\documentclass[lettersize,journal]{IEEEtran}
\usepackage{amsmath,amsfonts,amssymb,mleftright}
\usepackage{algorithmic}
\usepackage{algorithm}
\usepackage{array}
\usepackage[caption=false, font=footnotesize,  subrefformat=parens,labelformat=parens]{subfig}

\usepackage{textcomp}
\usepackage{stfloats}
\usepackage{url}
\usepackage{verbatim}
\usepackage{graphicx}
\usepackage{cite}
\usepackage{makecell}
\usepackage{varwidth}

\usepackage{xparse}
\usepackage{xcolor}
\usepackage{cases}
\usepackage{cleveref}
 \usepackage{multirow}
 \usepackage{tabularx,booktabs}
 
\usepackage{colortbl}
\usepackage{booktabs,ragged2e}
\usepackage[flushleft]{threeparttable}
 \usepackage{enumitem}
\setlist[enumerate]{label=\roman*}

\long\def\comment#1{}

\DeclareMathOperator*{\argmin}{arg\,min}

\newfont{\bbb}{msbm10 scaled 700}

\newfont{\bb}{msbm10 scaled 1100}

\newcommand{\av}{{\bf a}}

\newcommand{\ev}{{\bf e}}

\newcommand{\gv}{{\bf g}}

\newcommand{\mv}{{\bf m}}

\newcommand{\uv}{{\bf u}}

\newcommand{\vv}{{\bf v}}
\newcommand{\xv}{{\bf x}}
\newcommand{\yv}{{\bf y}}
\newcommand{\zv}{{\bf z}}

\newcommand{\Am}{{\bf A}}
\newcommand{\Bm}{{\bf B}}

\newcommand{\Fm}{{\bf F}}
\newcommand{\Gm}{{\bf G}}
\newcommand{\Hm}{{\bf H}}
\newcommand{\Id}{{\bf I}}

\newcommand{\Pm}{{\bf P}}
\newcommand{\Qm}{{\bf Q}}

\newcommand{\Sm}{{\bf S}}
\newcommand{\Tm}{{\bf T}}
\newcommand{\Um}{{\bf U}}

\newcommand{\Xm}{{\bf X}}
\newcommand{\Ym}{{\bf Y}}
\newcommand{\Zm}{{\bf Z}}

\newcommand{\deltav}{\hbox{\boldmath$\delta$}}

\newcommand{\muv}{\hbox{\boldmath$\mu$}}

\newcommand{\Lambdam}{\hbox{\boldmath$\Lambda$}}
\newcommand{\Deltam}{\hbox{\boldmath$\Delta$}}
\newcommand{\Sigmam}{\hbox{\boldmath$\Sigma$}}

\newcommand{\Thetam}{\hbox{\boldmath$\Theta$}}

\newcommand{\diag}{{\hbox{diag}}}

\newcommand{\trace}{{\hbox{tr}}}

\NewDocumentCommand{\evalat}{sO{\big}mm}{%
	\IfBooleanTF{#1}
	{\mleft. #3 \mright|_{#4}}
	{#3#2|_{#4}}%
}
\newtheorem{theorem}{Theorem}
\newtheorem{lemma}{Lemma}
\newtheorem{remark}{Remark}

\begin{document}

\title{Regularized Linear Discriminant Analysis Using a Nonlinear Covariance Matrix Estimator}

\author{Maaz~Mahadi,~ 
        Tarig~Ballal,~\IEEEmembership{Member, IEEE},
        Muhammad Moinuddin,~\IEEEmembership{Member, IEEE},
        Tareq Y. Al-Naffouri,~\IEEEmembership{ Senior Member, IEEE},
        and~Ubaid~M.~Al-Saggaf,~\IEEEmembership{Member, IEEE}.
\thanks{Maaz Mahadi, Tarig Ballal, and Tareq Y. Al-Naffouri are with Electrical and Computer Engineering Program, Computer, Electrical and Mathematical Sciences and Engineering (CEMSE), King Abdullah University of Science and Technology (KAUST), Thuwal 23955-6900, Kingdom of Saudi Arabia, (e-mail: {maaz.mahadi; tarig.ahmed; tareq.alnaffouri}@kaust.edu.sa).}
\thanks{Muhammad Moinuddin and Ubaid M. Al-Saggaf are with the Department of Electrical and Computer Engineering, King Abdulaziz University, Jeddah 21589, Saudi Arabia,
	and also with the Center of Excellence in Intelligent Engineering Systems
	(CEIES), King Abdulaziz University, Jeddah 21589, Saudi Arabia (e-mail: mmsansari; usaggaf@kau.edu.sa).}
	\thanks{© 2024 IEEE.  Personal use of this material is permitted.  Permission from IEEE must be obtained for all other uses, in any current or future media, including reprinting/republishing this material for advertising or promotional purposes, creating new collective works, for resale or redistribution to servers or lists, or reuse of any copyrighted component of this work in other works.}
}



\maketitle

\begin{abstract}
Linear discriminant analysis (LDA) is a widely used technique for data classification. The method offers adequate performance in many classification problems, but it becomes inefficient when the data covariance matrix is ill-conditioned. This often occurs when the feature space's dimensionality is higher than or comparable to the training data size. Regularized LDA (RLDA) methods based on regularized linear estimators of the data covariance matrix have been proposed to cope with such a situation. The performance of RLDA methods is well studied, with optimal regularization schemes already proposed. In this paper, we investigate the capability of a positive semidefinite ridge-type estimator of the inverse covariance matrix that coincides with a nonlinear (NL) covariance matrix estimator. The estimator is derived by reformulating the score function of the optimal classifier utilizing linear estimation methods, which eventually results in the proposed NL-RLDA classifier. We derive asymptotic and consistent estimators of the proposed technique's misclassification rate under the assumptions of a double-asymptotic regime and multivariate Gaussian model for the classes. The consistent estimator, coupled with a one-dimensional grid search, is used to set the value of the regularization parameter required for the proposed NL-RLDA classifier. Performance evaluations based on both synthetic and real data demonstrate the effectiveness of the proposed classifier. The proposed technique outperforms state-of-art methods over multiple datasets. When compared to state-of-the-art methods across various datasets, the proposed technique exhibits superior performance.
\end{abstract}
%

%
\IEEEpeerreviewmaketitle

\section{Introduction}
The origins of linear discriminant analysis (LDA) date back to the pioneering work of Fisher \cite{1936-fisher-use}, \cite{2012-devijver-pattern}. Owing to its simplicity and efficiency, LDA has been used in many applications, including microarray data classification \cite{2015-chen-shrunken}, detection \cite{2013-jin-motor}, face recognition \cite{1997-belhumeur-eigenfaces} and hyperspectral image classification \cite{2017-wang-locality}, to name a few. Fisher discriminant analysis aims at maximizing the \textit{between-class} distance to \textit{within-class} variance ratio of two statistical datasets. The technique has been successfully applied in classification and dimensionality reduction tasks \cite{1962-anderson-introduction} \cite{2020-khalili}. In its simplest form, LDA assumes a common covariance matrix for all the classes, and it ideally requires knowledge of the true statistical mean and covariance matrix. However, these parameters are practically unattainable. Computational methods need to be applied to estimate these statistical parameters from available data samples \cite{2020-khalili}.

Despite its good performance in various applications, LDA methods do not perform well when the data covariance matrix is \textit{ill-conditioned} or \textit{singular}. Such situations arise in many practical scenarios when the dimensionality of the
feature space of the data is higher than, or comparable to,
the size of the training data (the observed examples used
to estimate the covariance matrix) \cite{2009-bandos-classification}.

A remedy for this problem is to \textit{shrink}, or \textit{regularize}, the sample covariance matrix by combining it with a biased (target) matrix in a weighted average manner. The resulting matrix is the \textit {regularized sample covariance matrix}, and the parameter that controls the shrinkage is the shrinkage parameter or regularization parameter given by $\gamma \in (0,1)$. Linear combinations of the sample covariance matrix and a target matrix or multi-target matrices are very common in the literature, e.g.,  \cite{2004-ledoit-well, 2005-schafer-shrinkage,2019-ollila-optimal,2010-chen-shrinkage,2008-stoica-using, 2011-fisher-improved,2013-halbe-regularized,2012-chen-shrinkage,1989-friedman-regularized,2014-lancewicki-multi,2019-zhang-improved,2021-raninen-linear}. Usually, the covariance matrix estimation problem is formulated based on some optimization criteria, e.g., the mean square error \cite{2004-ledoit-well,2019-ollila-optimal}. Another form of regularization dates back to Horel \cite{1962-horel-applications} through his work on ridge regression, which works simply by replacing the unbiased least-squares estimator with a biased one that incorporates a  positive parameter, $\gamma  \in(0,\infty)$. This helps to improve the estimation in the mean-square-error sense. Di Pillo \cite{1976-di-application} applied Horel's method to estimate the sample covariance matrix used in LDA.

Besides these linear shrinkage-based estimators, there are nonlinear shrinkage estimators where
 the modification of the eigenvalues of the sample covariance matrix is achieved through a nonlinear transformation as presented in \cite{2012-ledoit-nonlinear,2015-ledoit-spectrum,2018-ledoit-optimal,2020-ledoit-analytical,2020-ledoit-power}. These methods are widely recognizable in finance applications \cite{2020-ledoit-power}. 
 The method presented in \cite{2012-ledoit-nonlinear} and \cite{2015-ledoit-spectrum} relies on the numerical inversion of a deterministic function, the quantized eigenvalues sampling transform (QuEST) function, which nonlinearly maps population eigenvalues into sample eigenvalues. The motivation for using nonlinear estimators instead of linear ones is the fact that a linear estimator is essentially a first-order approximation to a fundamentally nonlinear problem that governs the relation between the sample and population eigenvalues \cite{2012-ledoit-nonlinear}. 
 
In \cite{2014-abadir-design}, the estimation is done by splitting the sample data into two parts; one to estimate the eigenvectors of the covariance matrix, and the other to estimate the eigenvalues associated with these eigenvectors. The method performs well by averaging over many permutations of the sample split. A recent method proposed by Ledoit and Wolf \cite{2020-ledoit-analytical} combines the benefits of the aforementioned method. For the LDA problems, one can estimate the covariance matrix and invert it to obtain an estimation of the inverse covariance matrix, or the \textit{precision matrix} \cite{2005-schafer-shrinkage}, \cite{2012-ledoit-nonlinear}, \cite{2019-ollila-optimal}.

A better estimation approach for LDA is to estimate the inverse of the sample covariance matrix (the precision matrix). This provides a direct estimation and reduces the numerical errors induced by estimating the covariance matrix and then performing matrix inversion \cite{2012-ledoit-nonlinear, 2017-kuismin-estimation, 2020-kang-improvedprecision, 2020-bilgrau-targeted, 2019-van-generalized}. Estimating the precision matrix based on the aforementioned shrinkage/ridge estimators can be viewed as maximizing a non-familiar $l_2$-penalized log-likelihood function \cite{2016-van-ridge} which results from linear estimators. However, alternative ridge estimators for the precision matrix that resemble the original ridge $ l_2$-penalty have been proposed in the literature resulting in nonlinear estimators of the covariance matrix \cite{2016-van-ridge}, \cite{2017-kuismin-precision}.

Selecting a proper value of the regularization parameter is challenging \cite{2021-auguin-large}. Similar to \cite{2020-khalili} and \cite{2015-zollanvari}, the regularized parameter is selected by minimizing the estimated misclassification probability. However, the analysis and the derivation of the estimated probability error differ because of the newly proposed inverse covariance matrix estimator. The proposed estimator estimates the inverse pooled covariance matrix and represents an implicit nonlinear estimator of the pooled covariance matrix. The estimator is used to develop an RLDA classifier that we refer to as the NL-RLDA. 

The contributions of this paper can be summarized as follows:
\begin{itemize}
	\item Starting from the optimal score function of the Bayes classifier, we propose a nonlinear estimator of the pooled covariance matrix for RLDA. The application of the proposed estimator to RLDA results in our NL-RLDA classifier. 
	\item Under the assumption of data Gaussianity and the double asymptotic regime, we derive the asymptotic performance of the proposed classifier based on the misclassification rate.
	\item We derive a general consistent estimator of the misclassification rate performance of the proposed classifier.
	\item We utilize the derived misclassification rate expression to optimally tune the regularization parameter associated with the proposed covariance matrix estimator.  
\end{itemize}

Based on synthetic and real datasets, our results demonstrate that the proposed method can provide an advantage over other LDA-based classifiers that employ different covariance matrix estimators. The results also show the proposed method to be competitive when compared to other types of (non-LDA-based) classifiers. 

The remainder of this paper is organized as follows. In Section~\ref{sec:back}, we present background material on the basic LDA and RLDA approaches. Section~\ref{sec:proposed} introduces the proposed nonlinear covariance matrix estimator. The main results obtained by applying the proposed covariance matrix estimator to LDA and performance evaluation results based on synthetic and real data are presented in Section~\ref{sec:Performance Evaluation}. The conclusion of this paper is stated in Section~\ref{sec:conc}.

\subsection{Notations}
Uppercase boldface letters denote matrices, while lowercase boldface letters denote vectors. Scalars are denoted by normal lowercase letters. The superscript notation, $(.)^T$  denotes the transpose of a matrix or a vector, while $\mathbb{E}(.)$ denotes the statistical expectation operator, and $\text{tr}[.]$ is the trace of a matrix. $\mathbb{R}$, $\mathbb{R}^+$, and $\mathbb C$, respectively, denote real, positive-real, and complex fields of a dimension specified by a superscript. The variable $z$ denotes a complex variable. The notation $x \asymp y$ stipulates that $x$ and $y$ are asymptotically equivalent, i.e., $\vert x - y \vert \xrightarrow{\text{a.s.}}0 $, where a.s. denotes almost-sure convergence. The $l_2$-norm of a vector $\av$ is denoted by $\Vert \av \Vert_2$, and the 2-induced norm of a matrix $\Am$ is denoted by $\Vert \Am \Vert_2$, while the identity matrix of dimension $n$ is denoted by ${\bf I}_n$.

\section{Background}
\label{sec:back}
\subsection{Linear Discriminant Analysis (LDA)}
We consider a binary classification problem with two different populations $\mathcal{P}_0$, labeled as $\mathcal{C}_0$, and $\mathcal{P}_1$, labeled as $\mathcal{C}_1$. Each population has a multivariate Gaussian distribution with a common covariance matrix $\Sigmam \in \mathbb{R}^{p \times p}$ and different means $\boldsymbol{\mu}_i \in \mathbb{R}^{p \times 1}, i=0,1$. A random sample of $n = n_0 + n_1$ observations arranged in a matrix $\Xm = [{\xv}_1~ \xv_2 \cdots \xv_n]$, $\xv_j \in \mathbb{R}^{p \times 1}$,  such that $n_0$ observations are drawn  from $\mathcal{P}_0$ to constitute a set $\mathcal{G}_0 = \{ \xv_j \in \mathcal{C}_0\}_{j=1}^{n_0}$, . Another set, $\mathcal{G}_1 = \{ \xv_j \in \mathcal{C}_0\}_{j=1+n_0}^{n_0+n_1} $, is formed using $n_1$ random observations from $\mathcal{P}_1$. The probability of drawing an observation from $\mathcal{P}_0$ is $\pi_0$, and from $\mathcal{P}_1$ is $\pi_1$. 

We are looking for a discriminant rule to assign an input data vector to the class it most probably belongs to. If the distribution parameters of the observations are known, we can use the  score function \cite{2013-gareth-introduction}

\begin{align}		
{W}^{\text{Bayes}}(\xv) &=\left(\xv- {{{\boldsymbol {\mu}}_{0}+{\boldsymbol {\mu}}_{1}}\over {2}}\right) ^{T} {\Sigmam^{-1}}\left ({\boldsymbol {\mu}}_{0}-{\boldsymbol {\mu}}_{1}\right) 
\label{eqn:WBayes}
\end{align}
to assign the input to a class based on the following rule:

\begin{equation*}
    \xv \in
    \begin{cases}
    \mathcal{C}_0,    & {\text{if}}~ {W}^{\text{Bayes}}(\xv)> \tau,\\
    \mathcal{C}_1,    & {\text{if}}~ {W}^{\text{Bayes}}(\xv)\leq \tau,\\
    \end{cases}
    \end{equation*}
where $\tau = \log \dfrac{\pi_1}{\pi_0}$.

The score function in \eqref{eqn:WBayes} is the optimum Bayes classifier, which requires knowledge of the first- and second-order moments of the two classes. Practically, however, the statistics of each class are not known. Instead, we only have estimations of the covariance matrix and the mean vectors obtained from (limited) training data samples in the form of a \textit{pooled sample covariance matrix} and \textit{sample mean} vectors. The sample mean of class $i$ is calculated as
\begin{align}
\mv_{i}&= \dfrac{1}{n_i}\sum_{ \xv_{j} \in \mathcal{G}_i}  \xv_{j}, \label{eqn:sampleMean}
\end{align}
and the pooled sample covariance matrix is obtained using
\begin {equation}
\label{eqn:sampleCov}
\Sm= \dfrac{1}{n-2} \sum_{i\in\{0,1\}} (n_i-1)\Sm_i,
\end{equation}
where $n=n_0+n_1$, and $$\Sm_{i} = \dfrac{1}{n_i-1}\sum_{ \xv_{j} \in \mathcal{G}_i} (\xv_j - \mv_i )(\xv_j - \mv_i )^T.$$ 
Now, we can rewrite (\ref{eqn:WBayes}) based on the estimated statistics as
\begin{align}
\widehat{W}^{\text{LDA}}(\xv) &=\left(\xv- {{{\bf {m}}_{0}+{\bf {m}}_{1}}\over {2}}\right) ^{T} {\bf{S}}^{-1}\left ({ \mv}_{0}-{ \mv}_{1}\right).
\label{eqn:hatWlda}
\end{align}

This changes the classification rule to
\begin{equation}
\label{eqn:classify_x}
\xv \in
\begin{cases}
    \mathcal{C}_0,    & {\text{if}}~ \widehat{W}^{\text{LDA}}(\xv)> \hat{\tau},\\
    \mathcal{C}_1,    & {\text{if}}~ \widehat{W}^{\text{LDA}}(\xv)\leq \hat{\tau},\\
    \end{cases}
\end{equation}
where $\hat{\tau} = \log \dfrac{\hat{\pi}_1}{\hat{\pi}_0}, \hat{\pi}_0 = \dfrac{n_0}{n}$, and $\hat{\pi}_1 =\dfrac{n_1}{n}.$

The conditional  probability of misclassifying an observation from one class is defined as \cite{2015-zollanvari}
\begin{equation}
\varepsilon _{i}^{\text{LDA}} \triangleq \mathbb{P}[(-1)^{i}\widehat{W}^{\text{LDA}}(\xv)\leq (-1)^{i}\hat{\tau}\mid \xv\! \in \! \mathcal{C} _{i},\mathcal{G} _{0},\mathcal{G} _{1}],
\end{equation}
which can be expressed as \cite{2015-zollanvari}
\begin{equation} 
\varepsilon _{i}^{\text{LDA}}=\Phi \left ({{(-1)^{i+1}G\big (\muv_{i},\mv_{0}, \mv_{1}, \Sm\big)+(-1)^{i} {\hat{\tau}}}\over {\sqrt{D\big ( \mv_{0},\mv_{1}, \Sm, {{\Sigmam}}\big)}}}\right),
\end{equation}
where $\Phi $  denotes the cumulative distribution function of the standard Gaussian distribution, and

\begin{align}
G\big (\muv_{i}, \mv_{0}, \mv_{1}, \Sm\big)\triangleq &\, \left(\muv_{i}- {{ \mv_{0}+ \mv_{1}}\over {2}}\right) ^{T} \Sm^{-1}\left ( \mv_{0}- \mv_{1}\right), \\
D\big ( \mv_{0}, \mv_{1}, \Sm, {\Sigmam}\big)\triangleq &\, ( \mv_{0}- \mv_{1})^{T} \Sm^{-1} {\Sigmam} \Sm^{-1}( \mv_{0}-\mv_{1}).
\end{align}

The total (true) probability of misclassification is given by
\begin{equation}
	\varepsilon^{}= \pi_0\varepsilon_0 + \pi_1 \varepsilon_1. 
	\label{eqn:misclass}
\end{equation}

\subsection{Regularized Linear Discriminant Analysis (RLDA)}

We are interested in computing \eqref{eqn:hatWlda} using the available sample means and pooled sample covariance matrix. The latter raises problems when the size of the observed (or training) data is not much larger than the feature dimension. That covers the case where the feature dimension is comparable to the size of the observed data, i.e., $p\sim n$, and the case where $p>n$. In the former case, ${\bf{S}}$, can be ill-conditioned, and hence, the computation of \eqref{eqn:hatWlda} will potentially suffer from the occurrence of large errors. Indeed, for $p>n$, the matrix ${\bf{S}}$ is non-invertible.

To overcome the sample covariance matrix ill-conditioning and invertibility issues in LDA, we can estimate the covariance matrix and then invert the estimate, or we can directly estimate the inverse covariance matrix.
A number of regularized estimators of the covariance matrix and the precision matrix have been proposed in the literature.

Then, the RLDA score function can be expressed as
\begin{align}
\widehat{W}^{\text{RLDA}}(\xv) &=\left(\xv- {{{\bf {m}}_{0}+{\bf {m}}_{1}}\over {2}}\right)^{T} {\bf{H}}\left ({\mv}_{0}-{\mv}_{1}\right), 
\label{eqn:hatWrlda}
\end{align}
where $\Hm $ is an estimator of the inverse covariance matrix. For example, $\Hm $ can be a regularized covariance matrix estimator \cite{1989-friedman-regularized}, \cite{2015-zollanvari}. A popular choice of $\Hm $ is given as the inverse of regularized (linear) covariance matrix estimators of the forms (see \cite{2020-khalili}, \cite{2004-ledoit-well}, \cite{2019-ollila-optimal}, \cite{2015-zollanvari},  and \cite{2007-guo-regularized})
		\begin{subnumcases}
			{\Hm^\text{L}=}
			(\gamma \Sm + {\bf I}_{p})^{-1}, & $\gamma \in (0,\infty)$, \label{positive-subnum a}
			\\
			(\Sm + \gamma{\bf I}_{p})^{-1}, &  $\gamma \in (0,\infty)$,     \label{positive-subnum b}
            \\ 
            (\gamma \Sm_i + (1-\gamma) \Fm)^{-1}, &  $\gamma \in (0,1)$,
            \label{positive-subnum c}
		\end{subnumcases} 
where $\Fm$ is a target matrix that can have different structures \cite{2005-schafer-shrinkage}. These estimators are usually described as linear estimators.  
A widely used criterion for setting the regularization parameter is to minimize the mean square error between the true covariance matrix and the regularized estimator \cite{2004-ledoit-well}, \cite{2019-ollila-optimal}.

We can also directly estimate the inverse of the covariance matrix based on penalizing the log-likelihood function using linear estimators, or nonlinear estimators \cite{2016-van-ridge}, \cite{2017-kuismin-precision}. 

The conditional probability of misclassification for the score function (\ref{eqn:hatWrlda}) is given by \cite{2015-zollanvari}
\begin{equation} 
\label{eqn:epsilon^RLDA}
\varepsilon _{i}^{\text{RLDA}}=\Phi \left ({{(-1)^{i+1}G\big (\muv_{i},\mv_{0}, \mv_{1}, \Hm\big)+(-1)^{i} {\hat{\tau}}}\over {\sqrt{D\big ( \mv_{0},\mv_{1}, \Hm, {{\Sigmam}}\big)}}}\right),
\end{equation}
where
\begin{align}
G\big (\muv_{i}, \mv_{0}, \mv_{1}, \Hm\big)\triangleq&\, \left(\muv_{i}- {{ \mv_{0}+ \mv_{1}}\over {2}}\right) ^{T} \Hm\left ( \mv_{0}- \mv_{1}\right), \label{eqn:G} \\
D\big ( \mv_{0}, \mv_{1}, \Hm, {\Sigmam}\big)\triangleq&\, ( \mv_{0}- \mv_{1})^{T} \Hm {\Sigmam} \Hm( \mv_{0}-\mv_{1}). \label{eqn:D}
\end{align}

\begin{remark}
    \label{rem:1}
    It should be noted here that for $\varepsilon _{i}^{\text{RLDA}}$ to be well defined, it requires that $D\big ( \mv_{0}, \mv_{1}, \Hm, {\Sigmam}\big) \neq 0$. In the case where $D\big ( \mv_{0}, \mv_{1}, \Hm, {\Sigmam}\big) = 0$, the performance evaluation becomes a trivial task, as the misclassification rate can readily be deduced from \eqref{eqn:hatWlda} and \eqref {eqn:classify_x} to be equal to $\min (\hat{\pi}_0, \hat{\pi}_1)$. In this case, LDA is not sensitive to the data, and hence should not be applied to the classification task.
\end{remark}

\section{The proposed Regularized LDA Approach}
\label{sec:proposed}
In this section, we present our RLDA approach based on a proposed estimator. After developing the proposed classifier, we will derive the asymptotic performance and a consistent estimator of the proposed classifier's performance. We will also highlight how to obtain the regularization parameter using the derived consistent estimator. 

We start by establishing the proposed inverse covariance matrix estimator. Then, we apply it to obtain the score function of the proposed classifier. 

To motivate our approach, we start from the  optimal score function (\ref{eqn:WBayes}), which can be written in the form
\begin{align}
\label{eqn:LDAnew}
W^{\text{Bayes}}(\xv)  &= {{1}\over{2}} \sum_{i\in\{0,1\}} (-1)^i\left(\xv - \muv_i\right)^T  {\Sigmam^{-1}} \left(\xv - \muv_i\right)     \\ 
& = {{1}\over{2}} \sum_{i\in\{0,1\}} (-1)^i \left(\xv - \muv_i\right)^T  {\Sigmam^{-\frac{1}{2}}} {\Sigmam^{-\frac{1}{2}} \left(\xv - \muv_i\right) } \nonumber\\ 
& = {{1}\over{2}} \sum_{i\in\{0,1\}} (-1)^i  \gv_i^T\gv_i ,\quad i=0,1, \label{eqn:WLDA2}
\end{align}
where ${\bf {{g}}}_i\triangleq {\Sigmam^{-\frac{1}{2}}}\left(\xv - \muv_i\right)$. Note that for $j, k\in\{0, 1\}, j\neq k$, $\gv_j \sim \mathcal{N}(\boldsymbol{0}, \Id_p)$ when $\xv$ is drawn from class $j$; otherwise, if $\xv$ is from class $k$, we have $\gv_k \sim \mathcal{N}(\muv_k-\muv_j, \Id_p)$.  

We want to find a good estimate of the score function (\ref{eqn:WLDA2}) when only the sample means and the pooled sample covariance matrix are available. Unlike the RLDA approach discussed in the previous section, we start from the very basic definition of ${ \bf g}_i$ required in \eqref{eqn:WLDA2} to obtain the LDA score. From that definition, we observe that we can utilize the following linear model: 
\begin{align}
\av_i &=   \left( \Sm^\frac{1}{2} + \Deltam \right) \gv_i + \deltav_i,\quad i=0, 1,
\label{eqn:linearModel}
\end{align}
where $\av_i\triangleq \xv - \mv_i$, $\Delta = \Sigmam^\frac{1}{2} - \Sm^\frac{1}{2}$, and $\deltav_i = \muv_i - \mv_i$. 

The model \eqref{eqn:linearModel} is well recognized and arises naturally in many practical problems, albeit with different matrix structures~\cite{1997-chandrasekaran-parameter}. We can use this model to find an estimate of the vector $\gv_i$ required to compute the LDA score function. One possible approach is to seek an estimate $\hat{\gv}_i$ of $\gv_i$ which minimizes the maximum possible residual norm. This results in the following optimization~\cite{1997-chandrasekaran-parameter}: 
\begin{align}
&\min _{\hat{\mathbf{g}}_i } \Vert (\mathbf {S}^{\frac{1}{2}} + \Deltam) \hat{\mathbf{g}}_i  - {\bf a}_i\Vert_{2}, \quad i=0, 1 \nonumber \\
& \text{subject to:}\quad \Vert \Deltam \Vert_{2} \le \alpha,
\label{eqn:BDU}
\end{align}
where $\alpha$ is a positive scalar bound. 

It is shown in \cite{1997-chandrasekaran-parameter} that the solution of \eqref{eqn:BDU} is given by 
\begin{align}
\hat{\gv}_i &= (\Sm + \gamma_i {\bf I}_p)^{-1}\Sm^{\frac{1}{2}}{\bf a}_i, \quad i=0, 1,
\label{eqn:ghat}
\end{align}
where $\gamma_i \in \mathbb{R}^+$. It can be observed that the solution \eqref{eqn:ghat} has an identical form to that of the regularized least-squares estimator of $\gv_i$ based on the model \eqref{eqn:linearModel} when $\Deltam$ is equal to zero. Hence, we can think of $\gamma_i$, simply, as a regularization parameter. In \cite{1997-chandrasekaran-parameter}, it is shown that each parameter $\gamma_i$ can be obtained by solving a nonlinear equation. Unfortunately, the formula in \cite{1997-chandrasekaran-parameter} requires explicit knowledge of the perturbation bound $\alpha$. Therefore, for now, the values of the regularization parameters will remain unknown. 

Next, we can substitute the estimates \eqref{eqn:ghat} in \eqref{eqn:WLDA2} to obtain the expression of the score as a function of $\Sm$, $\av_i$ and $\gamma_i$. To simplify the development of the proposed approach, it is found compelling to set $\gamma_0=\gamma_1 = \gamma$ before substituting \eqref{eqn:ghat} in \eqref{eqn:WLDA2} to obtain
\begin{align}
&\tilde{W}=   {{1}\over{2}} \sum_{i\in\{0,1\}} (-1)^i  \hat{\gv}_i^T\hat{\gv}_i \nonumber \\
&= {{1}\over{2}} \sum_{i\in\{0,1\}} (-1)^i \left(\xv - \mv_i\right)^T \tilde{\Hm}\left(\xv - \mv_i\right), 
\label{eqn:proposedWeights}
\end{align}
where 
\begin{align}
\label{eqn:HNL} 
\tilde{\Hm}
&\triangleq \Sm^{\frac{1}{2}}(\Sm + \gamma{\bf I}_p)^{-2}\Sm^{\frac{1}{2}} \nonumber\\
&= (\Sm + \gamma{\bf I}_p)^{-2} \Sm\\
&= \Sm(\Sm + \gamma{\bf I}_p)^{-2}. \nonumber
\end{align}

Setting $\gamma_0=\gamma_1$ serves two purposes. First, it simplifies the derivation of the asymptotic performance and regularization parameter selection algorithm which will be discussed later. Second, it preserves the linearity of the score function,  as using two different regularization parameters prevents the two quadratic terms in \eqref{eqn:proposedWeights} from canceling each other. We can perceive \eqref{eqn:proposedWeights} as a simple estimator of \eqref{eqn:LDAnew} where $\mv_i$ are plugged in as estimates of $\muv_i$. On the other hand, the inverse covariance matrix $\Sigmam^{-1}$ is replaced with the \textit{nonlinear} estimator $\tilde{\Hm}$ given in \eqref{eqn:HNL}.  

\subsection{Interpretation of the proposed (inverse) covariance matrix estimator}
First, consider the eigenvalue decomposition of the pooled sample covariance matrix $\Sm = \Um \Lambdam \Um^\text{T}$, where $\Um \in \mathbb{R}^{p \times p}$ is the orthogonal matrix of the eigenvectors and $\Lambdam \in \mathbb{R}^{p \times p}$ is the diagonal matrix of the eigenvalues of $\Sm$. Given this decomposition, we can write \eqref{eqn:HNL} as
\begin{align}
\label{eqn:HNL EVD} 
\tilde{\Hm}
= \Um \Lambdam \left(\Lambdam + \gamma \Id_p \right)^{-2} \Um^\text{T},
\end{align}
which represents the inverse of the covariance matrix $\tilde{\Sm}\triangleq \Um \Lambdam^\dagger \left(\Lambdam + \gamma \Id_p \right)^{2} \Um^\text{T} =  \Sm^\dagger \left(\Sm + \gamma \Id_p \right)^{2}$, where $(.)^\dagger$ is the matrix pseudo-inverse. It is clear that in $\tilde{\Sm}$ the eigenvalues undergo a nonlinear transformation, and $\tilde{\Sm}$ is an obvious nonlinear function of the pooled sample covariance matrix. That said, in this paper the expression of $\tilde{\Sm}$ is not used, we rather focus on that of its inverse $\tilde{\Hm}$.

Given \eqref{eqn:HNL EVD}, any element ($k,l$) of $\tilde{\Hm}$ can be expressed as
\begin{align}
\label{eqn:HNL EV sum} 
[\tilde{\Hm}]_{k,l} 
= \sum_{d=1}^p \tilde{f} (\lambda_d)\frac{1}{\lambda_d} \left(\ev_k^\text{T} \uv_d \right)  \left(\uv_d^\text{T} \ev_l \right), k, l \in \{1, \cdots, p\},
\end{align}
where $\tilde{f}(\lambda_d) \triangleq \lambda_d^2/\left(\lambda_d + \gamma \right)^2$, $\lambda_d$ is the $d$-th eigenvalue, $\uv_d \in \mathbb{R}^{p}$ is the $d$-th eigenvector, and $\ev_k \in \mathbb{R}^{p}$ is the standard basis vector of order $k$ and likewise for $\ev_l$.

The parameters $\tilde{f}(\lambda_d)$ in \eqref{eqn:HNL EV sum} can be viewed as \textit{weights} or \textit{filter coefficients} that act on the contributions of the eigenvalues of $\Sm$ to the inverse covariance matrix estimate. We can easily see that these coefficients are zero for zero eigenvalues. Generally, these coefficients are small for small eigenvalues, and they increase as the eigenvalues increase to approach unity for very large eigenvalues, i.e., $\lim_{\lambda_d\to\infty} \tilde{f}(\lambda_d) = 1$. This property of the coefficients allows for a complete removal of, and a sizeable reduction in, the contributions of the zero and small eigenvalues, respectively. On the other hand, the contributions of the large eigenvalues are adequately emphasized. Hence, we expect the overall impact of the coefficients $\tilde{f}(\lambda_d)$ to lead to improvement in estimation accuracy.

As an example, it should be noted here that for the (linear) covariance matrix estimator of the form \eqref{positive-subnum b}, the corresponding inverse covariance matrix estimator can be expressed in a form similar to \eqref{eqn:HNL EV sum}, yet with coefficients $f^\text{L}(\lambda_d) \triangleq \lambda_d/\left(\lambda_d + \gamma \right)$. Assuming that $\lambda_1>\lambda_2>\cdots>\lambda_p$, we can define the ratios 
\begin{align}
\label{eqn:coeff ratio L} 
\rho^\text{L}(\lambda_d) 
\triangleq \frac{f^\text{L}(\lambda_d)}{f^\text{L}(\lambda_1)} 
= \frac{\lambda_1 \lambda_d + \gamma \lambda_d}{\lambda_1 \lambda_d + \gamma \lambda_1}
\leq 1, \forall d=1,\cdots,p;
\end{align}
and
\begin{align}
\label{eqn:coeff ratio NL} 
\tilde{\rho}(\lambda_d) 
&\triangleq \frac{\tilde{f}(\lambda_d)}{\tilde{f}(\lambda_1)} 
= \left(\frac{\lambda_1 \lambda_d + \gamma \lambda_d}{\lambda_1 \lambda_d + \gamma \lambda_1}\right)^2 
=\left(\rho^\text{L}(\lambda_d) \right)^2 \nonumber \\
&\leq \rho^\text{L}(\lambda_d)
\leq 1, \quad  \forall d=1,\cdots,p.
\end{align}
These ratios reflect the relative contributions of various eigenvalues in the computation of the respective inverse covariance matrix. It is obvious that $\rho^\text{L}(\lambda_d) = \tilde{\rho}(\lambda_d)= 1$ only for $\lambda_d = \lambda_1$, and $\rho^\text{L}(\lambda_d) = \tilde{\rho}(\lambda_d)= 0$ for any $\lambda_d=0$. For the rest of the eigenvalues, \eqref{eqn:coeff ratio NL} clearly indicates that for any value of $\gamma$, the relative contribution of those eigenvalues is smaller in \eqref{eqn:HNL} than in \eqref{positive-subnum b}. This implies that the proposed covariance matrix estimator de-emphasizes the contributions of the \textit{smaller} eigenvalues (to the inverse) more compared to the linear estimator in \eqref{positive-subnum b}. This can be especially useful when some of the non-zero eigenvalues of the sample covariance matrix are very small. Such a situation might arise, for example, due to the insufficiency of the number of training samples.

\begin{remark}
    \label{rem:2}
    If we consider classical consistency (when the number of data samples grows large), as $n$ increases (while $p$ is fixed), $\tilde{\mathbf{H}}$ will converge to $\boldsymbol{\Sigma}^{-1}$ only if $\gamma$ is adjusted properly. Obviously, $\gamma$ should approach zero as $n$ approaches $\infty$ if consistency is to be maintained. It is clear from this discussion that, by definition, classical consistency eliminates the $p>n$ case. On the other hand, if we consider the double asymptotic regime where $p$ and $n$ grow at a fixed ratio, in the $p>n$ case, the zero eigenvalues of $\tilde{\mathbf{H}}$ precludes consistency in this regime, as the eigenvalues of $\tilde{\mathbf{H}}$ will not converge to those of $\boldsymbol{\Sigma}^{-1}$. 
\end{remark}

In the following subsections, we establish the asymptotic performance of LDA based on the score function given by \eqref{eqn:proposedWeights}. Unless confusion arises, for simplicity of notations, we will use $\Hm$ instead of $\tilde{\Hm}$. In the following subsection, we define some quantities that simplify our derivations. 

\subsection{Definitions}
\label{sec:def}
To keep the notations compact, we use the definitions in Table~\ref{tab:def}, where $\Thetam$, $\Am$, $\Bm \in \mathbb{R}^{p \times p}$, $\sigma_i $, the eigenvalue $i$ of $\Sigmam$, $\tilde{n} = n-2$ and $c =\frac{p}{n}$.   

\begin{table}[h]
		\caption{Frequently used definitions}
		\label{tab:def}
		\setlength\tabcolsep{0pt} 
		
		\begin{tabular*}{0.9\columnwidth}{@{\extracolsep{0.3\columnwidth}} cl}
			\toprule
			Symbol & Definition 			 \\ 
			\midrule
			$e(z) = e$  & $\frac{1}{\tilde{n}} \text{tr} \Big[ \Sigmam \big(x\Sigmam-z{\bf I}_p\big)^{-1}\Big]$ \\
			$x(z) =x$  & $\dfrac{1}{1+e}$\\
			$\phi(z) = \phi$ & $ \frac{1}{\tilde{n}} \text{tr} \Big[ \Sigmam^2 \big(x\Sigmam-z{\bf I}_p\big)^{-2}\Big]$ \\
			$\tilde{\phi}(z) = \tilde{\phi}$ & $ \dfrac{1}{(1+e)^2}$ \\
			\addlinespace
			$\Qm(z) = \Qm$ & $ (\Sm- z {\bf I}_p)^{-1}$ \\
			$b(z) = b$        &   $ \frac{1}{\tilde{n}} \sum_{i=1}^{\tilde{n}} \frac{1}{\sigma_i(1-c-czb)-z}$ \\
			$w(z) = w $     & $1 - czb$ \\
 			$\Pm(z) = \Pm$  & $\Big(w\Sigmam-z{\bf I }_p\Big)^{-1}$  \\
			\addlinespace
			$\xi_{\Am,\Bm}$& $\frac{1}{\tilde{n}} \text{tr}\big[\Sigmam \Am \Sigmam \Bm \big]$ \\
			\bottomrule
		\end{tabular*}
\end{table}	

\begin{figure*}[!t]
    \normalsize
    \begin{align}
    &\eta_{\text{\small \Thetam}} = \Big(  \dfrac{1}{1-\phi\tilde{\phi}}-\dfrac{\phi \tilde{\phi}^\prime + \phi^\prime \tilde{\phi} + 2\frac{x^\prime}{x}(1-\phi \tilde{\phi})}{(1-\phi \tilde{\phi})^2}z \Big) \trace\Big[\Thetam \Sigmam (x\Sigmam-z{\bf I}_p)^{-2}\Big] +   2\Big(  \dfrac{ z-\frac{x^\prime}{x}z^2}{(1-\phi \tilde{\phi})^2} \Big) \trace\Big[\Thetam \Sigmam (x\Sigmam-z{\bf I}_p)^{-3}\Big] \nonumber\\
    &+ \dfrac{2(ww^\prime \xi_{P,P} + w^2 \xi_{P, P^\prime})^2+(1-w^2 \xi_{P,P})(w^{\prime 2}\xi_{P,P}+2ww^\prime \xi_{P,P^\prime}+w^2\xi_{P^\prime,P^\prime})}{(1-w^2\xi_{P,P})^3}z^2\trace[{\Thetam\Pm\Sigmam \Pm}] \nonumber\\
    &+ 2\dfrac{ww^\prime\xi_{P,P} + w^2\xi_{P,P^\prime}}{(1-w^2\xi_{P,P})^2}z^2\trace[{\Thetam \Pm\Sigmam\Pm^\prime}] + \dfrac{1}{(1-w^2\xi_{P,P})}z^2 \trace[{\Thetam \Pm^\prime\Sigmam \Pm^\prime}].
    \label{eqn:asym}
    \end{align}
    \hrulefill
    \vspace*{4pt}
\end{figure*}

\subsection{Asymptotic Performance}
In this section, we pursue an asymptotic analysis (in the double asymptotic regime) of the proposed RLDA classifier based on the proposed nonlinear estimator \eqref{eqn:HNL}. To this end, we utilize results from the field of random matrix theory (RMT) \cite{2016-muller-random}, \cite{2016-suliman-RMT}. A similar analysis for the RLDA classifier that uses \eqref{positive-subnum a} was conducted in \cite{2015-zollanvari}. However, applying the proposed estimator makes the analysis more involved. Before delving into the analysis, we state the following growth-rate assumptions required for the derivations \cite{2020-khalili}:
\begin{enumerate}[start=1,label={\bfseries Assumption~\arabic*:},leftmargin=*,labelindent=0em]
\item  (data scaling): $\frac{p}{n} \rightarrow d \in (0,\infty)$.
\item  (class scaling): $\frac{n_i}{n} \rightarrow d_i \in (0,\infty)$ 
\item (covariance scaling): $\lim \sup_p \Vert \boldsymbol{ \Sigma}\Vert < \infty$
\item  (mean scaling): $\lim \sup_p \Vert\muv_0-\muv_1 \Vert < \infty$
\end{enumerate}
Assumption~1 controls the growth rate in data, as it implies that both the sample size and the data dimension grow large with the same order of magnitude. Similarly, Assumption~2 controls the growth rate in training. Assumption~3 is usually met within the random matrix theory (RMT) framework \cite{2016-benaych-spectral}, and Assumption~4 guarantees that we can obtain a nontrivial classification error rate \cite{2020-khalili}.

\begin{theorem}
    \label{thm:1}
    Under Assumptions~(1)--(4), and $\Sm(\mv_0 - \mv_1) \neq \boldsymbol{0}$, the following relations hold true for the quantities defined in \eqref{eqn:G} and \eqref{eqn:D}:
    \begin{align}
        G\big (\muv_{i},\mv_{0}, \mv_{1}, \Hm\big)& \asymp \evalat{\widetilde{G}_i(z)}{z=-\gamma}  \\
        D\big (\mv_{0}, \mv_{1}, \Hm, {\Sigmam }\big)& \asymp \evalat{\widetilde{D}(z)}{z=-\gamma} ,	
    \end{align}
    where
    \begin{align}
        \widetilde{G}_i(z) &= \dfrac{(-1)^i}{2} (x-x^\prime z)\Bigg [ \boldsymbol{ \mu}^T\Sigmam \left(  x\Sigmam-z{\bf I} \right)^{-2}\boldsymbol{ \mu}\nonumber\\
        &+ \left( \dfrac{1}{n}_1-\dfrac{1}{n}_0 \right)\text{tr} \big[ \Sigmam^2 \left( x\Sigmam-z{\bf I} \right)^{-2} \big]  \Bigg],
        \label{eqn:Gasymp}
    \end{align}
    and 
    \begin{equation}
        \widetilde{D}(z) = { \eta}_{\boldsymbol{\mu \mu}^T} + \left( \dfrac{1}{n}_1+\dfrac{1}{n}_0 \right)\eta_{_\text{\tiny \Sigmam}}, 
    \end{equation}
    with ${ \eta}_{\boldsymbol{\mu \mu}^T} $ and $ \eta_{_\text{\tiny \Sigmam}}$ obtained from (\ref{eqn:asym}) using some of the definitions in Table~\ref{tab:def}.
\end{theorem}
\textit{Proof:} 
see Appendix B.

\subsection{The Consistent Estimator}
The following theorem provides a consistent estimator of the conditional error probability \eqref{eqn:epsilon^RLDA} using the proposed estimator.
\begin{theorem}
     \label{thm:2}
    Under Assumptions~(1)--(4), and $\Sm(\mv_0 - \mv_1) \neq \boldsymbol{0}$, the following conditional probability is a consistent estimator of the original conditional probability \eqref{eqn:epsilon^RLDA}: 
    \begin{equation} 
        \hat{\varepsilon} _{i}^{\text{NL-RLDA}} \triangleq \Phi \left ({{(-1)^{i+1}G\big ({\mv}_{i},\mv_{0}, \mv_{1}, \Hm\big)+ \frac{1}{n_i}\hat{\theta}_G+(-1)^{i} {{\hat{\tau}}}}\over {\sqrt{D_c\big ( \mv_{0},\mv_{1}, \Hm, \Sm\big)}}}\right),
        \label{eqn:consist}
    \end{equation}
    where 
    \begin{equation}
        \hat{\theta}_G = \dfrac{\hat{e}^\prime(\hat{x}-z\hat{x}^\prime)-{\dfrac{1}{n} \text{tr}[\Sm \Qm^2]}} {\hat{e}^\prime(1+\hat{e})^{-2}},	
        \label{eqn:thetaG}
    \end{equation}
    \begin{equation}
        \hat{e} = \dfrac{\dfrac{1}{\tilde{n}} \text{tr}[\Sm \Qm^2]}{1- \dfrac{1}{\tilde{n}} \text{tr}[\Sm \Qm]}.
        \label{eqn:consist eHat}
    \end{equation}
    
    In \eqref{eqn:consist}--\eqref{eqn:consist eHat} $\tilde{n}$ and $\Qm$ are defined in Section~\ref{sec:def}, while $D_c\big ( \mv_{0},\mv_{1}, \Hm, \Sm\big)$ is a consistent estimator of $D\big ( \mv_{0},\mv_{1}, \Hm, \Sigmam\big)$ given by
    \begin{align}
        &D_c\big ( \mv_{0},\mv_{1}, \Hm, \Sm \big) \nonumber\\
        &=z^2 (1+\hat{e})^4 \mv^T \Qm^2 \Sm\Qm^2 \mv + 2z(1+\hat{e})^2\mv^T \Qm^2 \Sm\Qm \boldsymbol\mv \nonumber \\
        & + ((1+\hat{e})^2+2z\hat{e}^{\prime}(1+\hat{e}))
        \mv^T \Qm\Sm\Qm \mv,
         \label{eqn:Dc}
    \end{align}
    where $\mv \triangleq \mv_0 - \mv_1.$
\end{theorem}

\textit{Proof:} see Appendix C.

Based on Theorem 2, We obtain the consistent estimator of the misclassification probability by substituting \eqref{eqn:consist} as follows:

\begin{align}
   \hat{\varepsilon}^{\text{NL-RLDA}} &= \hat{\pi}_0\hat{\varepsilon}_0^{\text{NL-RLDA}} + \hat{\pi}_1 \hat{\varepsilon}_1^{\text{NL-RLDA}} \label{eqn:vareps_NLRLDA} \\
    &= \dfrac{n_0}{n}\Phi \left ({{-G\big ({\mv}_{0},\mv_{0}, \mv_{1}, \Hm\big)+\frac{1}{n_0}\hat{\theta}_G+ \log \frac{\hat{\pi}_1}{\hat{\pi}_0} }\over {\sqrt{D_c\big ( \mv_{0},\mv_{1}, \Hm, \Sm \big)}}}\right)  \nonumber\\
    &+ \dfrac{n_1}{n}\Phi \left ({{G\big ({\mv}_{1},\mv_{0}, \mv_{1}, \Hm\big)+\frac{1}{n_1}\hat{\theta}_G- \log \frac{\hat{\pi}_1}{\hat{\pi}_0} }\over {\sqrt{D_c\big ( \mv_{0},\mv_{1}, \Hm, \Sm \big)}}}\right).
    \label{eqn:hat_epsilon}
\end{align} 

\begin{remark}
    \label{rem:3}
     In the $p>n$ case, it is possible to have $\Sm (\mv_0 - \mv_1) = \boldsymbol{0}$, even when $\mv_0 \neq \mv_1$ and $\Sm \neq \boldsymbol{0}$. This is due to the fact that $\Sm$ is not positive definite but rather positive semidefinite. In this case, the denominators in \eqref{eqn:hat_epsilon} can be equal to zero, and $\hat{\varepsilon}^{\text{NL-RLDA}}$ is n\textbf{}ot guaranteed to be a well-defined function. As alluded to in Remark~1, the performance of the RLDA in this case is trivial, and the classifier is ill-suited for the given classification task.
    
    The occurrence of $\Sm (\mv_0 - \mv_1) = 0$ when $\mv_0 \neq \mv_1$ and $\Sm \neq \boldsymbol{0}$ requires the vector of means difference to lie in the null space of the data sample covariance matrix. We notice that both $\Sm$ and $(\mv_0 - \mv_1)$ are \textit{continuous} random variables. Hence, $\Sm (\mv_0 - \mv_1)$ is a continuous random vector. This leads us to conclude that the probability of $\Sm (\mv_0 - \mv_1) = \boldsymbol{0}$ is equal to zero. For discrete data with reasonably large number of levels, the probability of $\Sm (\mv_0 - \mv_1)= \boldsymbol{0}$ will be very small. This is confirmed by evaluating $\Sm (\mv_0 - \mv_1)$ for various datasets.
\end{remark}

\subsection{Setting the regularization parameter value}
The performance of the classifier is highly dependent on the regularization parameter. It is desirable to find an optimum regularization parameter, $\gamma^{\text{opt}}$, that minimizes the classification error in \eqref{eqn:misclass}, i.e., 
\begin{equation}
\gamma^{\text{opt}} = \argmin\limits_{\gamma} {\varepsilon}.
\end{equation}
However, as pointed out in \cite{2016-bakir-efficient}, finding this optimum regularization parameter is infeasible because it requires perfect knowledge of the underlying distribution of the data. Even if the distribution parameters are known, solving the optimization problem to obtain the optimum $\gamma$ would not be easy as it requires the application of involved numerical techniques. 
As a practical alternative, we use \eqref{eqn:vareps_NLRLDA} as a surrogate of \eqref{eqn:misclass} to find the regularization parameter value that corresponds to the minimum estimated average error, i.e.,
\begin{equation}
	\gamma^{\text{o}} = \argmin\limits_{\gamma} \hat{{\varepsilon}}^{\text{NL-RLDA}}.
\end{equation}
To solve this minimization problem, we utilize a range-search technique. To this end, we define a grid of $\gamma$ values in an appropriate interval and evaluate (\ref{eqn:consist}) for each $\gamma$ value. We pick the value that coincides with the minimum estimated error as our regularization parameter.

\subsection{Summary of the proposed  classifier}
\label{subsec:Procedure1}
The following steps summarize the proposed classification approach:
\begin{itemize}
    \item \textbf{Training phase}
    \begin{enumerate}[label=\arabic*.]	
        \item  Estimate the sample means, $\mv_0$ and $\mv_1$, and the pooled covariance matrix, $\Sm$, from the training data using \eqref{eqn:sampleMean} and \eqref{eqn:sampleCov}, respectively.
        \item Initialize a grid range, then evaluate the consistent estimator of misclassification probability \eqref{eqn:hat_epsilon} at all grid points, and select the $\gamma$ value that estimates the minimum of \eqref{eqn:hat_epsilon}. 
    \end{enumerate}
    \item \textbf{Testing phase}
    \begin{enumerate}[label=\arabic*.]	 
        \item Using the test data sample, compute the score function according to \eqref{eqn:hatWrlda} with the proposed $\Hm$.
        \item Assign a class to the test data using the rule \eqref{eqn:classify_x}.
    \end{enumerate}
\end{itemize}

\subsection{Computational Complexity}
    It is easy to notice that implementing the proposed method according to Section ~\ref{subsec:Procedure1}  requires a complexity of $O(p^2n)$ for computing the SCM, and $O(p^3)$ for matrix multiplication and inversion which are needed in each grid-point. Hence, for the training phase, the overall complexity is $O(p^2n + kp^3)$, where $k$ is the number of grid points. The complexity can be reduced using the SVD trick when $p>n$ \cite{2009-hastie-elements}.
\section{Performance Evaluation}
\label{sec:Performance Evaluation}
In this section, we evaluate the proposed classifier's performance using both synthetic and real data. We compare the performance of the proposed classifier \footnote{The codes are made available via the link: {https://github.com/moaazhaj/NL-RLDA}} (summarized in Section~\ref{subsec:Procedure1}) with RLDA classifiers using different inverse covariance matrix estimators. These classifiers are the double asymptotic classifier, DA-RLDA \cite{2015-zollanvari}, \cite{2020-khalili}; alternative estimator Type I, Alt I-RLDA; Alternative estimator Type II, Alt II-RLDA; Ridge-type Operator for the Precision Matrix Estimator, ROPE-RLDA; the elliptical-based estimator classifier, ELL2-RLDA \cite{2019-ollila-optimal};  and the classifier based on the Ledoit-Wolf estimator, LW-RLDA \cite{2004-ledoit-well},  a linear estimator with diagonal entries of the SCM ($ \Fm = \diag(\Sm)$ ). We also benchmark nonlinear estimators of Quest1-RLDA and Quest2-RLDA \cite{2012-ledoit-nonlinear,2015-ledoit-spectrum}. All the aforementioned methods are RLDA classifiers based on different inverse covariance matrix estimators. Other techniques used for comparison but based on different principles are the logistic regression with ridge regression penalty, LR-RR; Support Vector Machine with 10-fold cross-validation (SVM-10fCV); and random forests (RF). Table~\ref{tab:methods} summarizes all the methods (with their settings) used for comparison and their implementation in R or Matlab.

	\begin{table*}[!h]
	\centering
	\caption{Classification methods used for comparison}
	\label{tab:methods}
	\begin{tabular}{@{}lp{6cm}lp{3cm}@{}}
		\toprule
		Method & Description & Matlab function or R-package & \begin{varwidth}{3cm}
		    Settings/Parameter(s) Selection Method
		\end{varwidth}  \\ \midrule
		DA-RLDA    &   \begin{varwidth}{6cm} RLDA based on double asymptotic G-consistent estimator of classification error \cite{2015-zollanvari}.   \end{varwidth} &  Similar to NL-RLDA  &   1-D search.   \\ 		
		ELL2-RLDA  &   RLDA based on CM estimation assumes elliptical distribution \cite{2019-ollila-optimal}.    & \verb|regscm|~\cite{2018-ollila-matlab} & Statistical estimation.         \\ 		
		LW-RLDA    &   RLDA based on Lediot-Wolf estimation method of the CM \cite{2004-ledoit-well}.  & \verb|regscm|~\cite{2018-ollila-matlab}  & Statistical estimation.        \\
		DIAG-RLDA  &    RLDA based on diagonal, unequal variances \cite{2005-schafer-shrinkage} & \verb|fitcdiscr|~      & 10-fold CV.    \\
		SCRDA      &       Shrunken Centroid Regularized Discriminant Analysis \cite{2007-guo-regularized}. & \verb|rda|~(R-package) & Two-parameters with 10-fold CV.  \\
		CRDA 		&     	Compressive Regularized Discriminant Analysis \cite{2018-tabassum-compressive}.  & 	 \verb|crda| &  Two-parameters with Statistical estimation.	\\
		LR-RR      &    Logistic regression with ridge regression penalty. &\verb|glmnet|~(R-package)    & 10-fold CV.      \\
		SVM-10fCV  &    SVM with 10-fold cross-validation. & \verb|e1071|~ (R-package) & 10-fold CV, linear kernel        \\
		RF         &    Random forest. &\verb|fitensemble|~ & `Bag' method, Number of trees  100.       \\
		Quest1-RLDA&    RLDA based on the Quest function 1 method (nonlinear shrinkage).    &   \verb|QuESTimate|    & Statistical estimation.       \\		
		Quest2-RLDA&    RLDA based on the Quest function 2 method (nonlinear shrinkage). & \verb|QuESTimate| & Statistical estimation. \\	
            Alt,I-RLDA & Alternative Type I Ridge Precision Estimator \cite{2016-van-ridge}.  & \verb|rags2ridges|   & LOOCV minimizing the log-likelihood, Target = \verb|default.target(S)| \\
             Alt,II-RLDA & Alternative Type II Ridge Precision Estimator, Target matrix is the zero matrix \cite{2016-van-ridge}. &\verb|rags2ridges|  & LOOCV minimizing the log-likelihood. \\
              ROPE-RLDA & Ridge-type Operator for the Precision matrix estimation \cite{2017-kuismin-precision}. & \verb|ROPE|    & 5-fold CV minimizing the log-likelihood.  \\
		\bottomrule
	\end{tabular}
\end{table*}

\subsection{Synthetic Data}

\begin{figure*}[!t]
\centering
\subfloat[$\nu^2=0.5,  n=50$]{\includegraphics[width=0.28\linewidth]{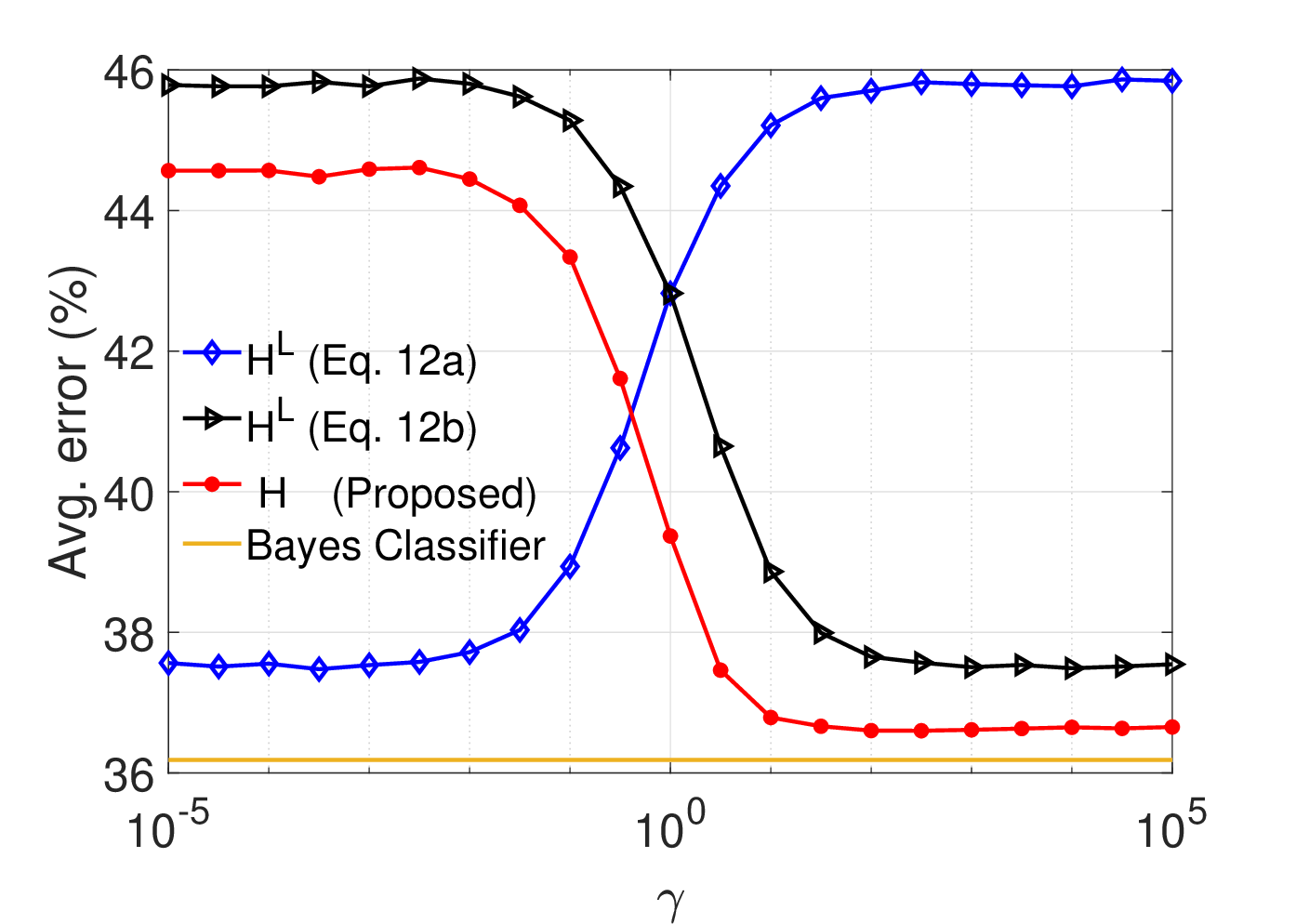}%
	\label{fig:fig1a}}
\hfil
\subfloat[$\nu^2=0.5,  n=100$]{\includegraphics[width=0.28\linewidth]{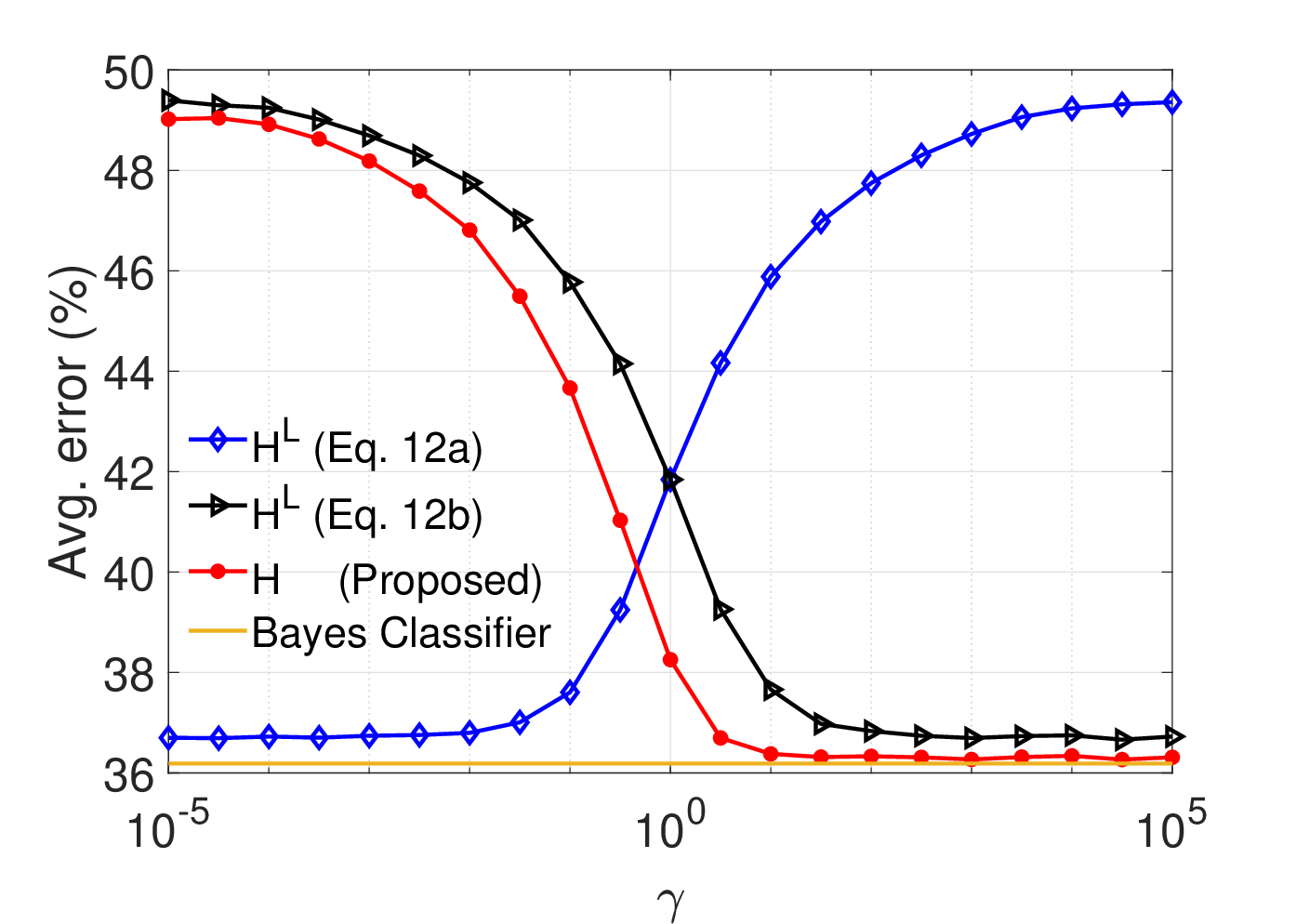}%
	\label{fig:fig1b}}
\hfil
\subfloat[$\nu^2=0.5, n=200$]{\includegraphics[width=0.28\linewidth]{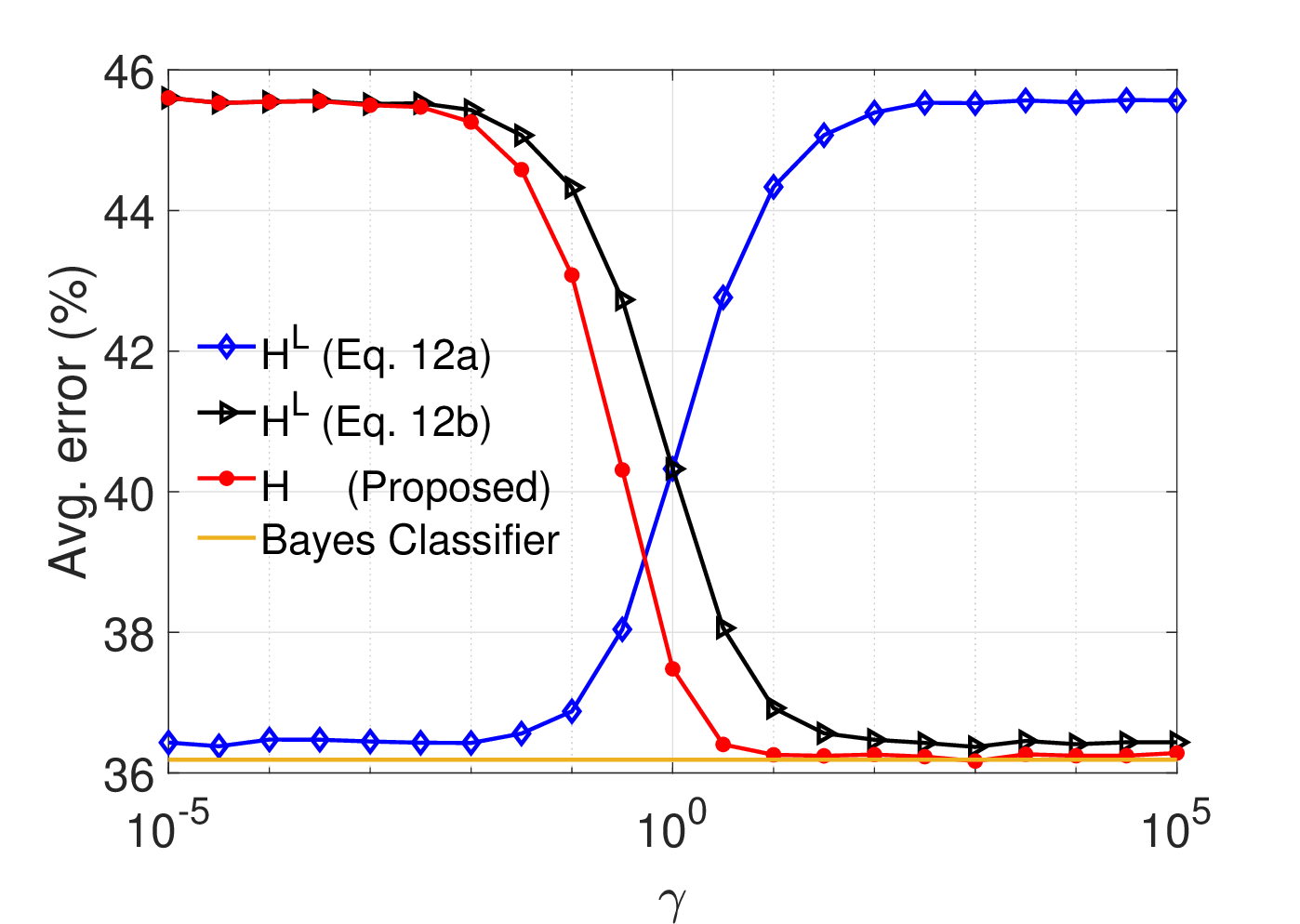}%
	\label{fig:fig1c}}
\\
\subfloat[$\nu^2=5, n=50$]{\includegraphics[width=0.28\linewidth]{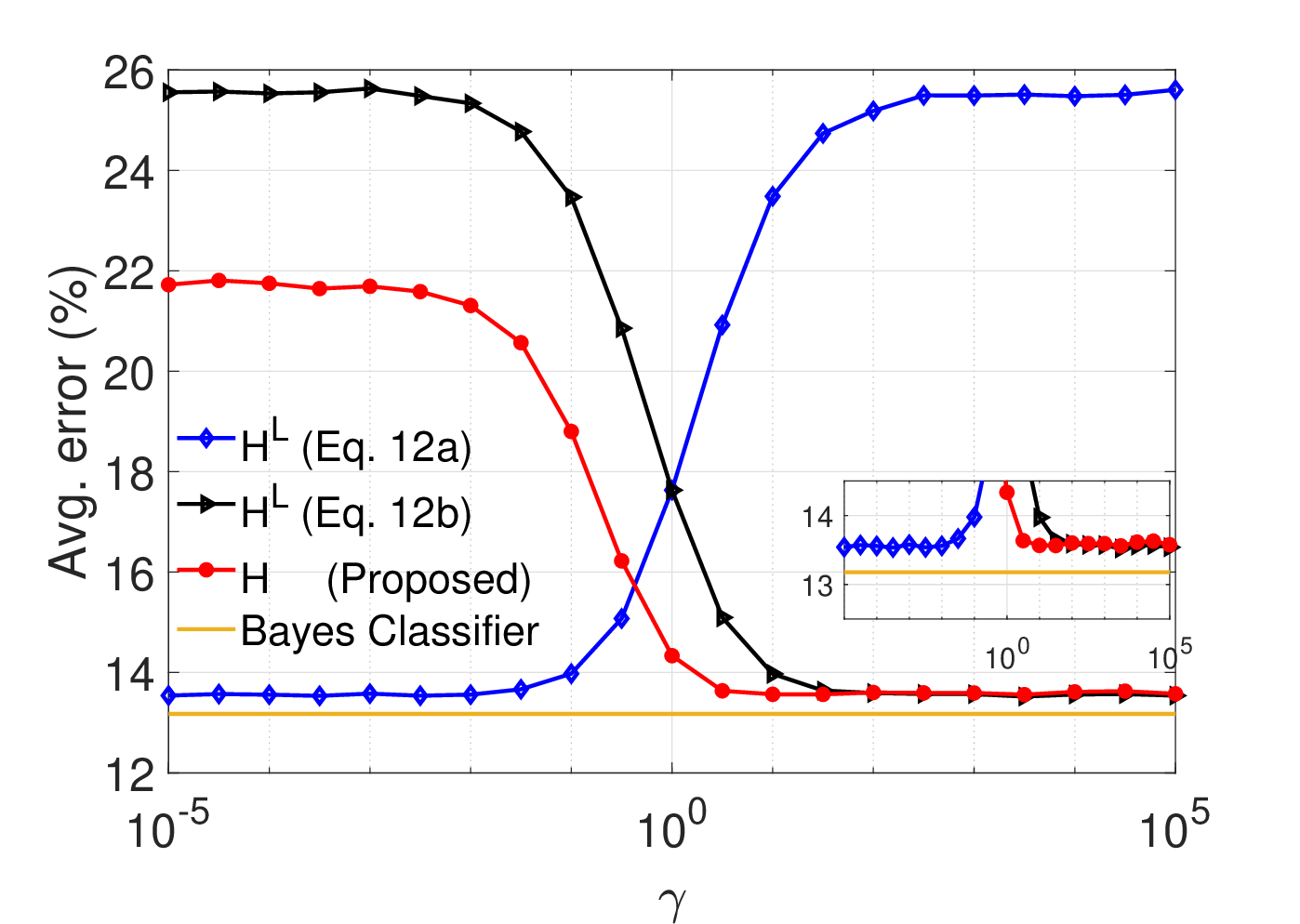}%
	\label{fig:fig1d}}
\hfil
\subfloat[$\nu^2=5, n=100$]{\includegraphics[width=0.28\linewidth]{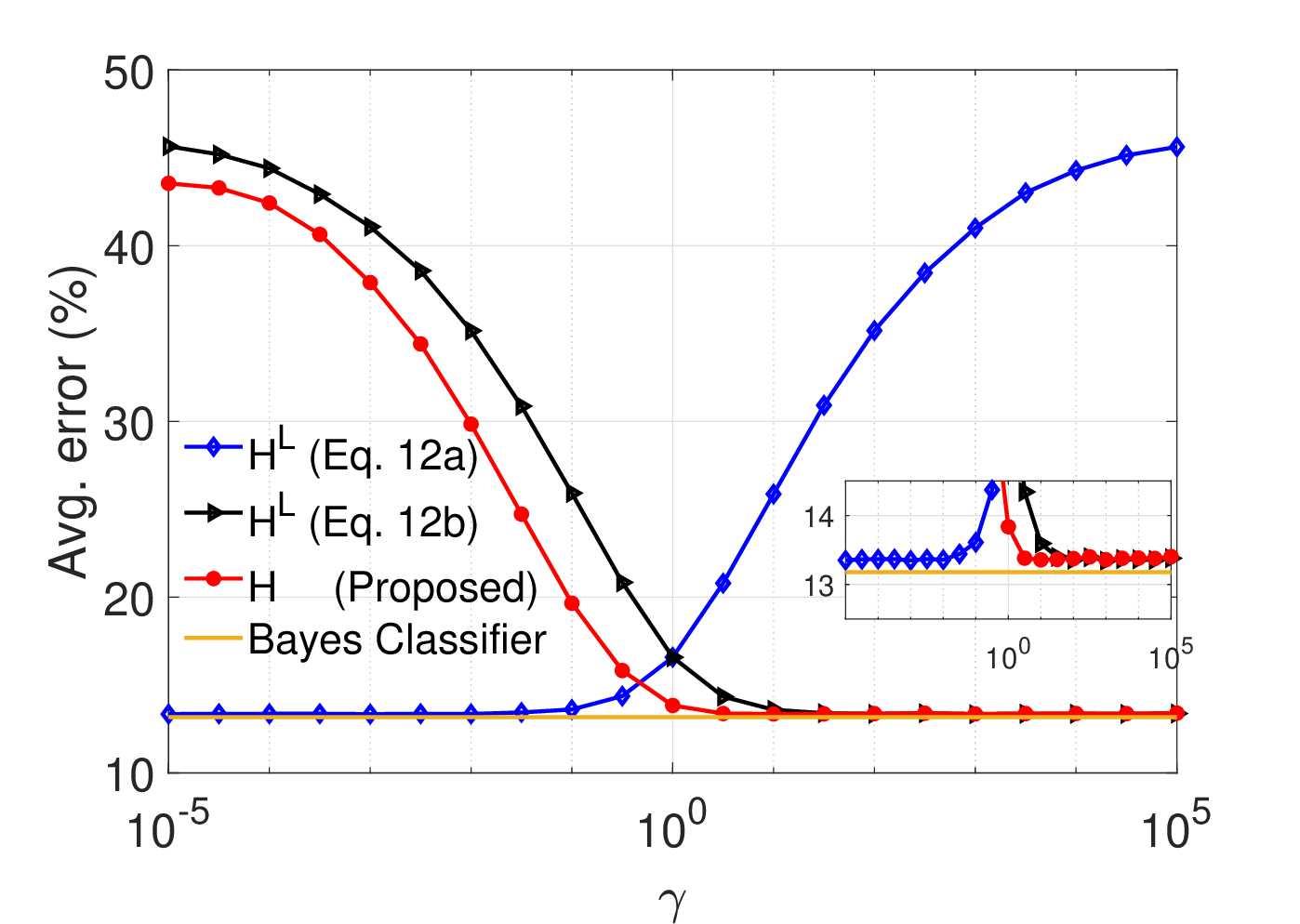}%
	\label{fig:fig1e}}
\hfil	
\subfloat[$\nu^2=5, n=200$]{\includegraphics[width=0.28\linewidth]{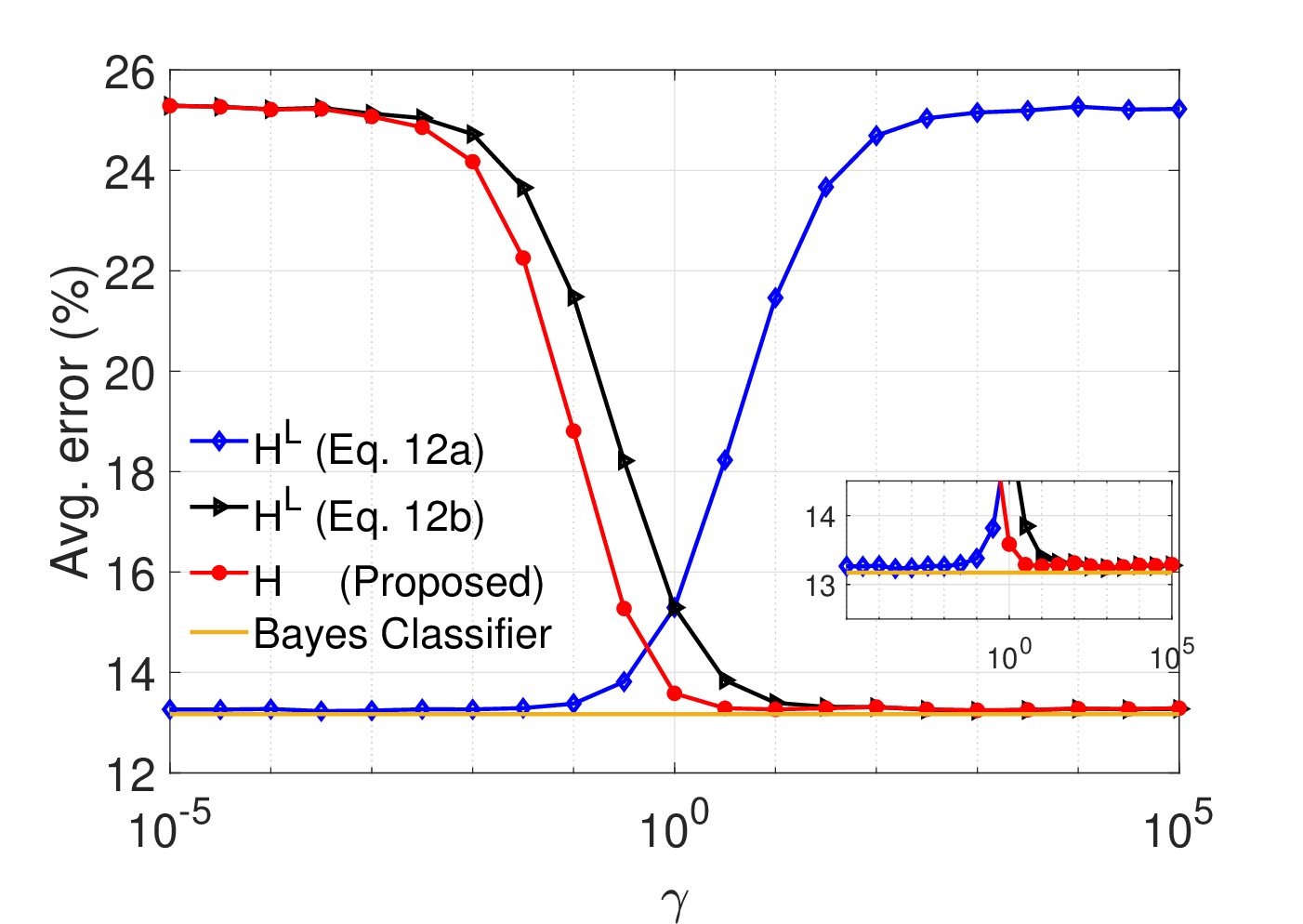}%
	\label{fig:fig1f}}
\\
\subfloat[$\nu^2=9,  n=50$]{\includegraphics[width=0.28\linewidth]{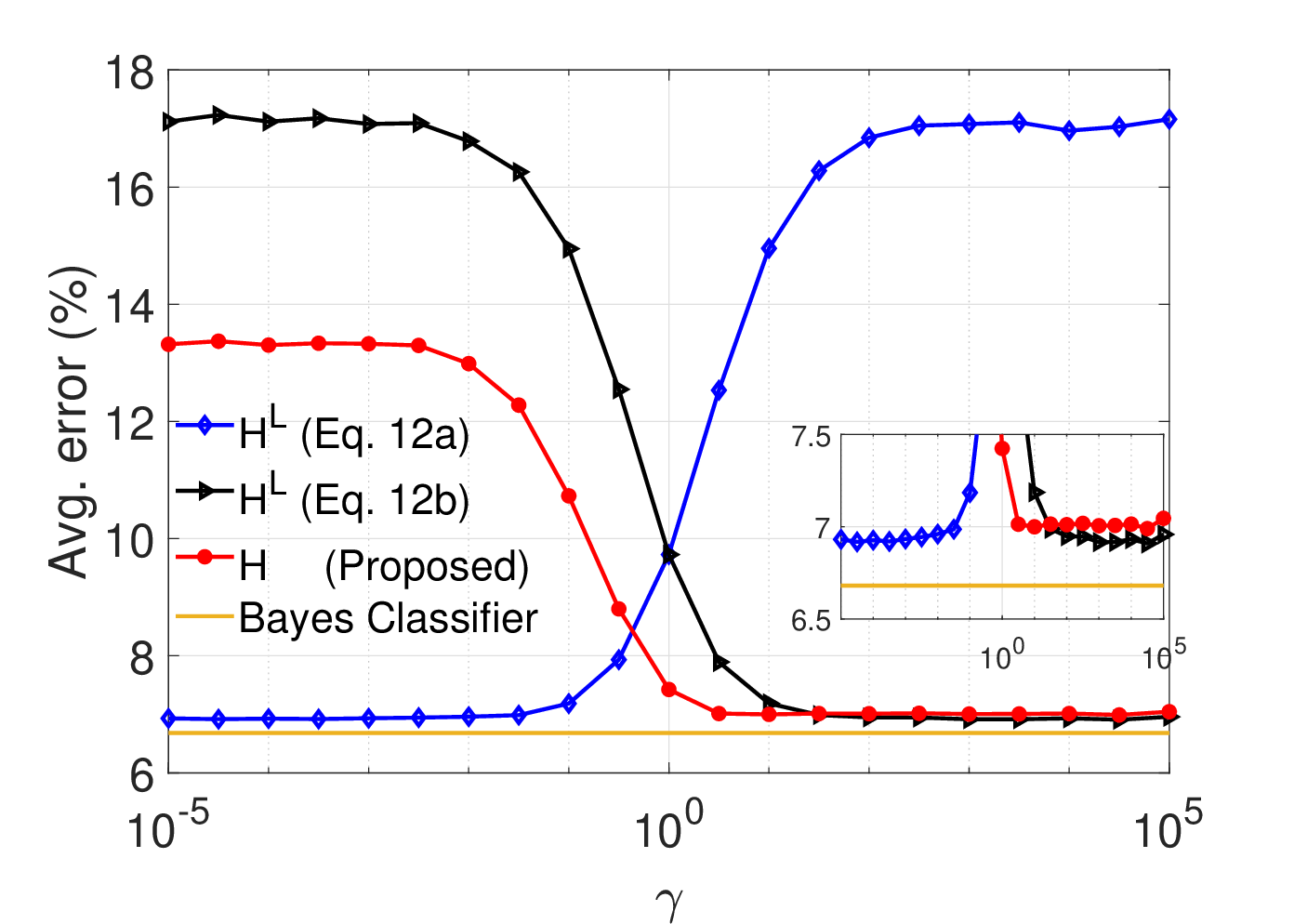}%
	\label{fig:fig1g}}
\hfil
\subfloat[$\nu^2=9, n=100$]{\includegraphics[width=0.28\linewidth]{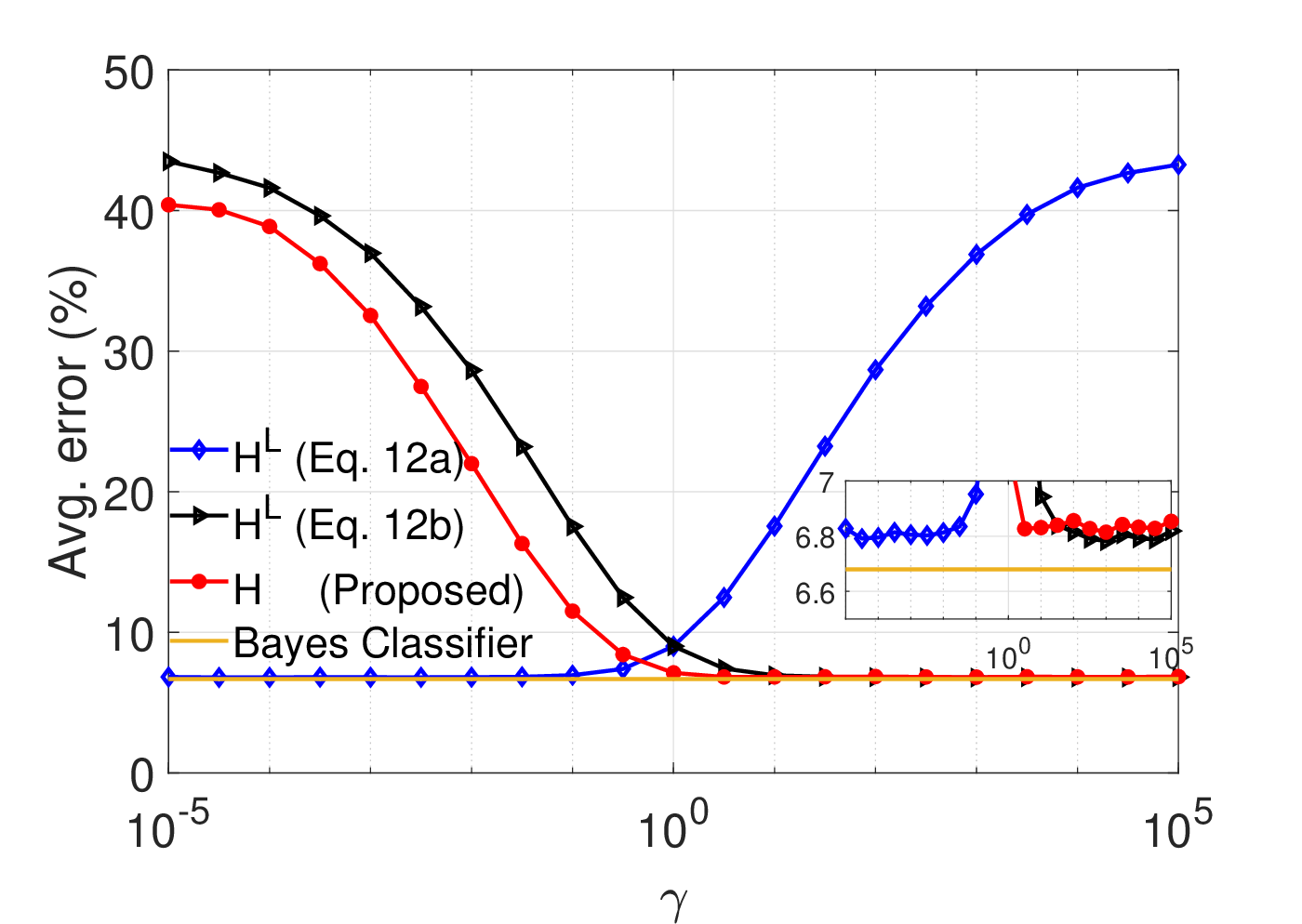}%
	\label{fig:fig1h}}
\hfil
\subfloat[$\nu^2=9, n=200$]{\includegraphics[width=0.28\linewidth]{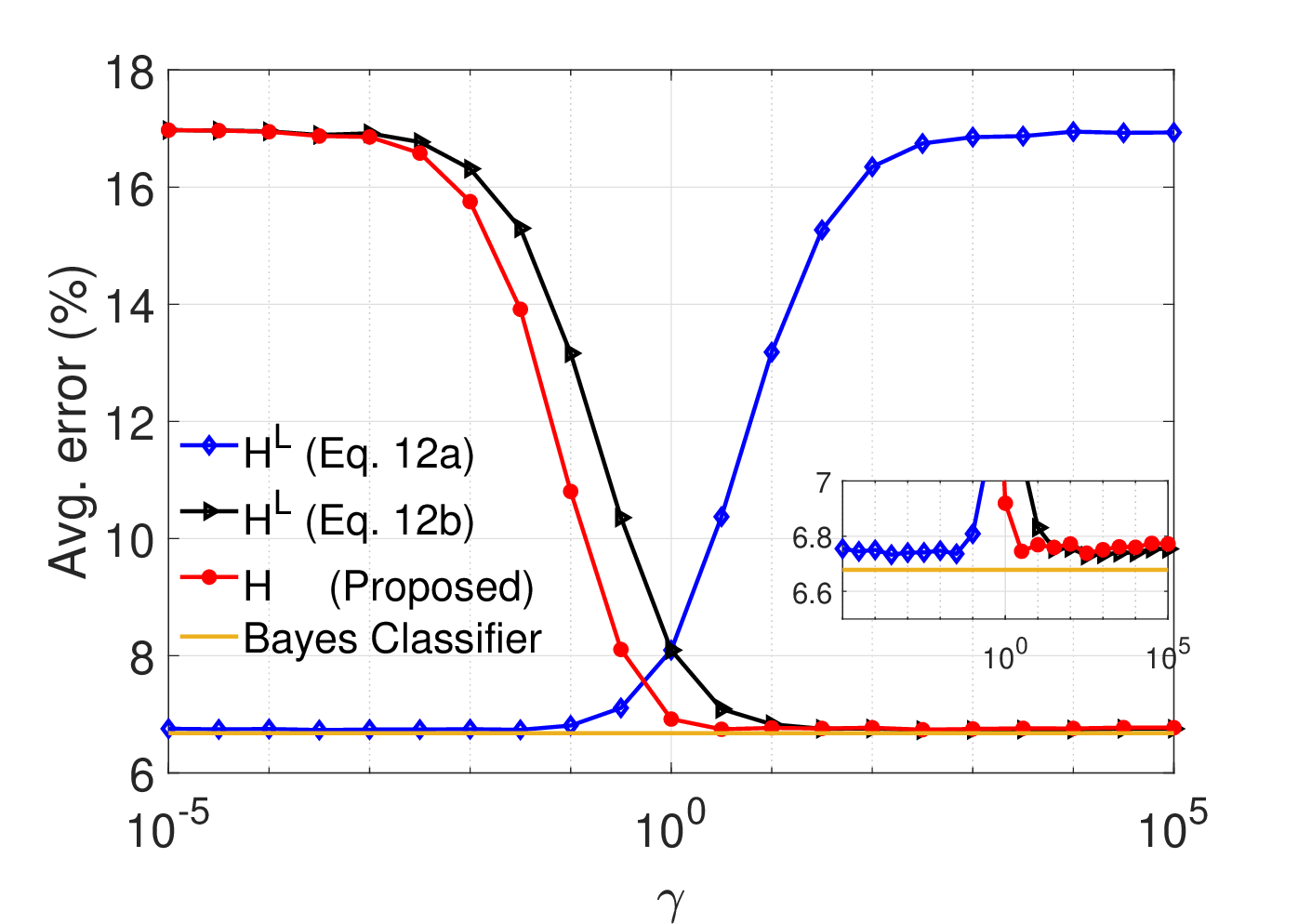}%
	\label{fig:fig1i}}
\\
\caption{RLDA classification error versus the regularization parameter ($\gamma$) for Gaussian data: A comparison of the proposed nonlinear covariance matrix estimator and commonly used linear estimators for different Mahalanobis distance ($\nu$) and training data size ($n$) values and a fixed $p=100$.}
\label{fig:fig1}	
\end{figure*}

\begin{figure*}[!t]
\centering
\subfloat[$\nu^2=0.5,  n=50$]{\includegraphics[width=0.25\linewidth]{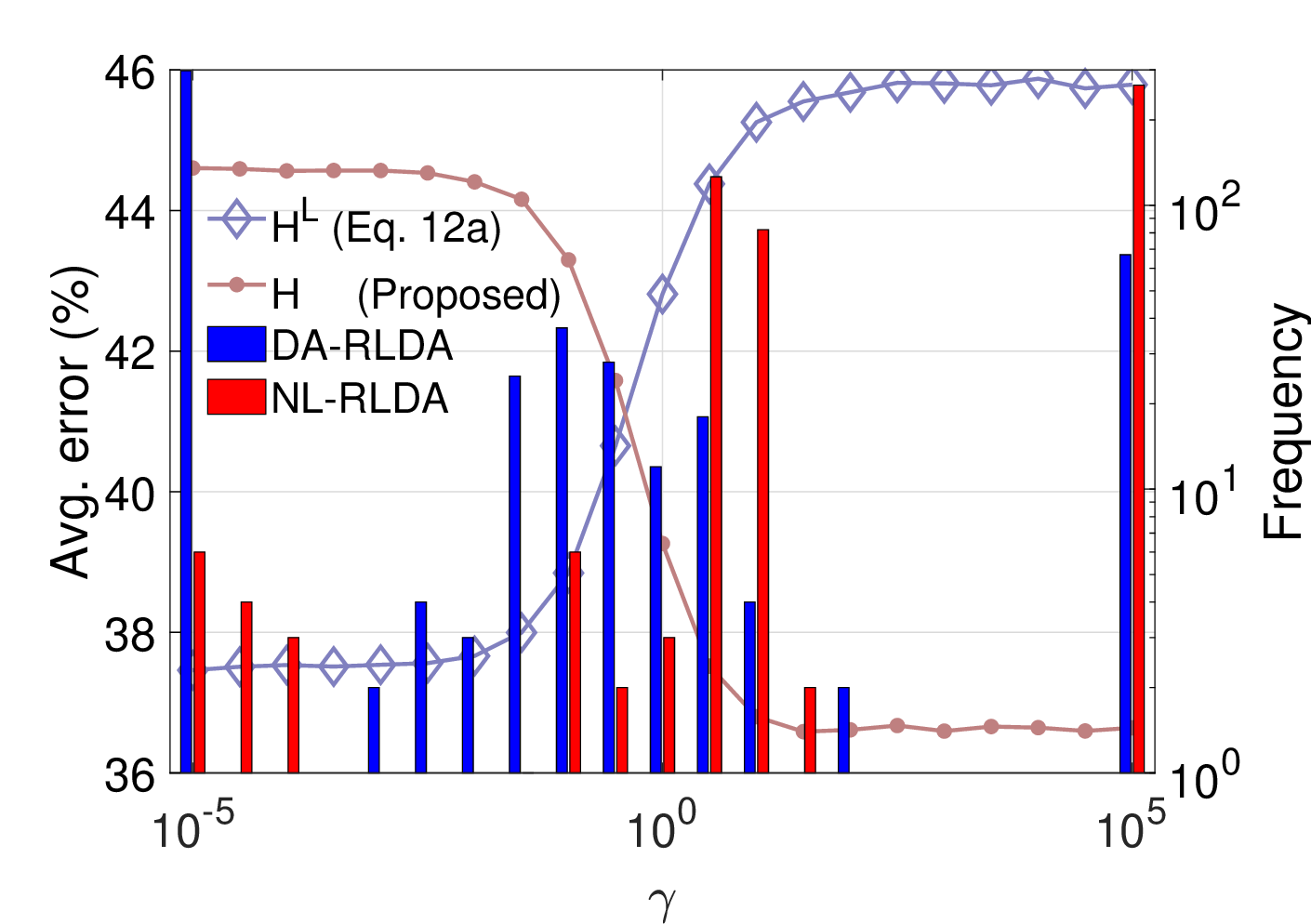}%
	\label{fig:p5bar50}}
\hfil
\subfloat[$\nu^2=0.5,  n=100$]{\includegraphics[width=0.25\linewidth]{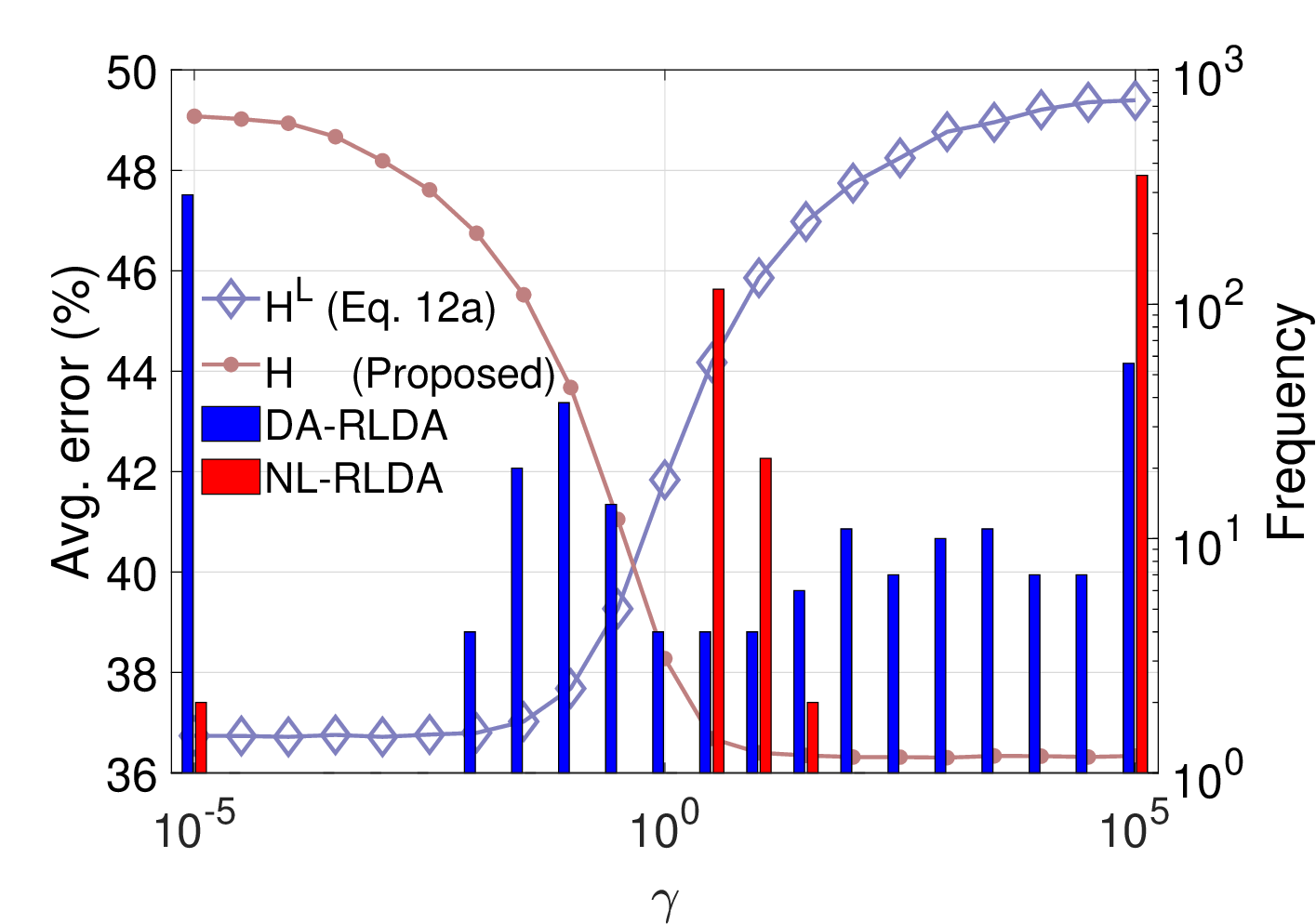}%
	\label{fig:p5bar100}}
\hfil
\subfloat[$\nu^2=0.5, n=200$]{\includegraphics[width=0.25\linewidth]{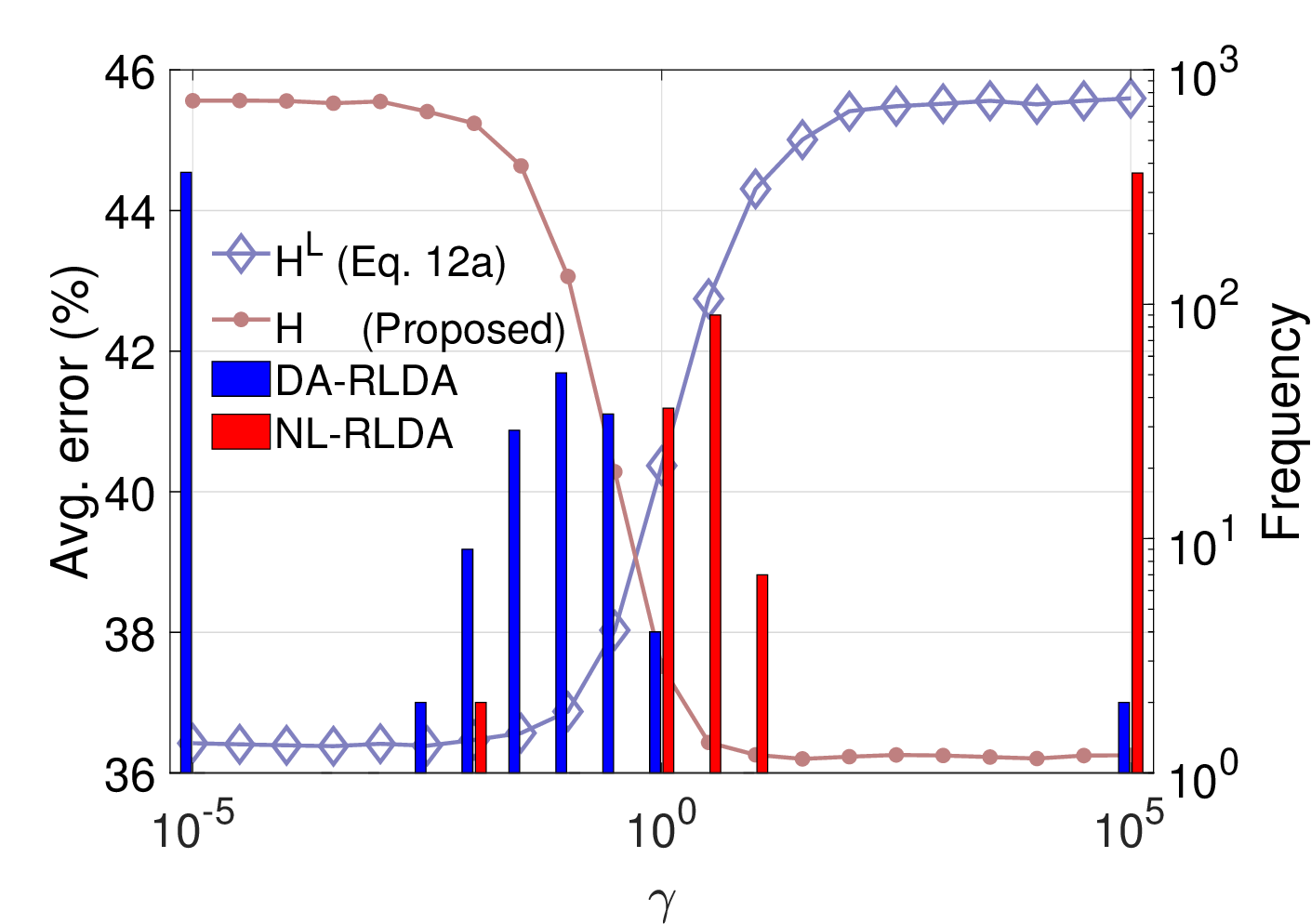}%
	\label{fig:p5bar200}}
\\
\subfloat[$\nu^2=5, n=50$]{\includegraphics[width=0.25\linewidth]{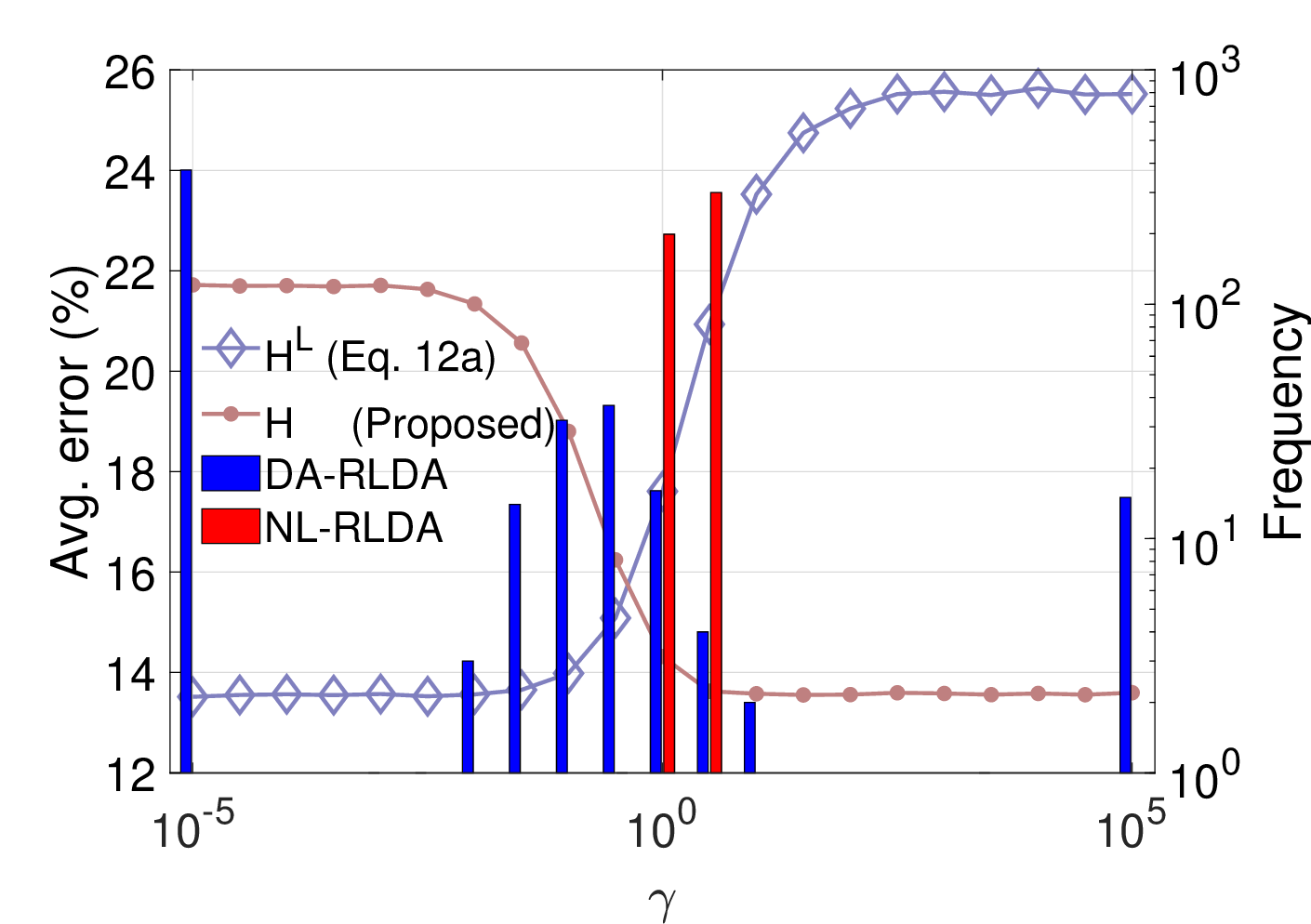}%
	\label{fig:5bar50}}
\hfil
\subfloat[$\nu^2=5, n=100$]{\includegraphics[width=0.25\linewidth]{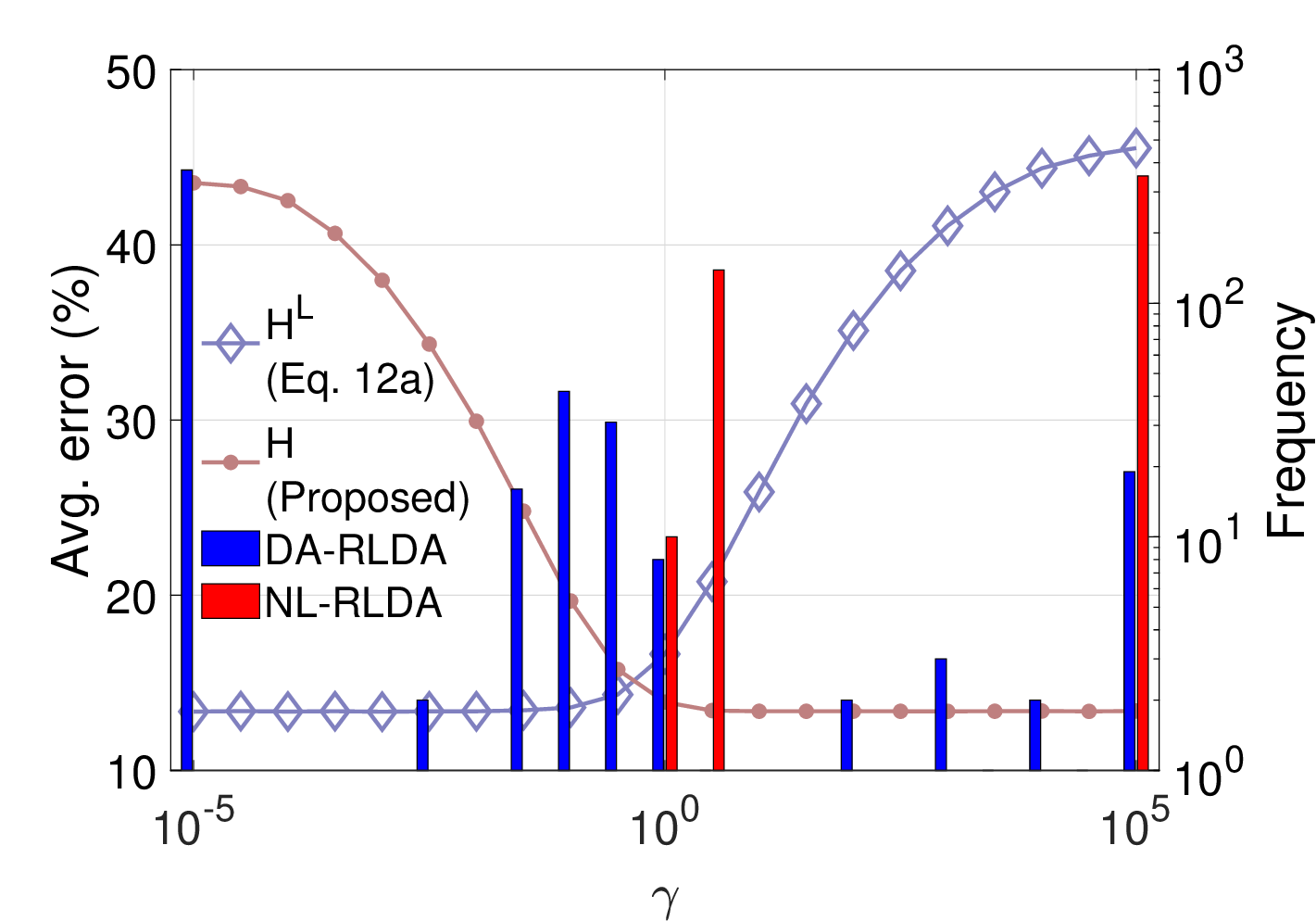}%
	\label{fig:5bar100}}
\hfil	
\subfloat[$\nu^2=5, n=200$]{\includegraphics[width=0.25\linewidth]{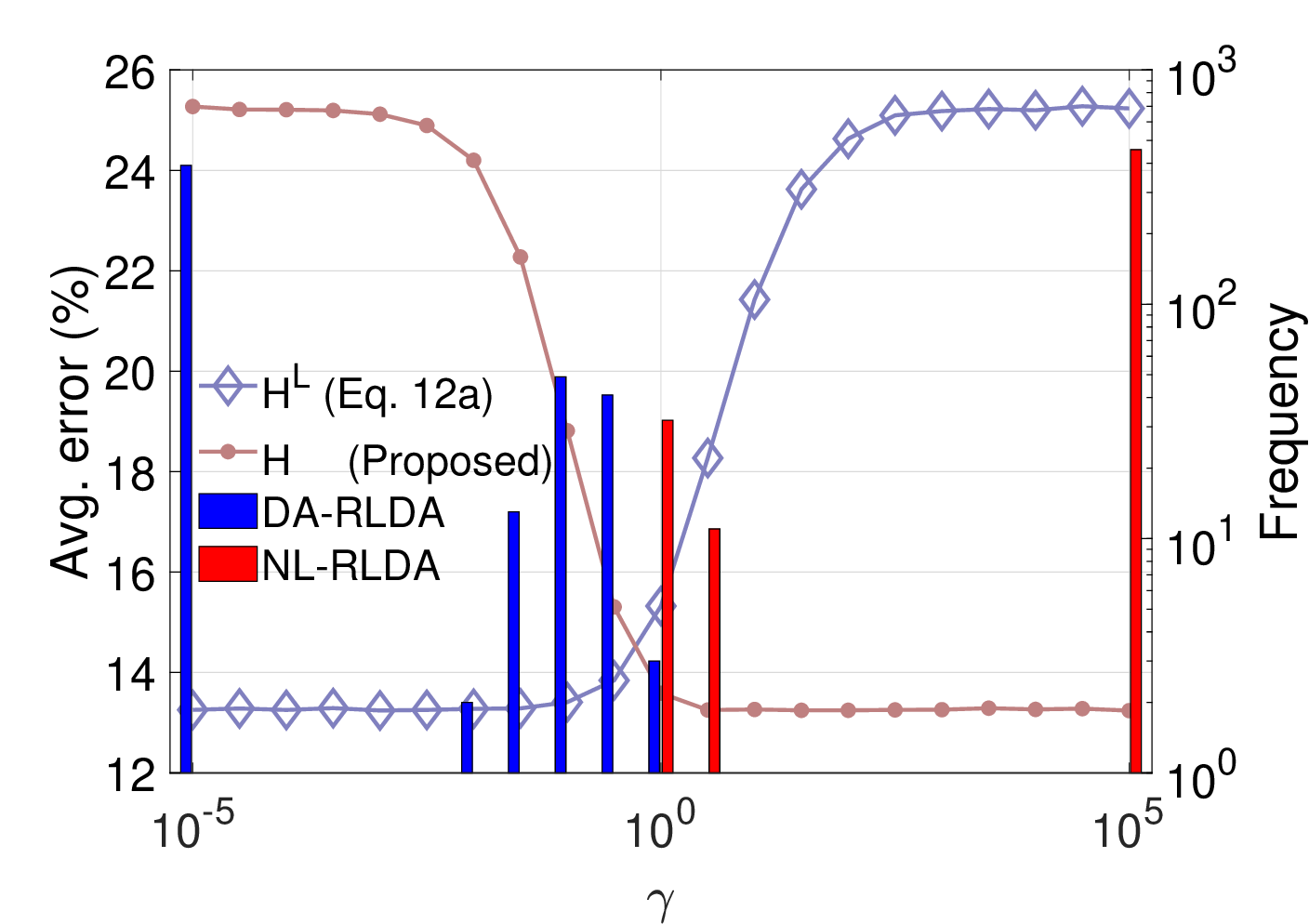}%
	\label{fig:5_bar200}}
\\
\subfloat[$\nu^2=9,  n=50$]{\includegraphics[width=0.28\linewidth]{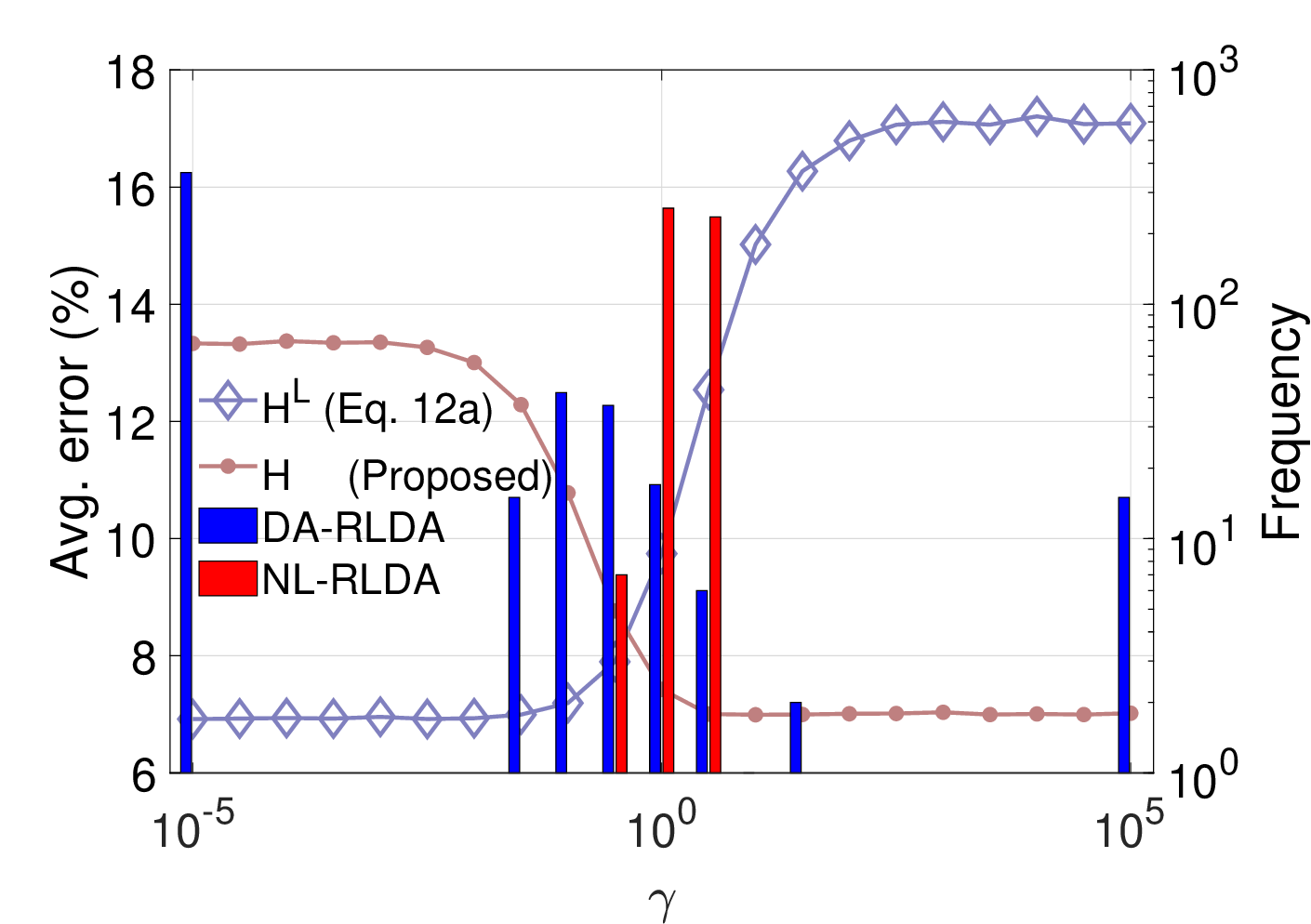}%
	\label{fig:9_bar50}}
\hfil
\subfloat[$\nu^2=9, n=100$]{\includegraphics[width=0.25\linewidth]{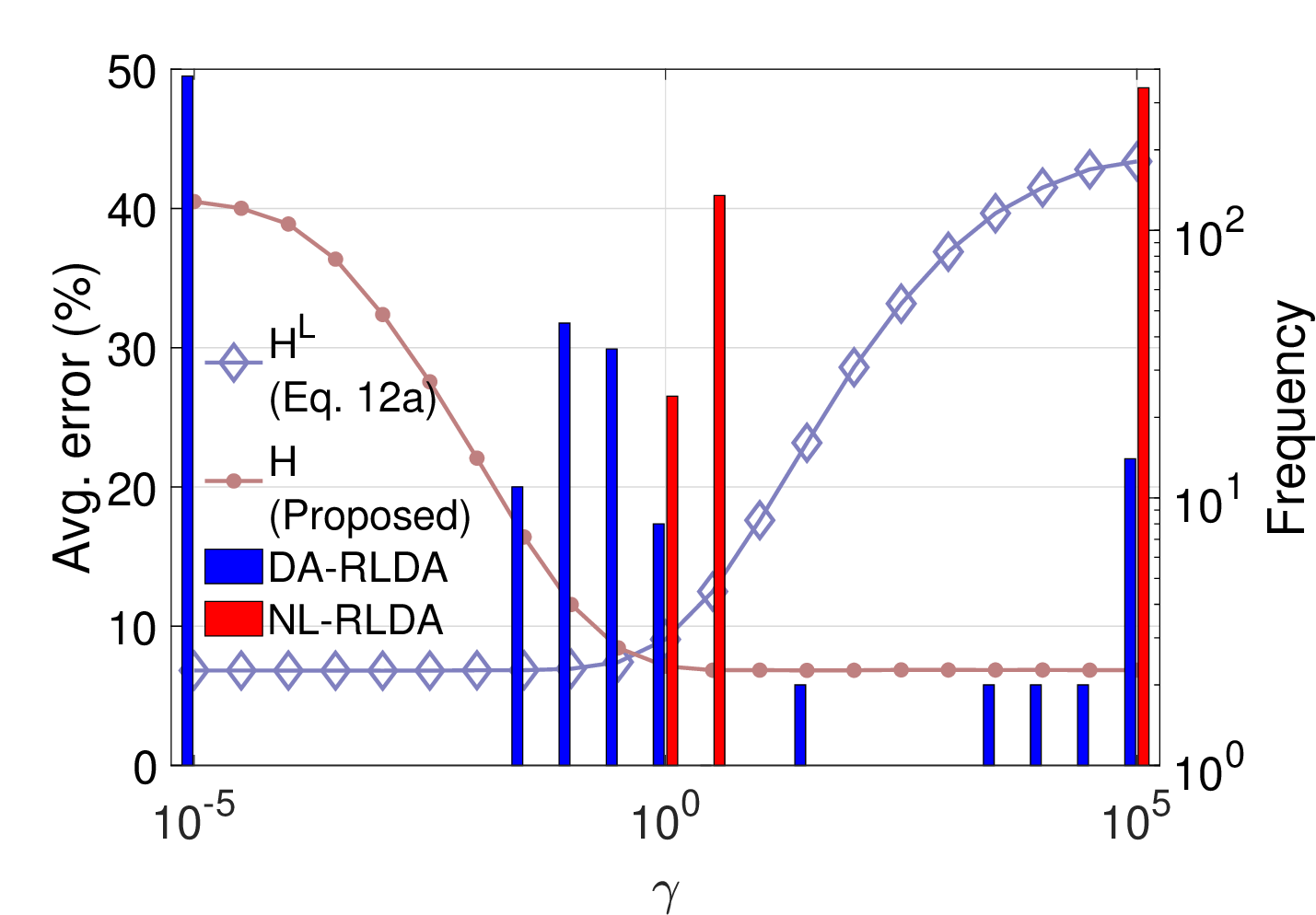}%
	\label{fig:9bar100}}
\hfil
\subfloat[$\nu^2=9, n=200$]{\includegraphics[width=0.25\linewidth]{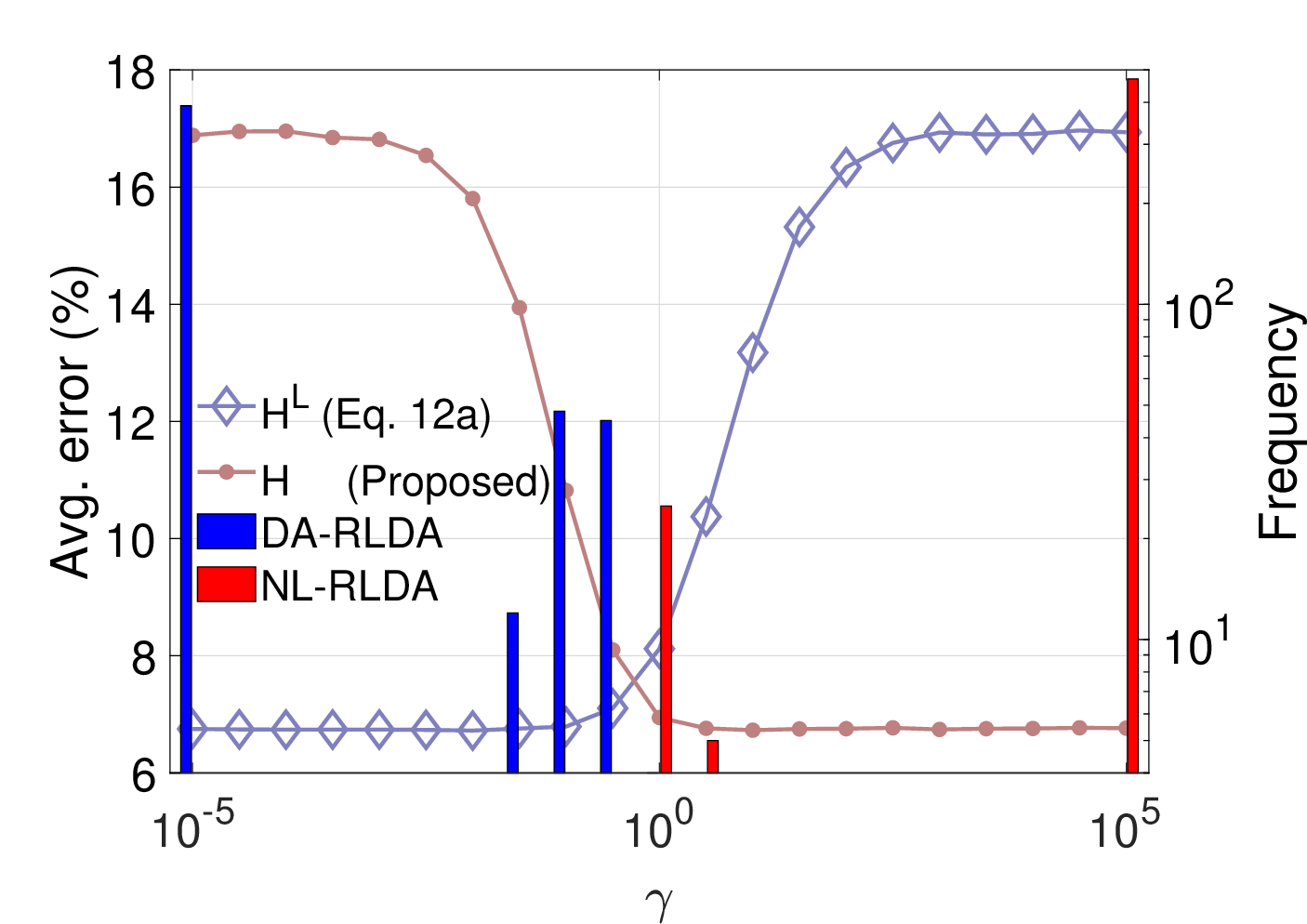}%
	\label{fig:9_bar200}}
\\
\caption{Histograms of the regularization parameter ($\gamma$) values together with RLDA classification error versus $\gamma$ curves for Gaussian data: A comparison of the proposed nonlinear covariance matrix estimator and commonly used linear estimators for different Mahalanobis distance ($\nu$) and training data size ($n$) values and a fixed $p=100$. }
\label{fig:barplot}	
\end{figure*}

\begin{figure*}
    \centering
    {\includegraphics[width=10cm, height=1cm]{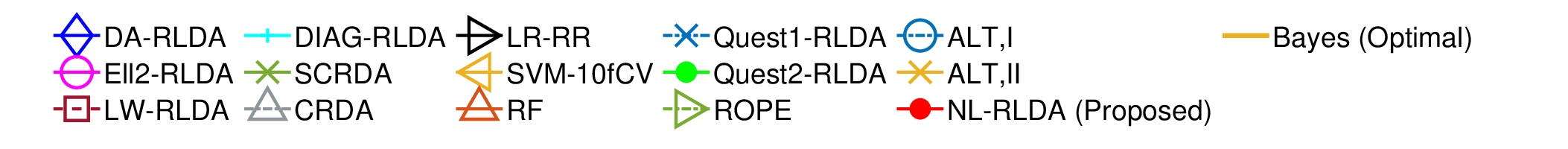} }\\
    \setkeys{Gin}{width=0.25\linewidth} 
    \subfloat[$\pi_0 = 0.5$, $p=50$.      \label{subfig:M1_50a}]{\includegraphics{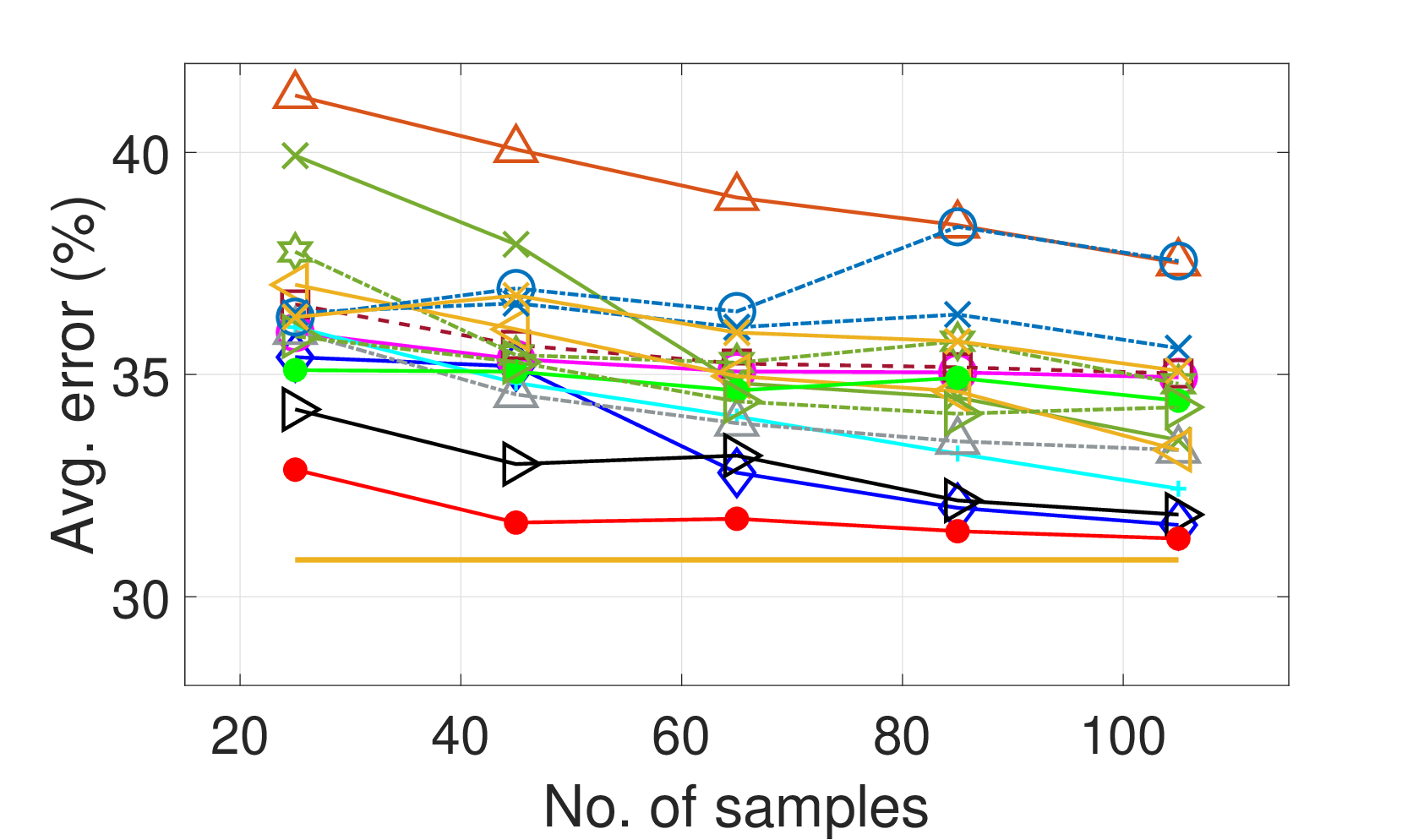} }\hfil
    \subfloat[$\pi_0 = 0.5$, $p=250$.    \label{subfig:M1_250a}]{\includegraphics{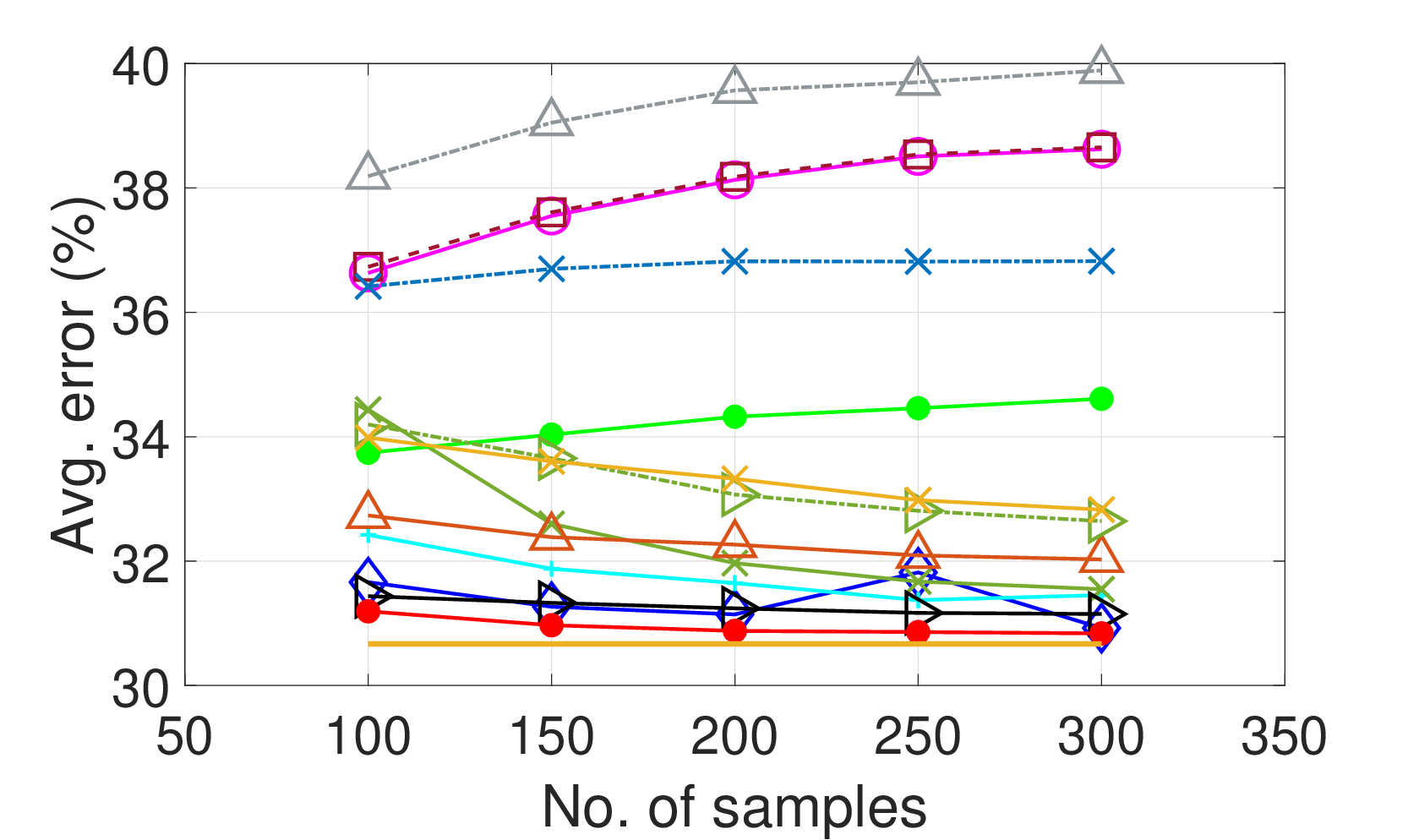} }\hfil
    \subfloat[$\pi_0 = 0.5$, $p=1000$.  \label{subfig:M1_1000a}]{\includegraphics{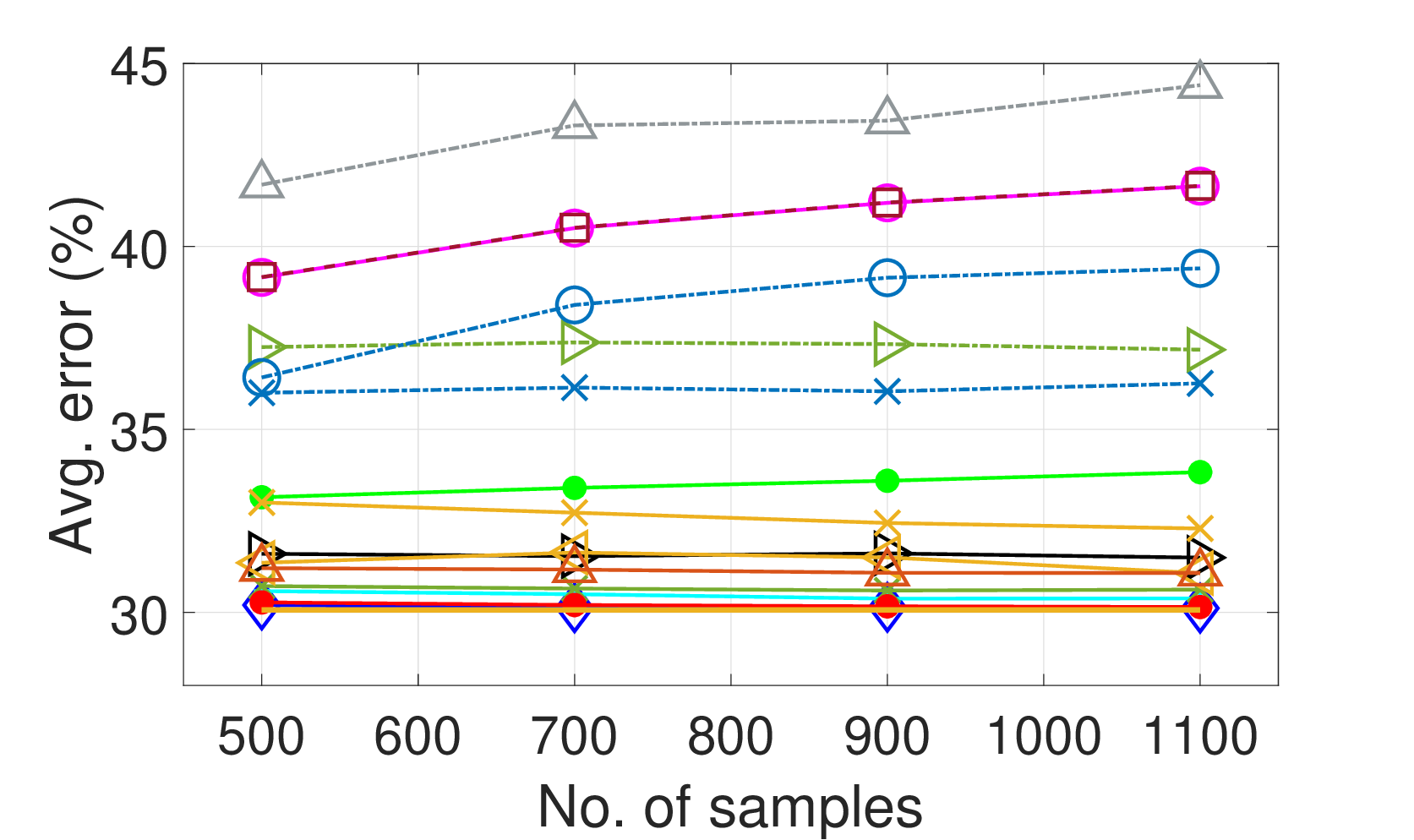} }\hfil
    \subfloat[$\pi_0 = 0.3$, $p=50$.\label{subfig:M1_50b}]{\includegraphics{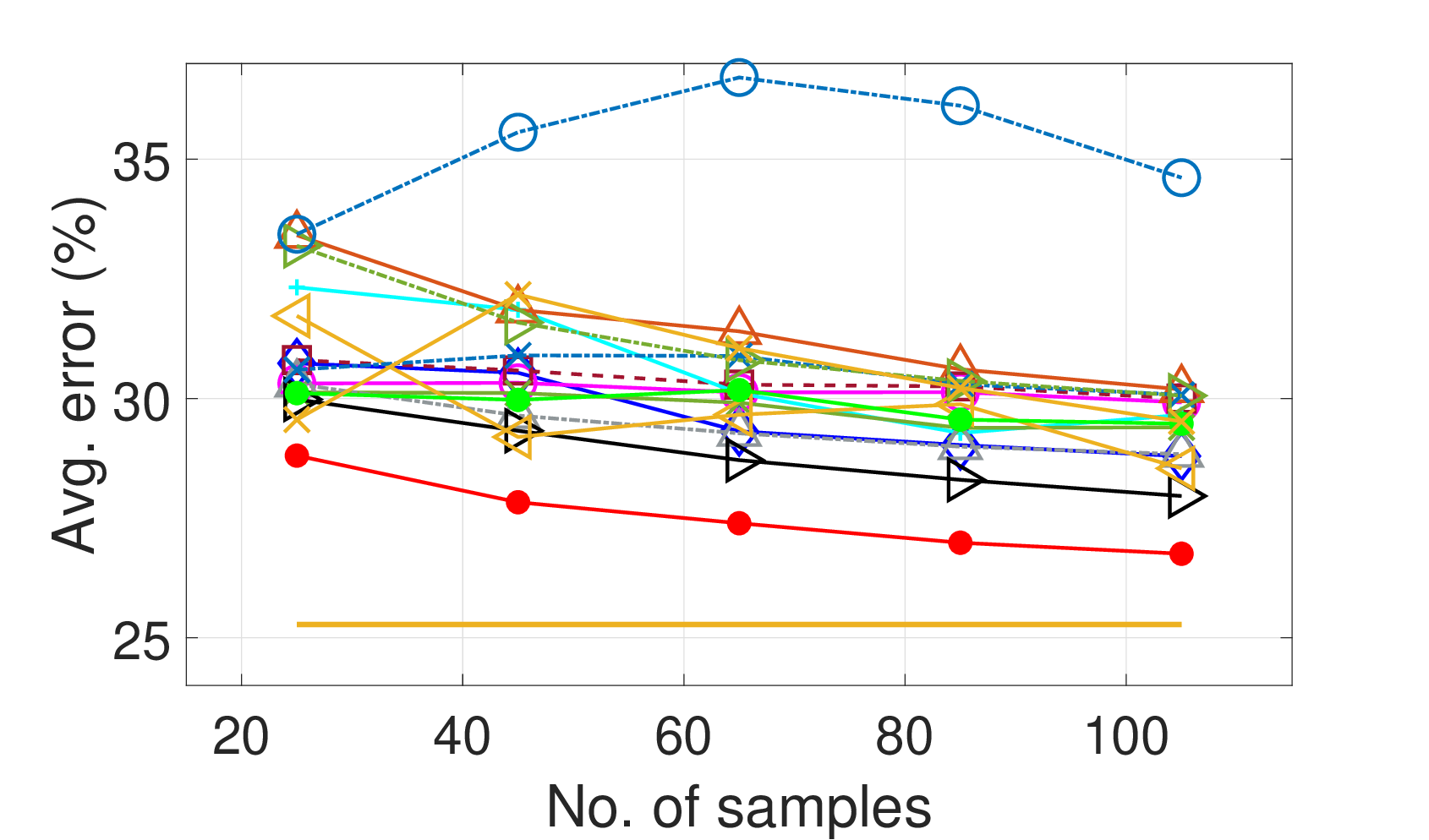} }\hfil
    \subfloat[$\pi_0 = 0.3$, $p=250$. \label{subfig:M1_250b}]{\includegraphics{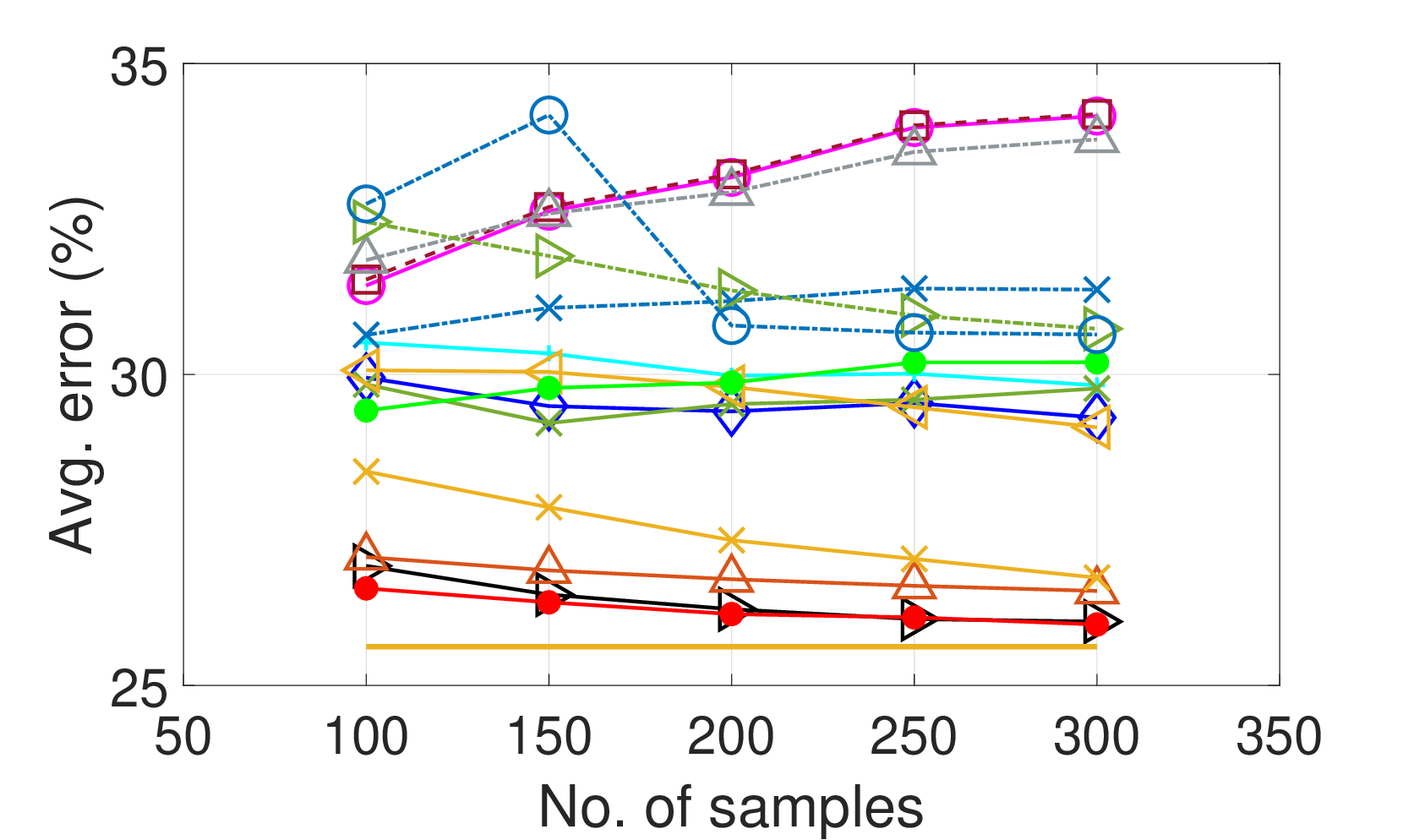} }\hfil
    \subfloat[$\pi_0 = 0.3$, $p=1000$. \label{subfig:M1_1000b}]{\includegraphics{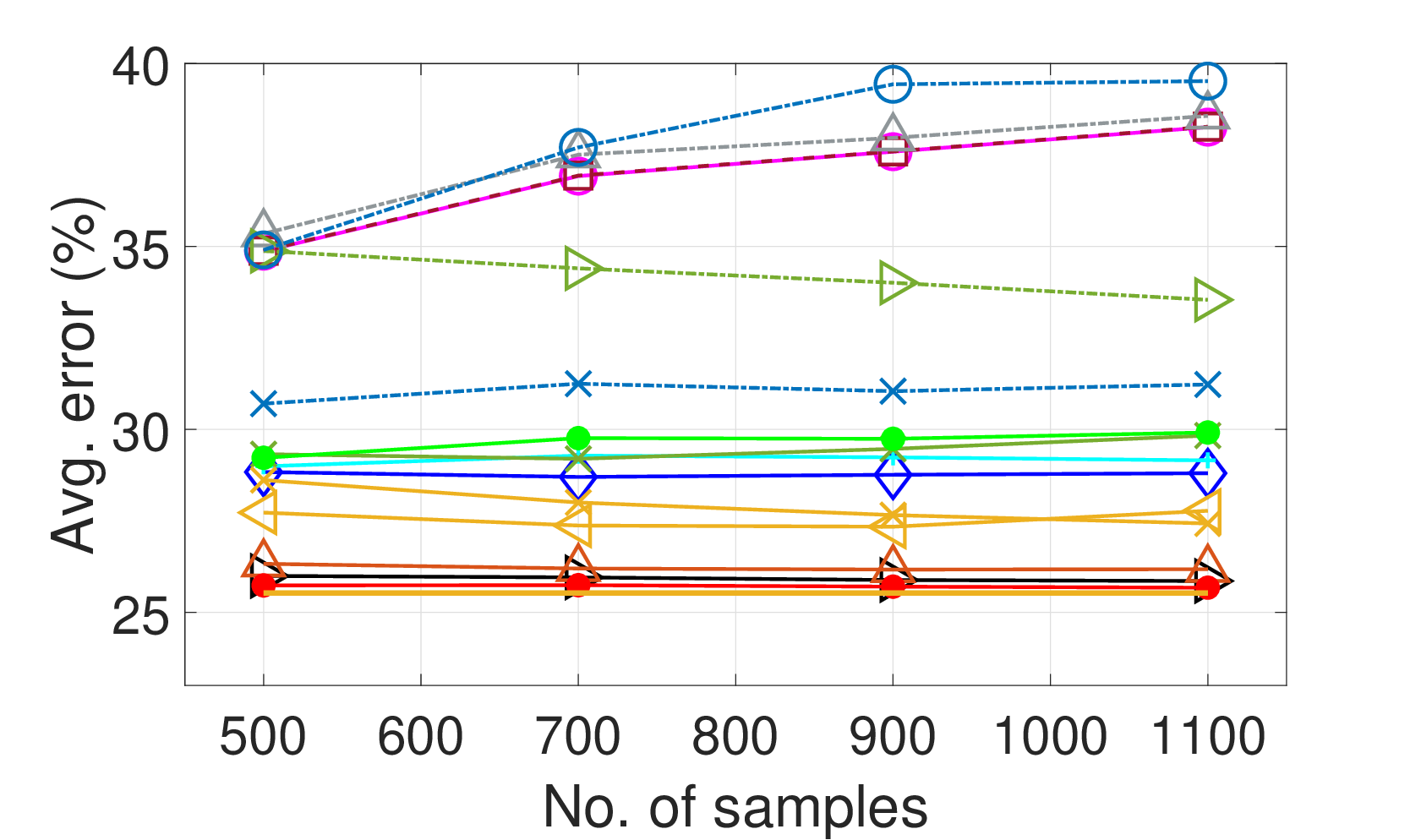} }
    
    \caption{Model 1:  The diagonal elements of the covariance are equal to 1, and the off-diagonal elements are  0.1.}
    \label{fig:Model 1}
\bigskip
    \subfloat[$\pi_0 = 0.5$, $p=50$.      \label{subfig:M2_50a}]{\includegraphics{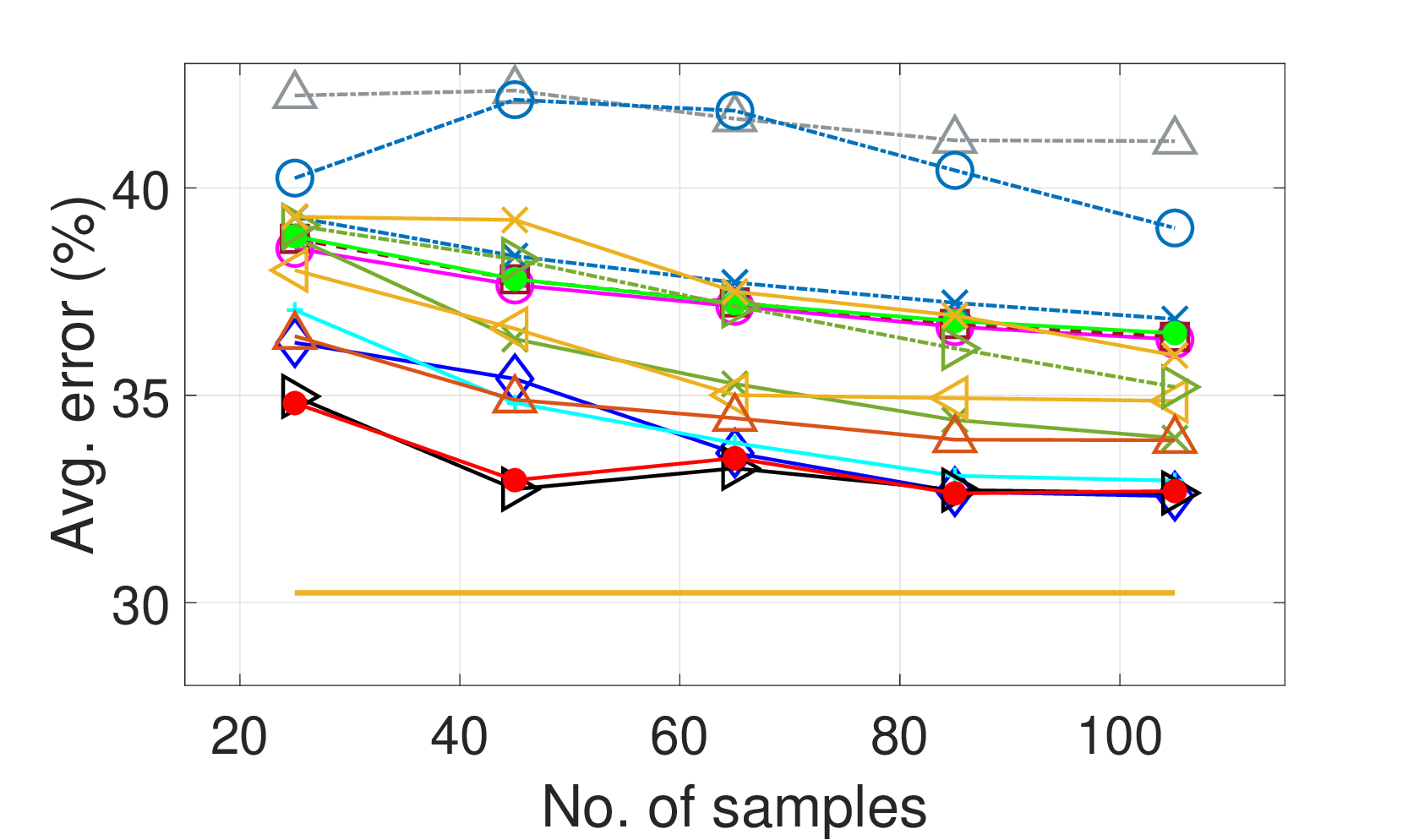} }\hfil
    \subfloat[$\pi_0 = 0.5$, $p=250$.    \label{subfig:M2_250a}]{\includegraphics{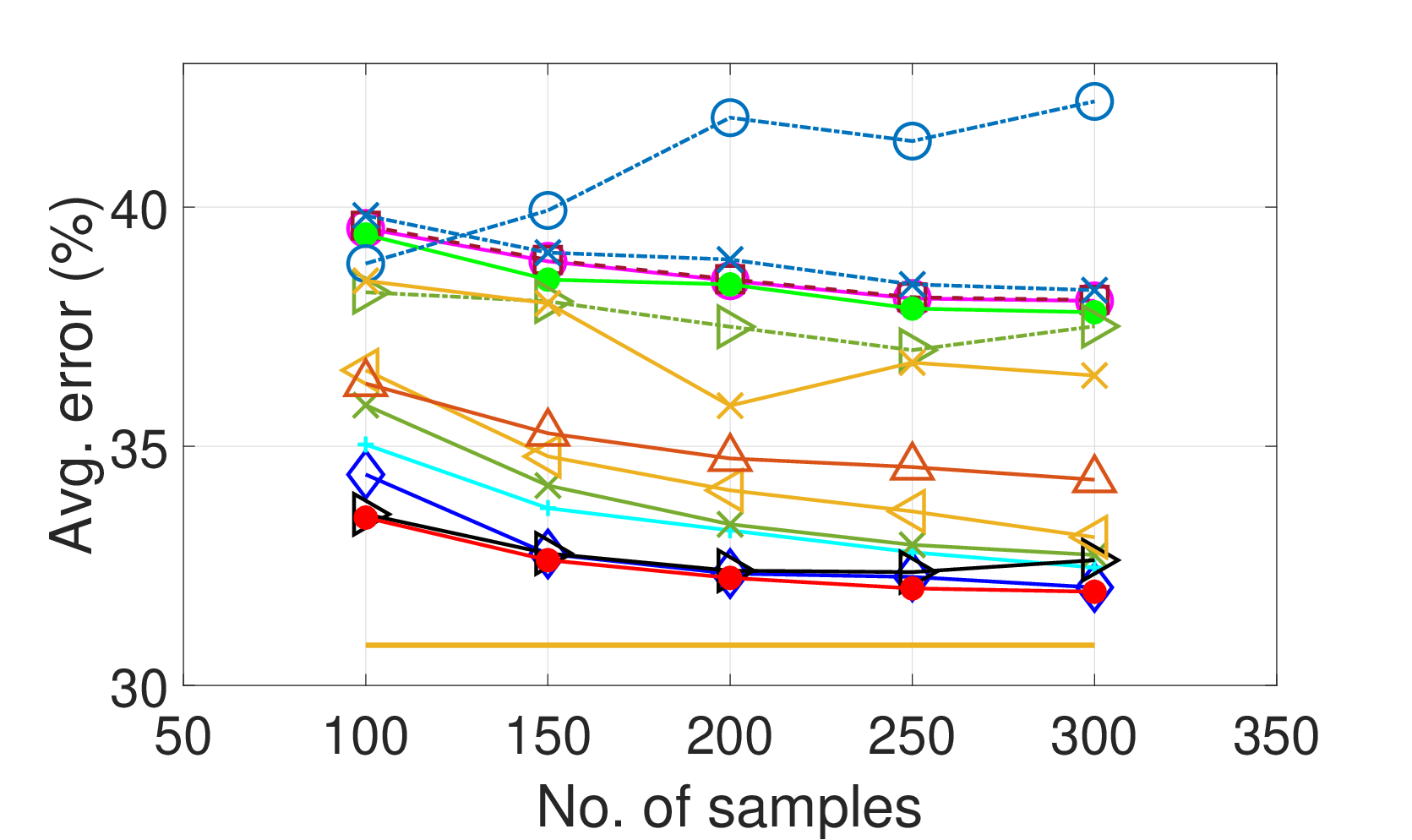} }\hfil
    \subfloat[$\pi_0 = 0.5$, $p=1000$.  \label{subfig:M2_1000a}]{\includegraphics{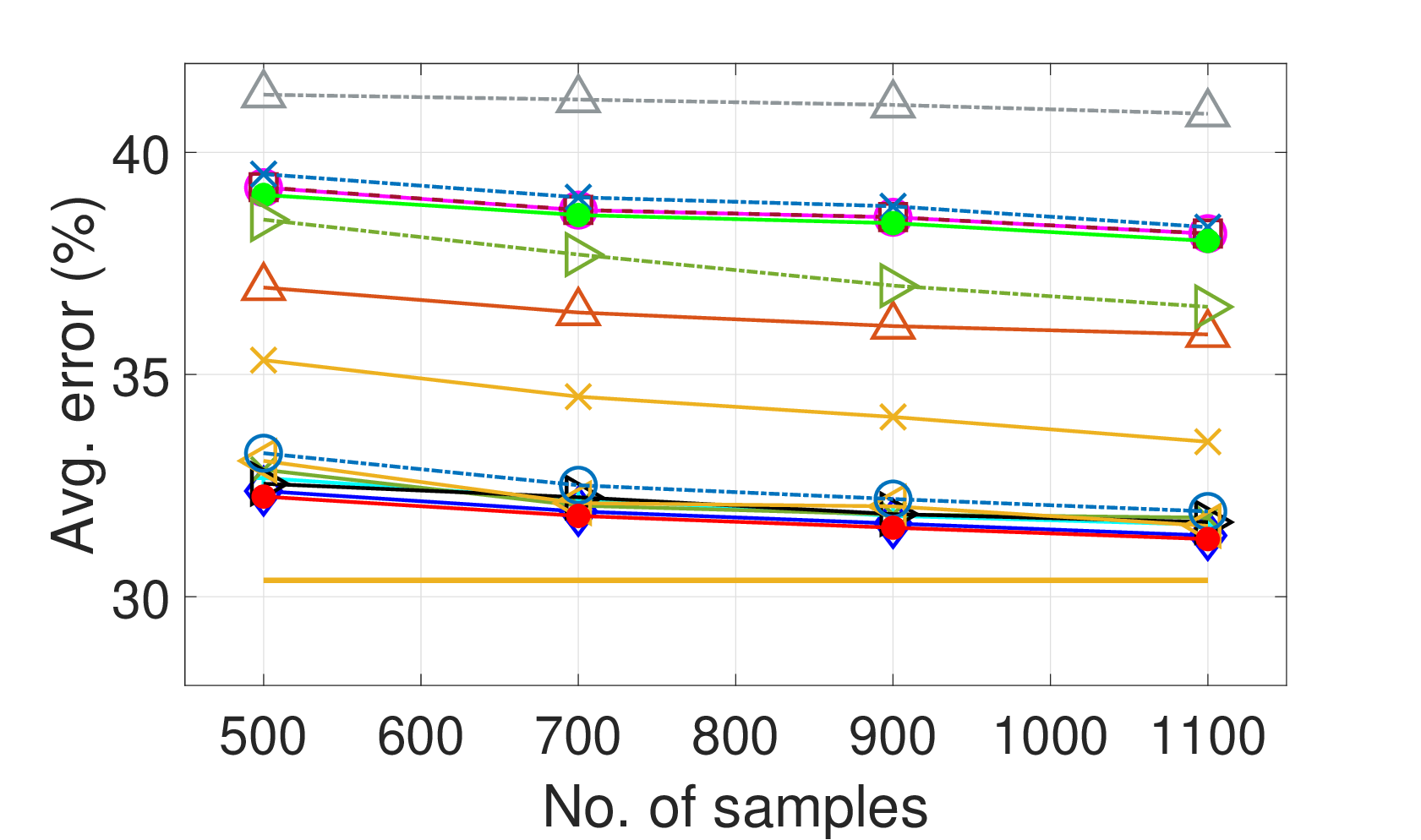} }\hfil
    \subfloat[$\pi_0 = 0.3$, $p=50$.\label{subfig:M2_50b}]{\includegraphics{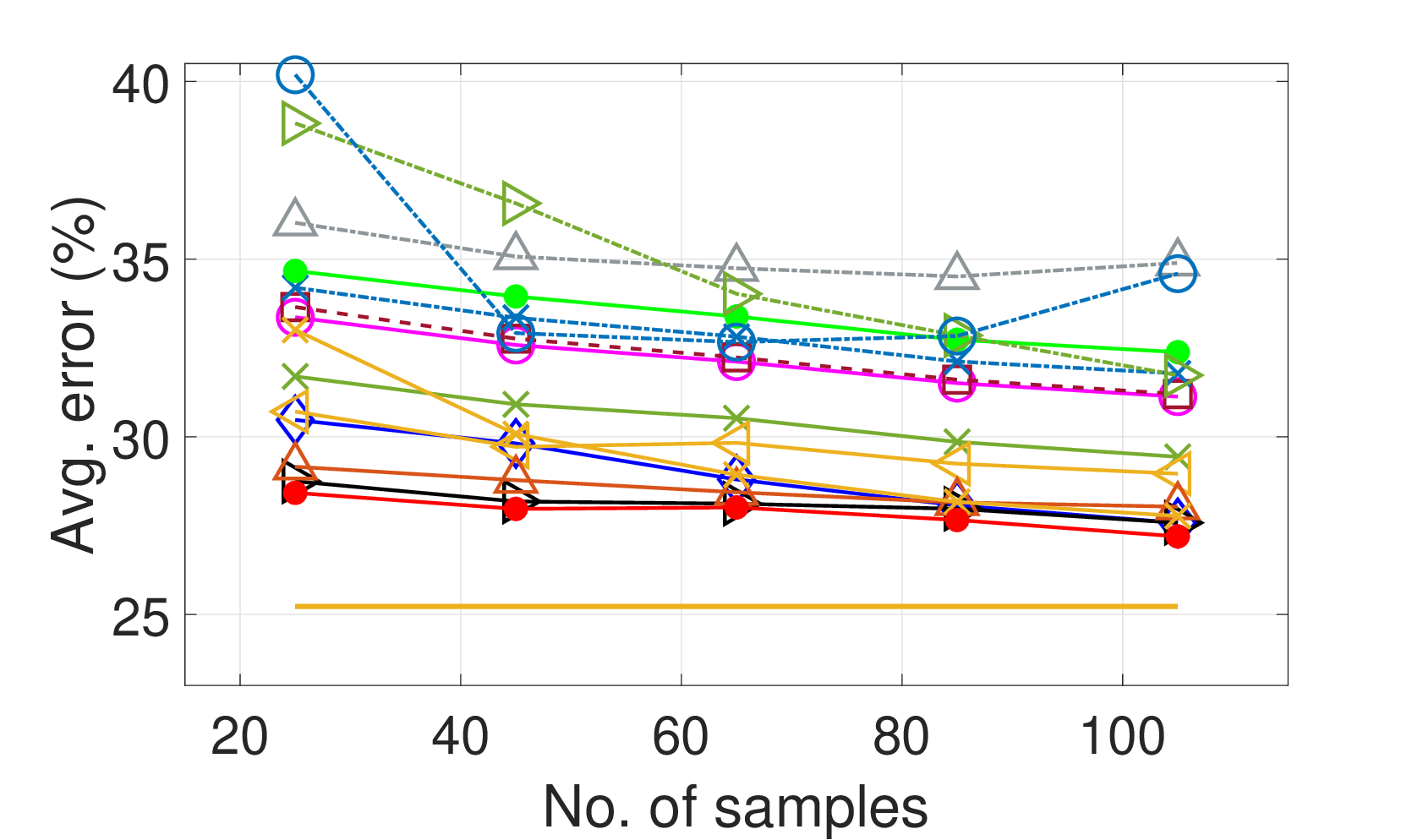} }\hfil
    \subfloat[$\pi_0 = 0.3$, $p=250$. \label{subfig:M2_250b}]{\includegraphics{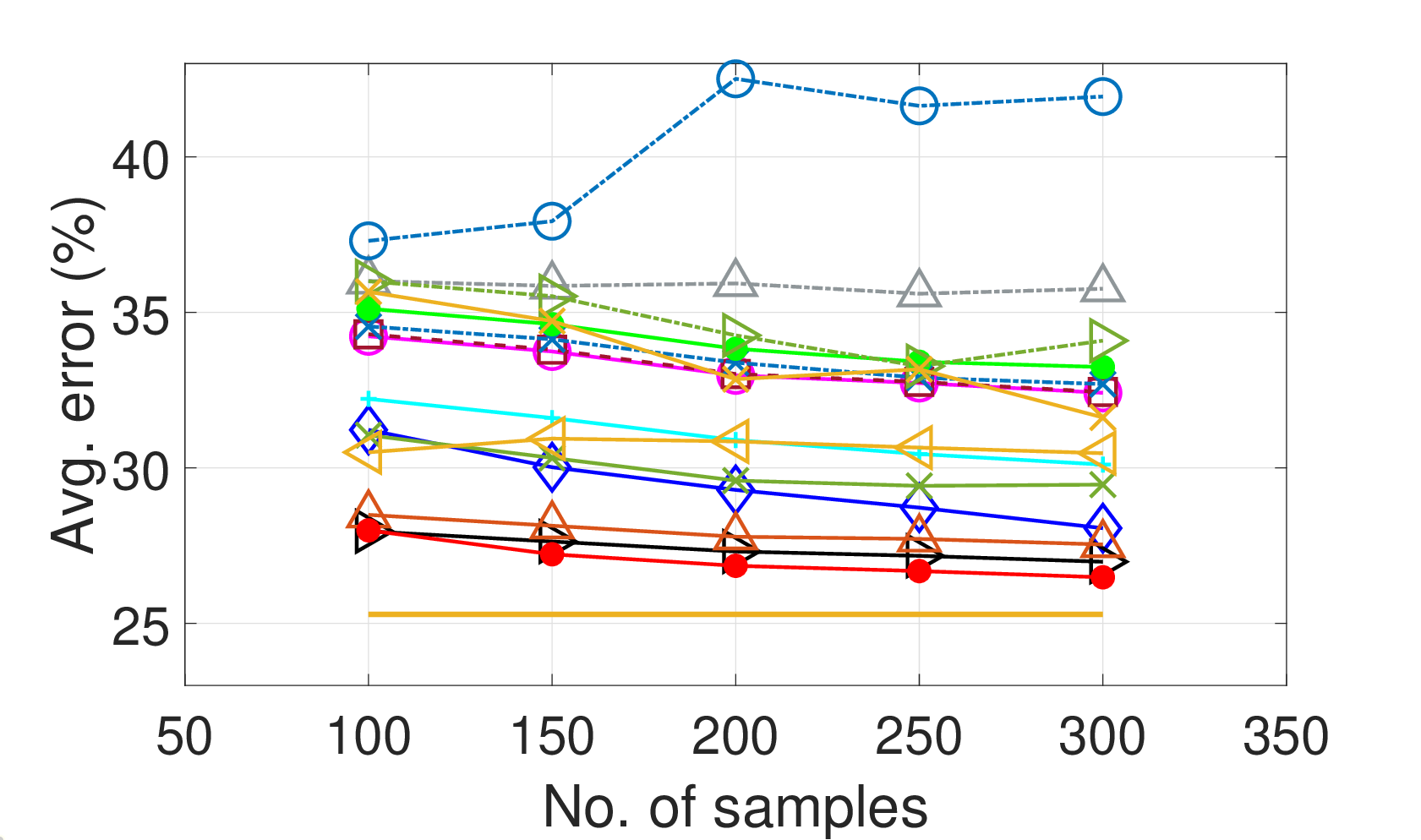} }\hfil
    \subfloat[$\pi_0 = 0.3$, $p=1000$. \label{subfig:M2_1000b}]{\includegraphics{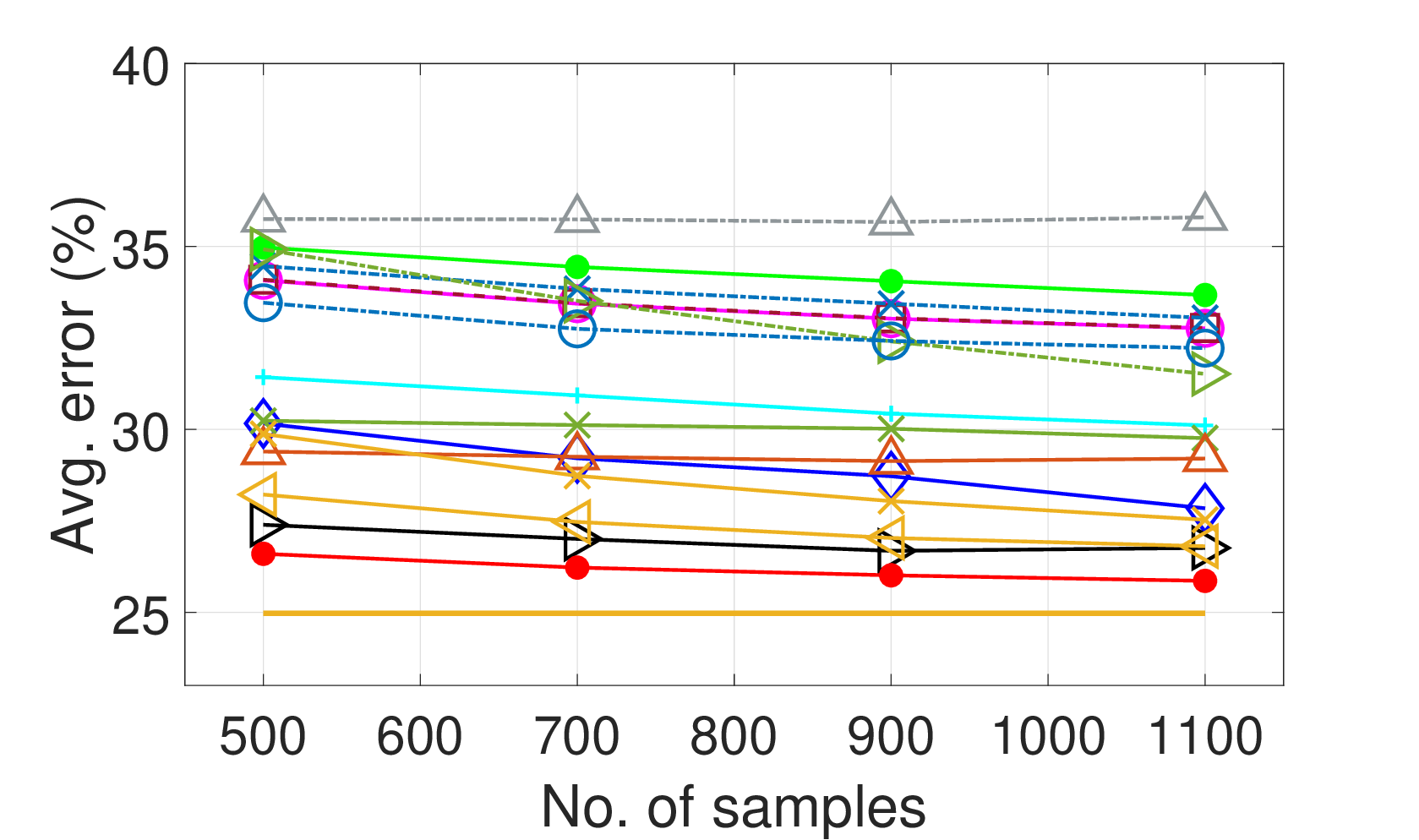} }
\caption{Model 2: The diagonal elements of the covariance are equal to 1, and the off-diagonal elements are $0.9^{|i-j|}$.}%
\label{fig:Model 2}
\bigskip
    \subfloat[$\pi_0 = 0.5$, $p=50$.      \label{subfig:M4_50a}]{\includegraphics{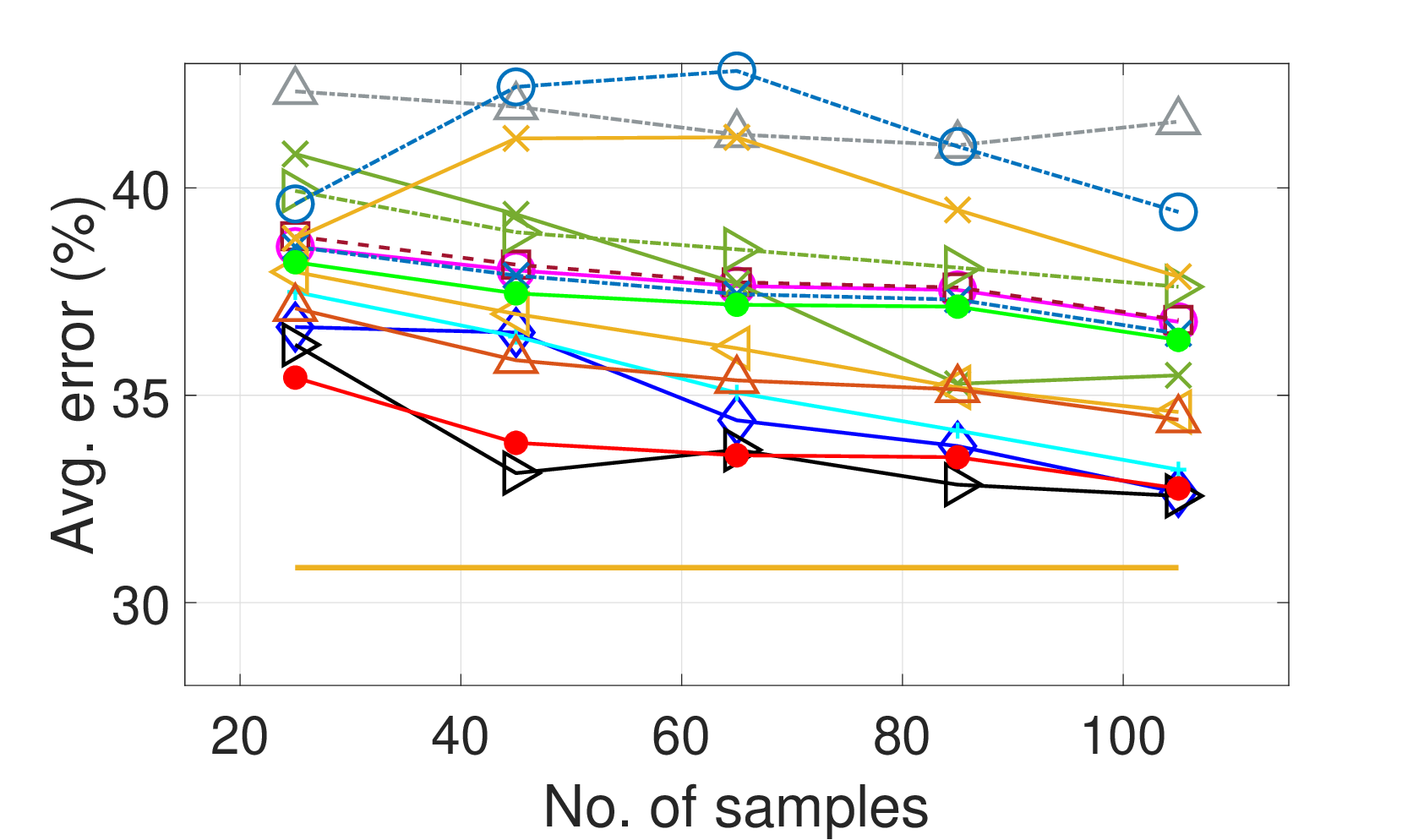} }\hfil
    \subfloat[$\pi_0 = 0.5$, $p=250$.    \label{subfig:M4_250a}]{\includegraphics{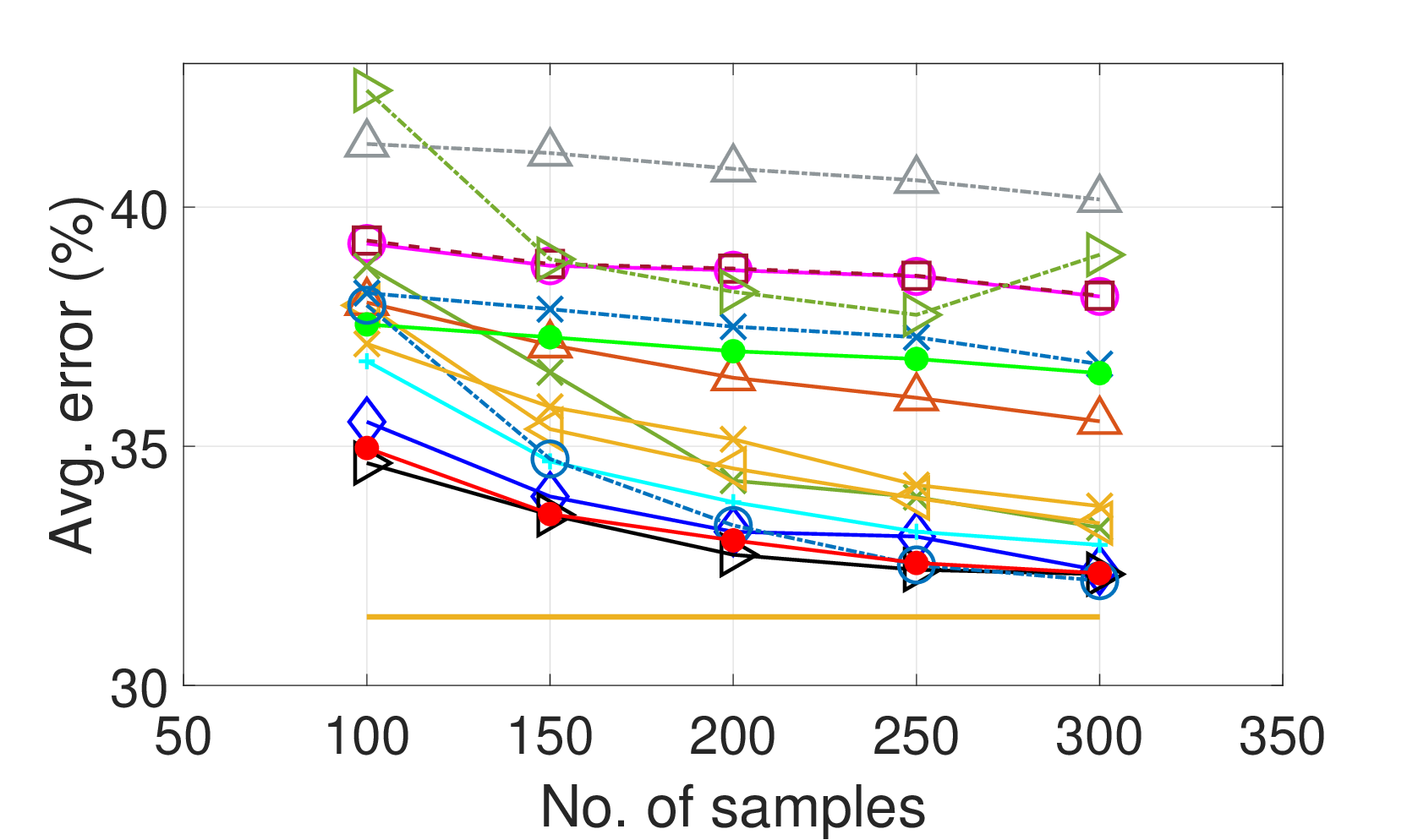} }\hfil
    \subfloat[$\pi_0 = 0.5$, $p=1000$.  \label{subfig:M4_1000a}]{\includegraphics{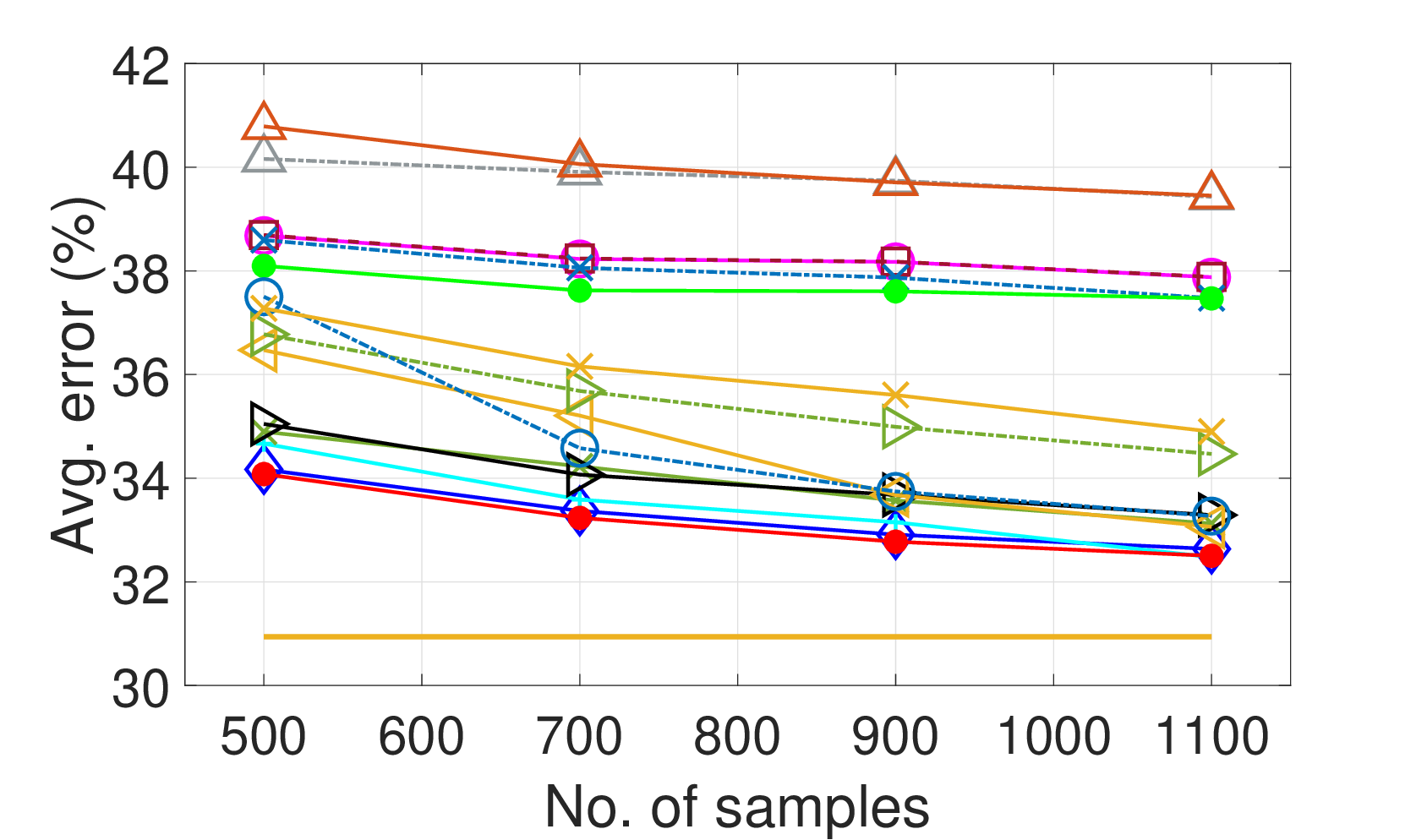} }\hfil
    \subfloat[$\pi_0 = 0.3$, $p=50$.\label{subfig:M4_50b}]{\includegraphics{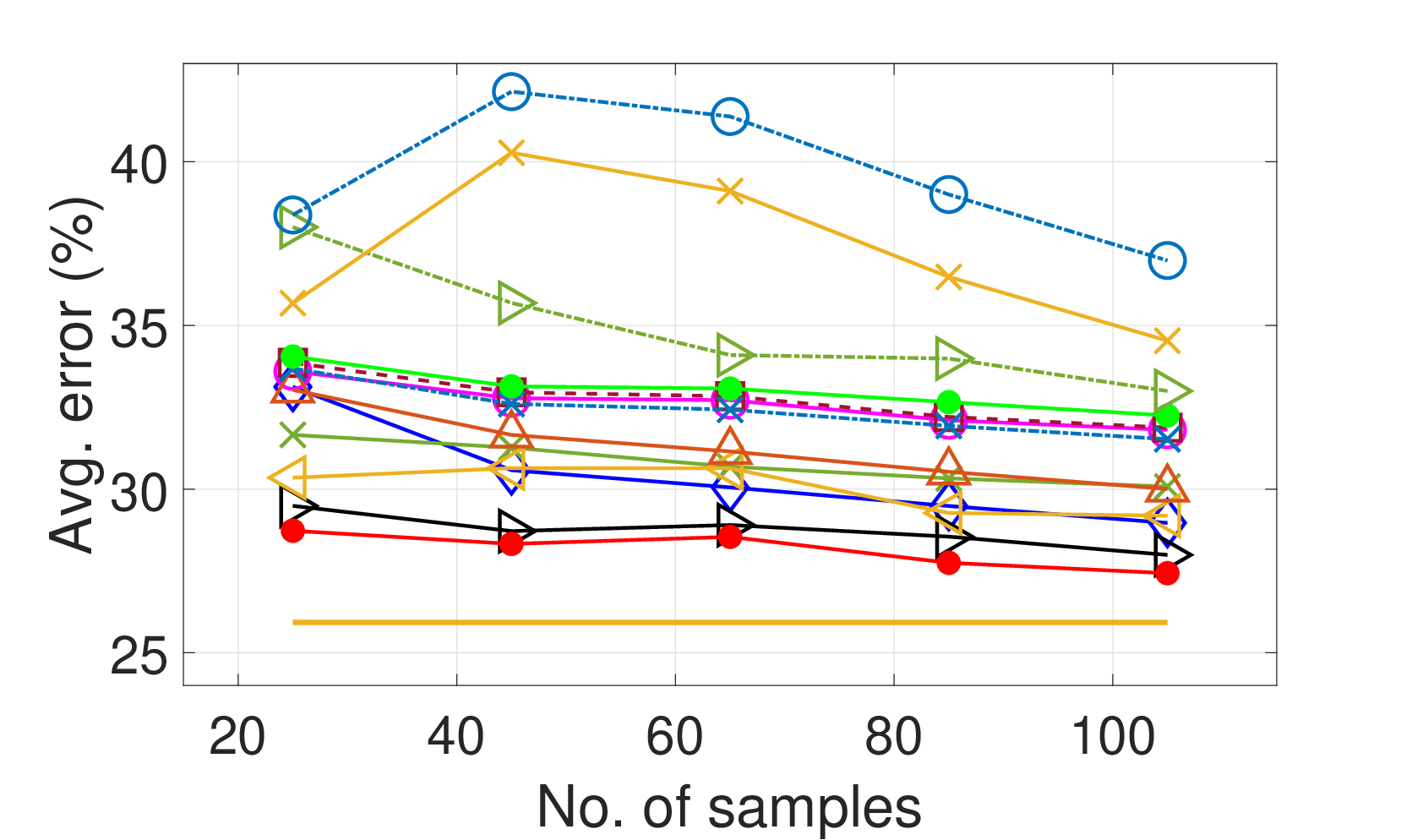} }\hfil
    \subfloat[$\pi_0 = 0.3$, $p=250$. \label{subfig:M4_250b}]{\includegraphics{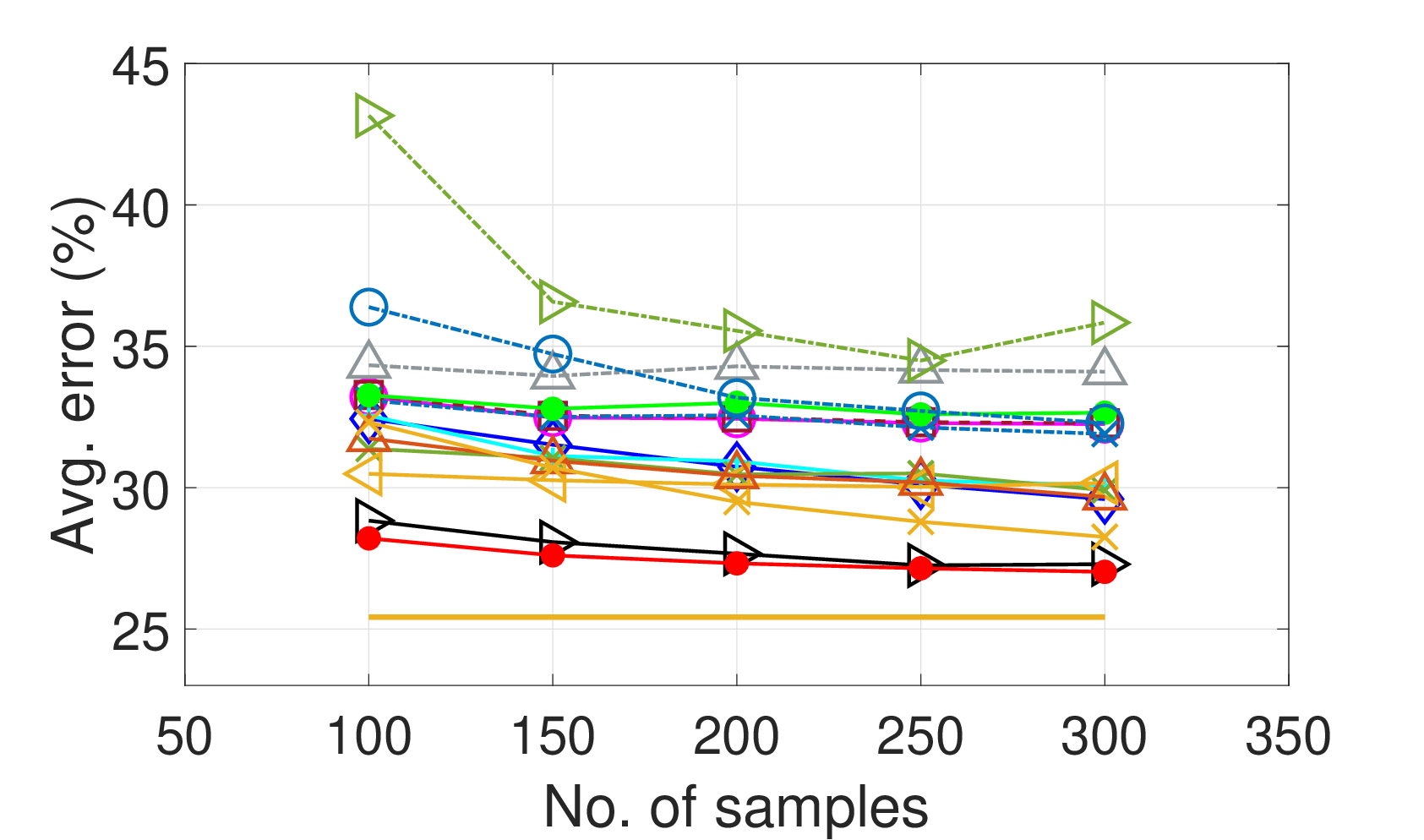} }\hfil
    \subfloat[$\pi_0 = 0.3$, $p=1000$. \label{subfig:M4_1000b}]{\includegraphics{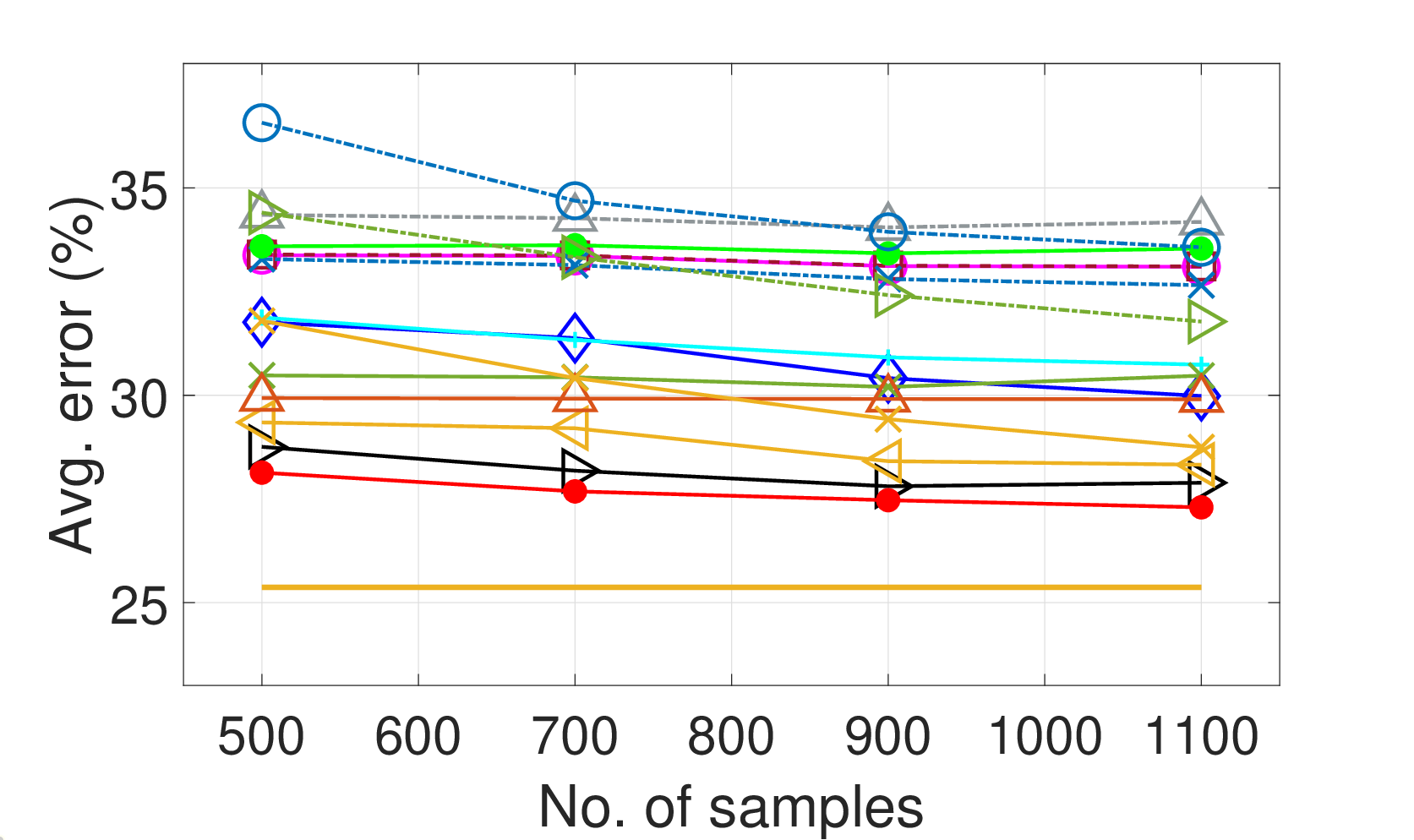} }
    \caption{Model 3: The main diagonal of the covariance matrix is  1, the following four sub-diagonal (lower and upper) elements are 0.9, the next five sub-diagonal  (lower and upper) elements are 0.3, and the remaining elements are zeros.}%
    \label{fig:Model 3}
\end{figure*}

To carry out the tests, we generate 10000 samples of Gaussian data that represent two different populations with a common covariance matrix $\boldsymbol{ \Sigma}$. The covariance matrix is generated such that all its diagonal and off-diagonal elements are equal to 1 and 0.1, respectively. On the other hand, the mean vectors are governed by the rule $\boldsymbol{\mu}_1 = -\boldsymbol{\mu}_0$, where $\boldsymbol{\mu}_0 =k \boldsymbol{1 }$. The parameter $k \in \mathbb{R}$ is calculated based on the squared \textit{Mahalanobis distance} $\nu^2 \triangleq  (\boldsymbol{\mu}_0-\boldsymbol{\mu}_1)^T \boldsymbol{ \Sigma}^{-1}(\boldsymbol{\mu}_0-\boldsymbol{\mu}_1)$. Mahalanobis distance measures the difference between the mean vectors in each group relative to the common within-group covariance matrix \cite{1999-mclachlan-mahalanobis}. 
We assign three different portions of the population data for the training phase, $\frac{1}{20}, \frac{1}{10}$ and $\frac{1}{5}$, which correspond to $n = 50, n=100,$ and $ n=200$, respectively. Equal sampling is performed on the two groups (i.e., $\hat{\pi}_0 =\hat{\pi}_1  = 0.5$ ). 

To implement the grid-search process required to set the regularization parameter value, we define a non-uniform grid of 21 $\gamma$ values in the interval $[1\times 10^{-5}, 1 \times 10^5]$ by using an exponential function $\gamma(j) = 10^{\frac{5j}{10}}$, with $ j = \{-10, -9 \cdots -1, 0, 1 \cdots 10\}.$ This allows a wide range of $\gamma$ values to be evaluated in the search process without excessively increasing the search complexity. 

Before evaluating the performance of the proposed classifier summarized in Section~\ref{subsec:Procedure1}, we examine the classification error variation with the regularization parameter value for different covariance matrix estimators by varying $\gamma$ through the interval $[10^{-5}, 10^5]$. For each $\gamma$ value, we draw $n$ training data samples and compute the sample mean and the sample pooled covariance matrix of each class. We use the remaining data for testing. For each test sample, we compute the score function and determine the class of that sample. We compute the classification error rate by averaging over $5000$ trials for each $\gamma$ value. The dimensionality of data is fixed at $p=100$. We generate data based on three different Mahalanobis distances; $\nu = \sqrt{0.5}, \nu = \sqrt{5}$ and $\nu = \sqrt{9}$. For each distance, we consider three training data sizes $n=50, 100 $, and $200$. 

In the following results, we estimate the probability of misclassification as the average error rate obtained from Monte Carlo simulations. Fig.~\ref{fig:fig1} plots the average error rates versus $\gamma$ for the proposed nonlinear estimator and the benchmark linear estimators along with the Bayes classifier, which is the optimum classifier that assumes perfect knowledge of the underlying data distribution \cite{2015-zollanvari}. From Fig.~\ref{fig:fig1}, we observe that, in all cases, the error curves of the two linear estimators are symmetric around $\gamma = 1$. For $\nu = \sqrt{0.5}$ (Fig.~\subref*{fig:fig1a}), the optimum average classification error is $36.1\%$. For the linear estimators, the minimum average error is $37.5$, while the minimum average error achieved by the proposed estimator is $36.6\%$. For the same Mahalanobis distance $\nu=\sqrt{0.5}$, the minimum error for all estimators decreases as the number of training samples increases to $n=100$ and decreases further at $n=200$. Inspecting the error plots for all $\nu$ and $n$ values, we can conclude that the proposed estimator’s advantage is more visible when both the Mahalanobis distance and the number of training samples are small. The gap between the proposed nonlinear estimator’s and the linear estimator's minimum error shrinks as either the Mahalanobis distance or the training data size increases. The minimum errors of all estimators become nearly the same for sufficiently large $\nu$ and $n$ values, as in Fig.~\subref*{fig:fig1i}.

We have, so far, examined the potential of different covariance matrix estimators dictated by their minimum classification error rates. Now, we look at the performance of the corresponding algorithms and examine their capability to achieve the minimum error. To illustrate this, we consider the proposed grid-search method and the grid-search method associated with the DA-RLDA classifier. We apply these two grid-search methods to obtain the regularization parameter for their respective covariance matrix estimators using the same data of Fig.~\ref{fig:fig1}. Fig.~\ref{fig:barplot} plots the histograms of the obtained $\gamma$ values for the NL-RLDA and DA-RLDA methods calculated from 500 trials. As a reference, we plot the classification-error-versus-$\gamma$ curves for the two methods along with the histograms. From these curves, we can see those good choices of the regularization parameter values for the NL-RLDA and DA-RLDA methods are, approximately, $\gamma\in[10^{0.5},10^5]$ and $\gamma\in[10^{-5},10^{-1.5}]$, respectively. The histograms demonstrate that the proposed approach consistently produces $\gamma$ values in the interval where the error is closest to its minimum value. On the other hand, for the benchmark algorithm, a proportion of the produced $\gamma$ values occur outside the desired interval. This finding implies an advantage of the proposed approach in terms of classification error, even when the (theoretical) minimum classification error is comparable to that of the DA-RLDA method. Up to now, we examined the performance of the proposed estimator along with its linear counterpart and showed some advantages. 

Next, we compare the proposed method with a wide range of competitive techniques that are summarized in Table~\ref{tab:methods}. Figs.~\ref{fig:Model 1} -- \ref{fig:Model 3} plot the average misclassification percentage versus the number of training samples for three different covariance matrix structures: 
    \begin{itemize}
        \item Model 1: The diagonal elements of the covariance matrix are equal to 1, and the off-diagonal elements are  0.1.
        \item Model 2: Autoregressive covariance matrix structure AR(0.9), the diagonal elements of the covariance matrix are equal to 1, and the off-diagonal elements are $0.9^{|i-j|}$.
        \item Model 3: The main diagonal of the covariance matrix is  1, the following four sub-diagonal (lower and upper) elements are 0.9, the next five sub-diagonal  (lower and upper) elements are 0.3, and the remaining elements are zeros.
    \end{itemize}
For each model, we consider dimensions, $p =50, 250$, and $100$ at prior probabilities $\pi_0 = 0.5$ and  $0.3$. As can be clearly seen from Fig. \ref{fig:Model 1} which plots the performance of Model 1, the proposed method shows superiority over all methods in most scenarios. This also can be concluded when examining the results of Model 2 (Fig.~\ref{fig:Model 2}) and Model 3 (Fig.~\ref{fig:Model 3}).

\begin{figure*}
    \centering
    \subfloat[$p =50 $, $n =  25$]{\includegraphics[width=0.3\linewidth]{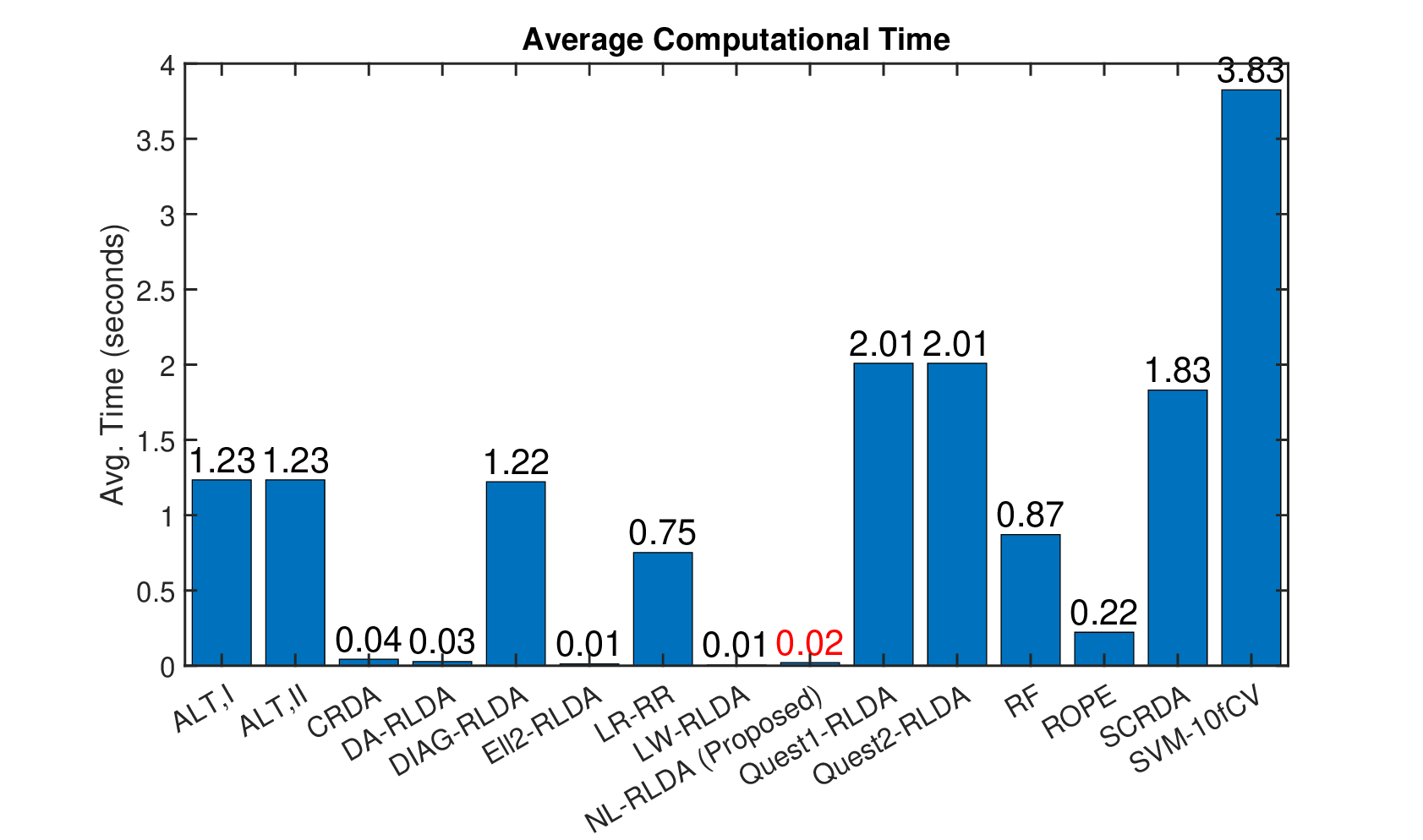}%
    	\label{subfig:ACT1}}
    \hfil
    \subfloat[$p =250 $, $n =  100$ ]{\includegraphics[width=0.3\linewidth]{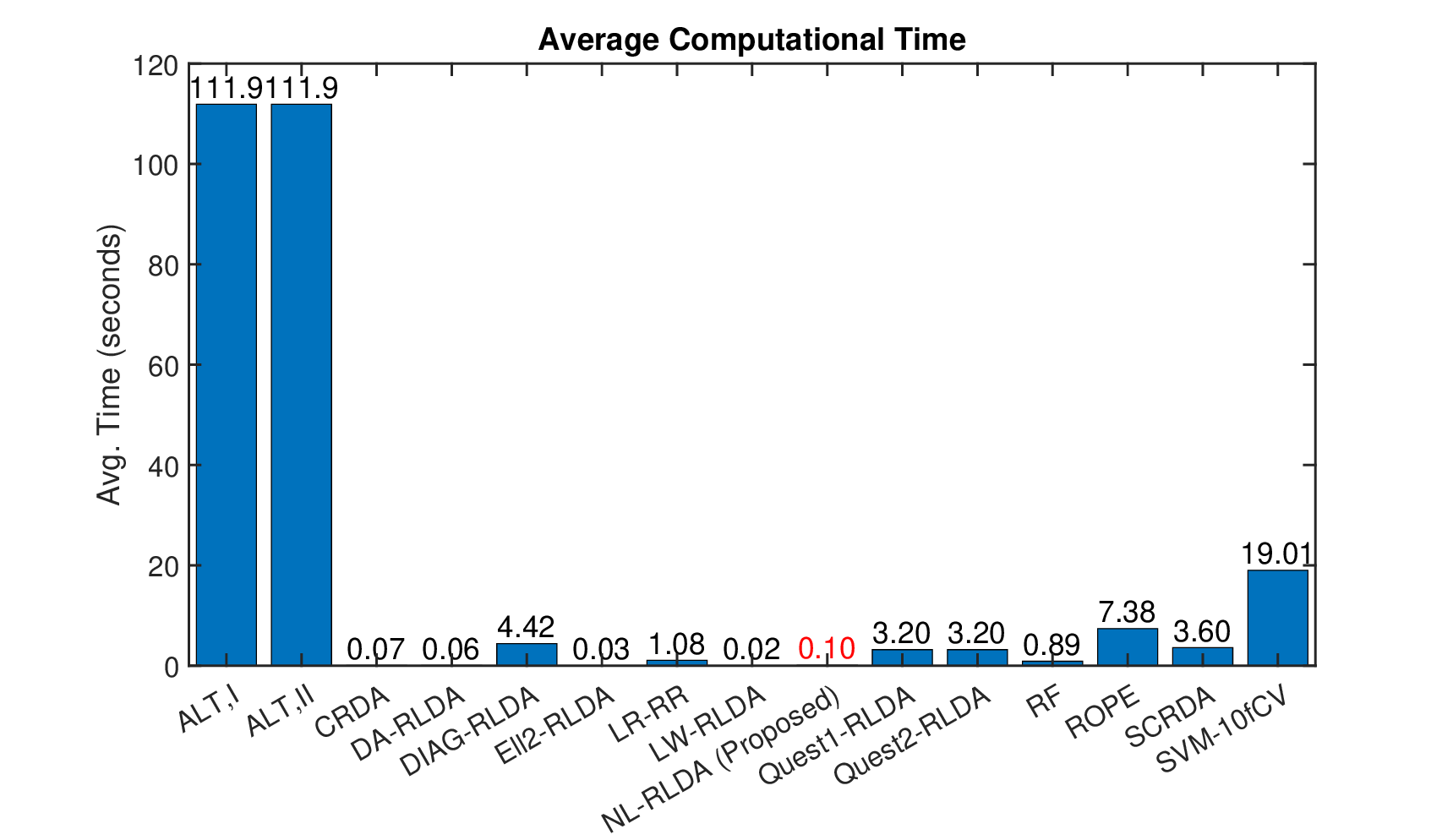}%
    	\label{subfig:ACT2}}
     \hfil
      \subfloat[$p =1000 $, $n =  500$ ]{\includegraphics[width=0.3\linewidth]{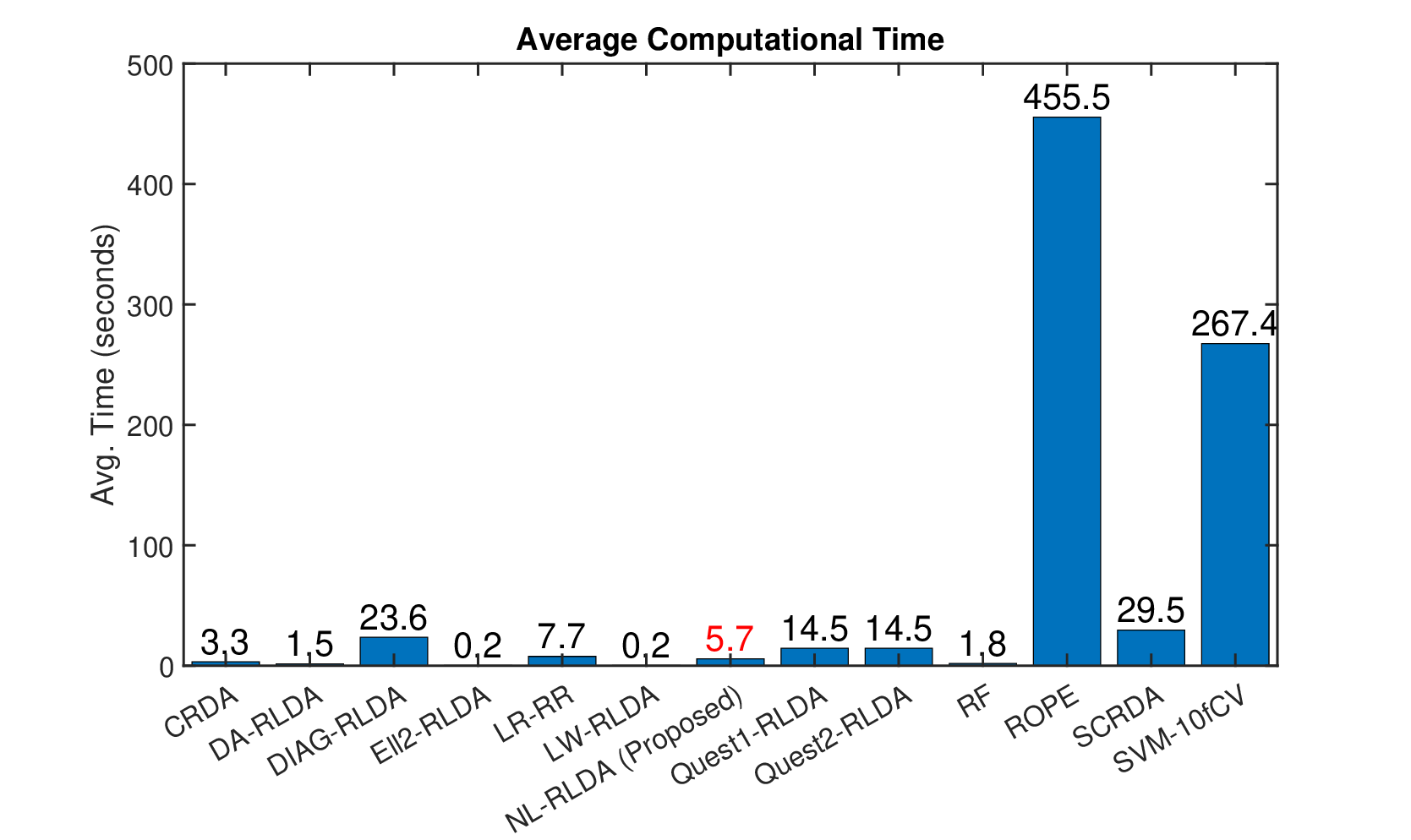}%
    	\label{subfig:ACT3}}
    \caption{The average computational time (in seconds) of all methods using Matlab R2019a running on a 64-bit, Core(TM) i7-2600K 3.40 GHz Windows PC, and  R version 4.0.1. For all methods that require cross-validation or searching, we consider a grid of 21 values.}
    \label{fig:ACT}

	\centering
	\subfloat[Phonemes (`aa',`ao')]{\includegraphics[width=0.35\linewidth]{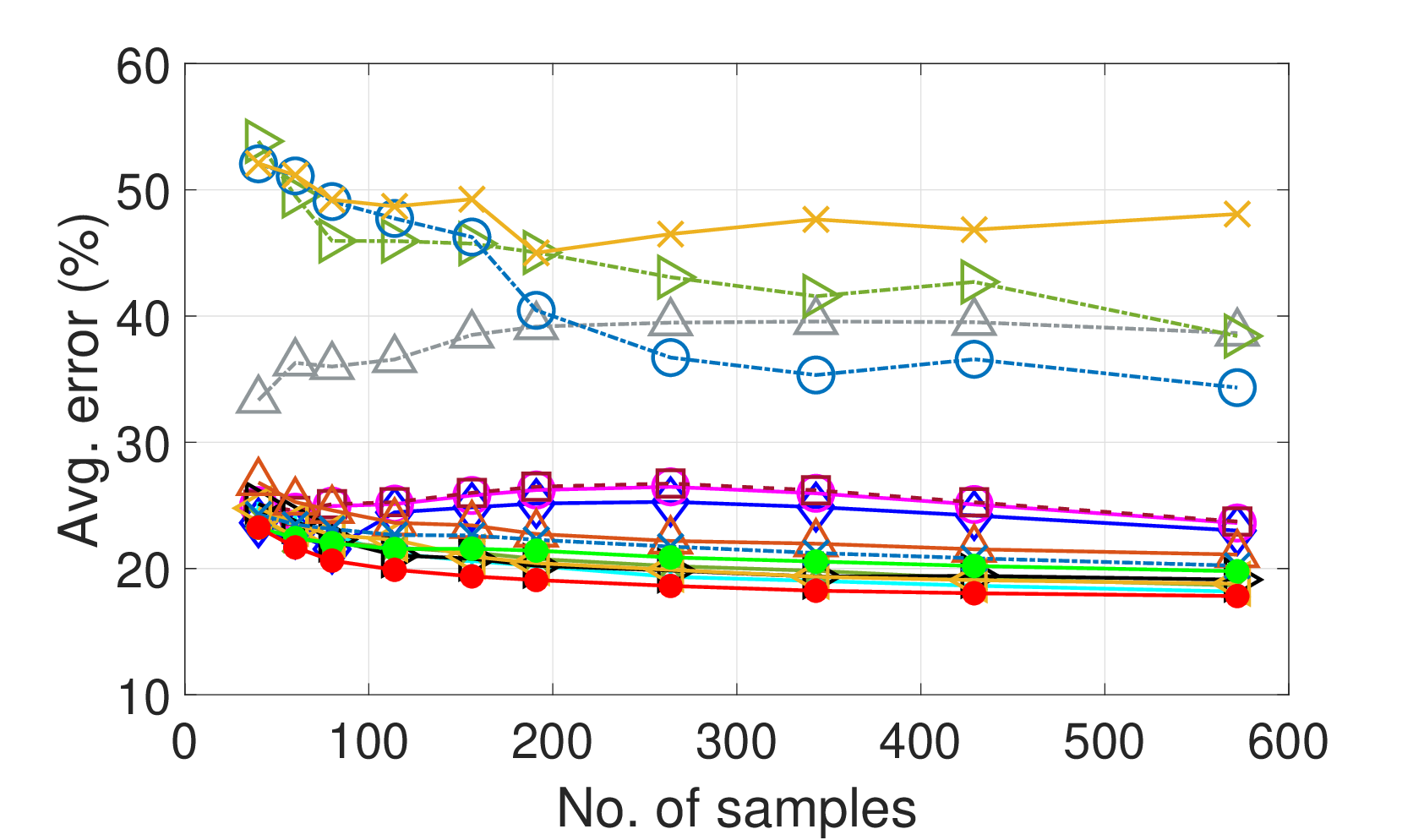}%
		\label{subfig:phonemes}}
	\hfil
	\subfloat[Statlog (gs,dgs)]{\includegraphics[width=0.35\linewidth]{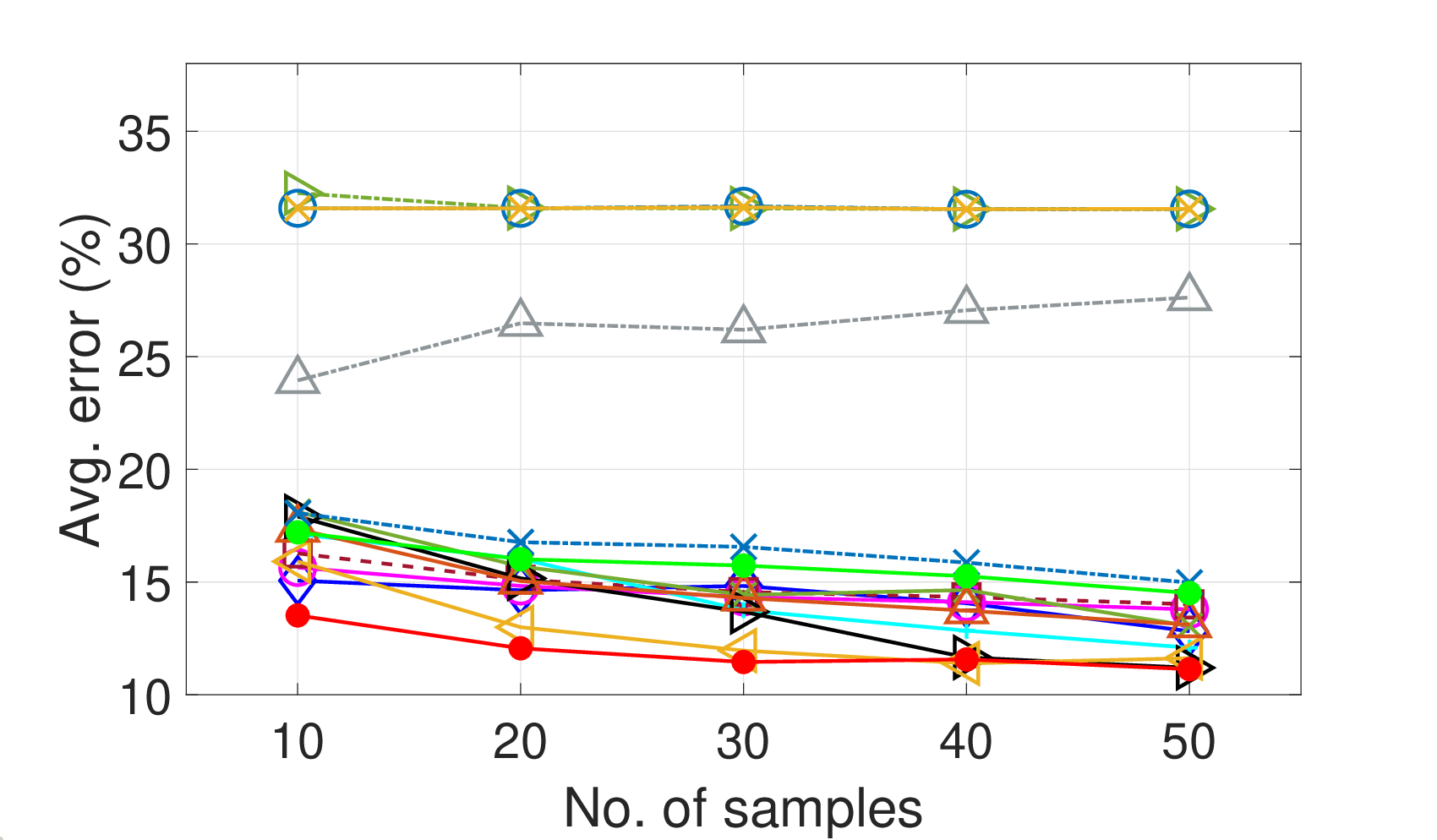}%
		\label{subfig:statlog}}
	\\
	\subfloat[USPS (4,9)]{\includegraphics[width=0.35\linewidth]{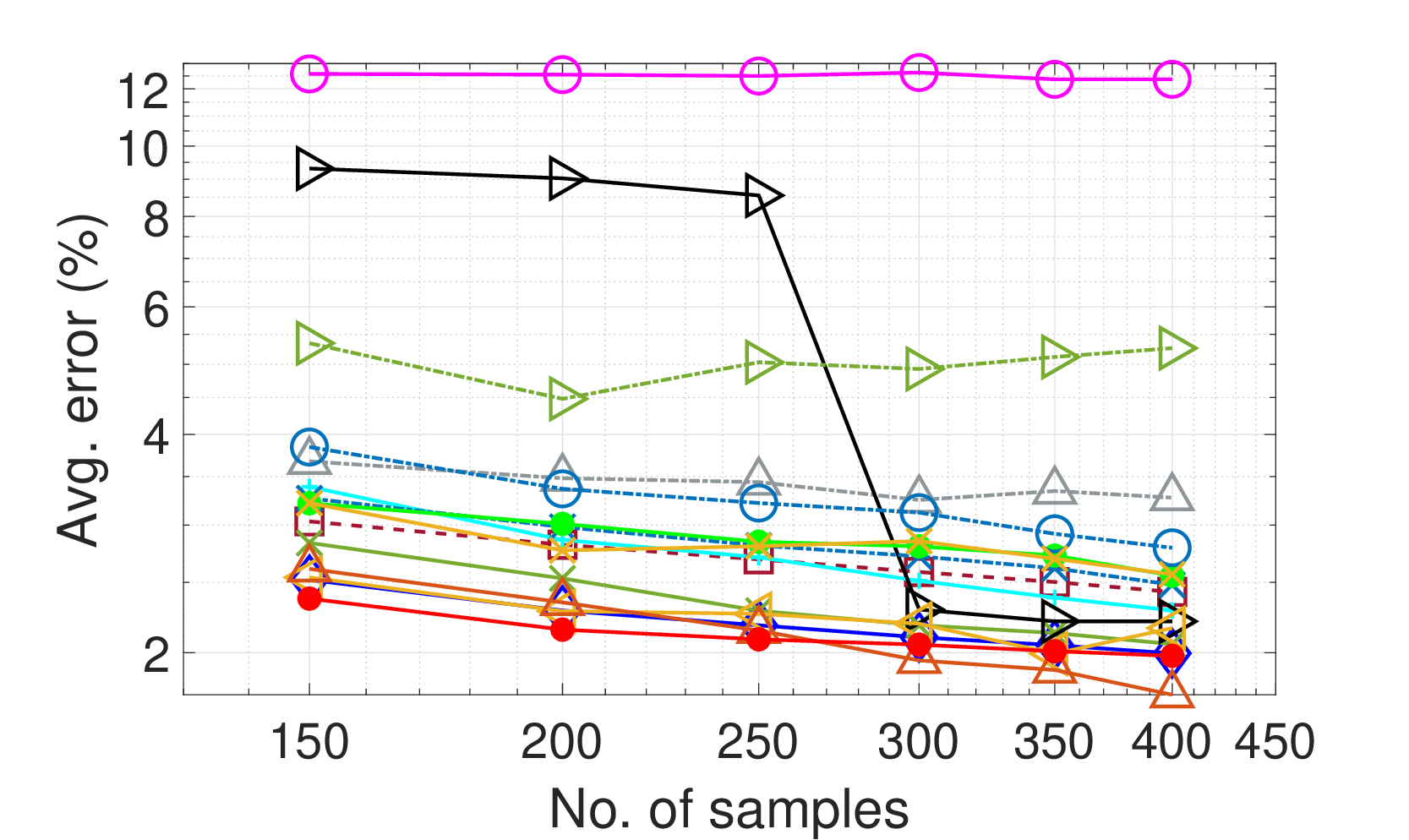}%
		\label{subfig:usps}}
	\hfil
	\subfloat[MINIST (3,8)]{\includegraphics[width=0.35\linewidth]{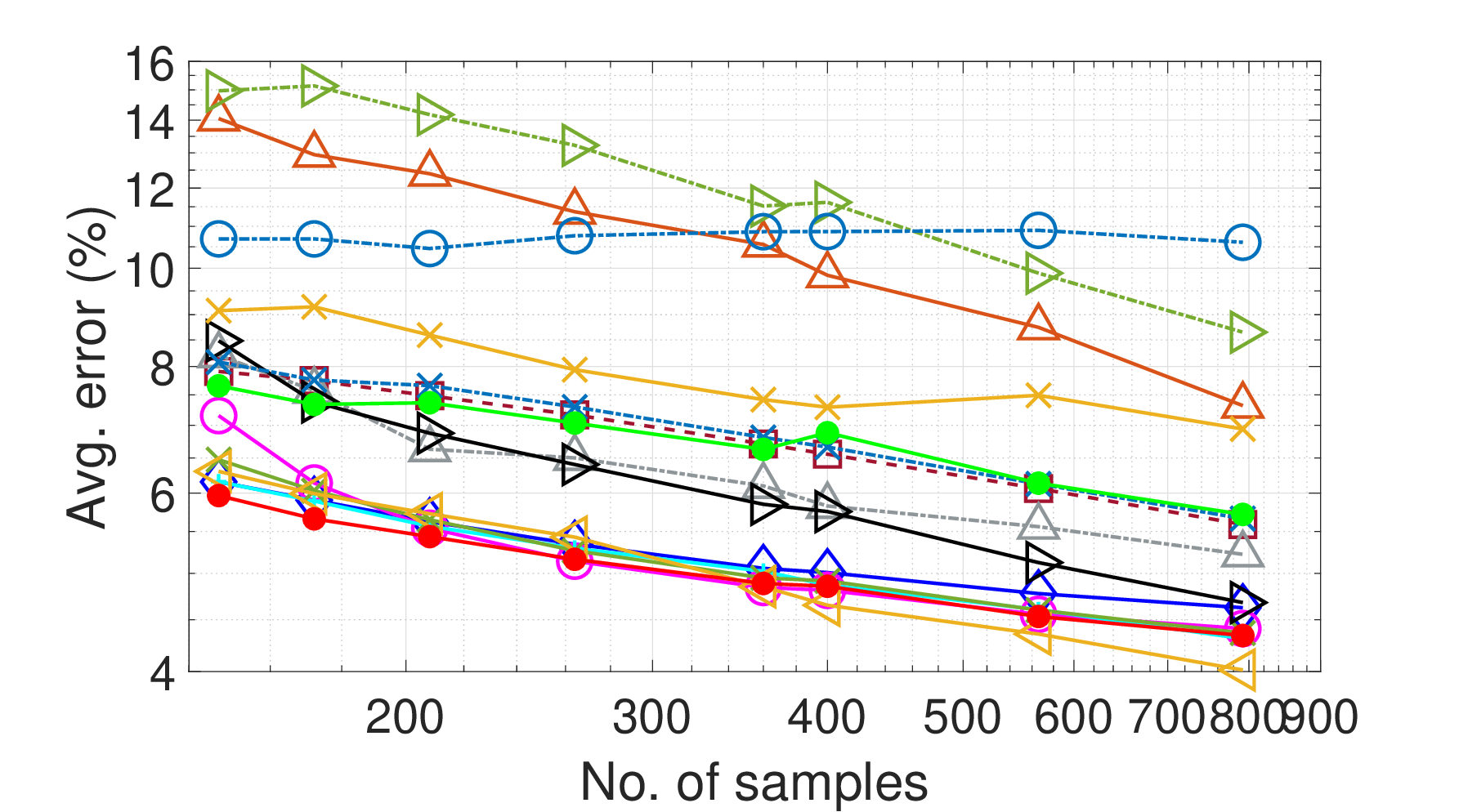}%
		\label{subfig:minist}}
    \\
    	\subfloat[WL (B,M ) ]{\includegraphics[width=0.35\linewidth]{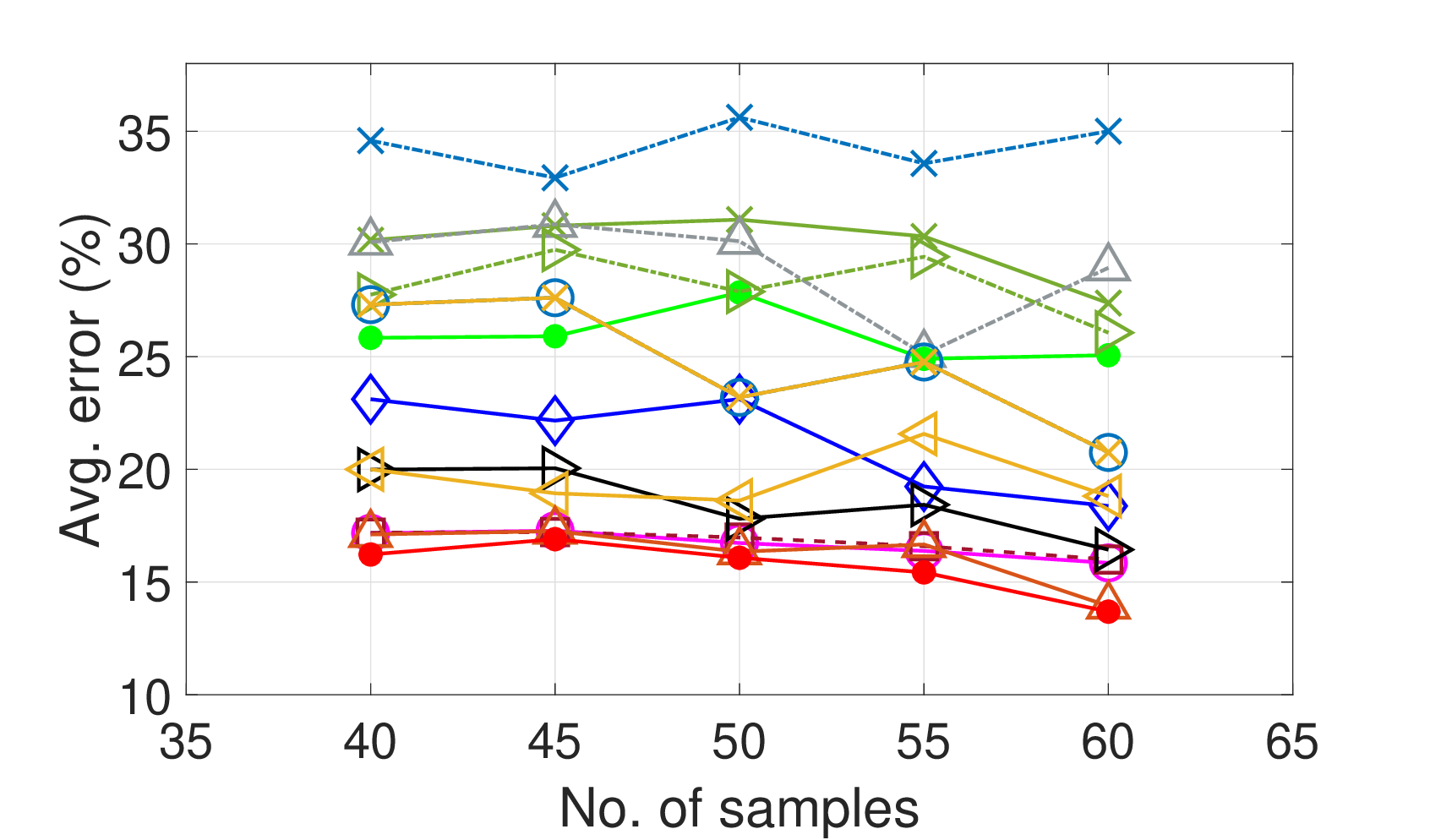}%
    	\label{subfig:WLBM}}
     \hfil
     \subfloat[WL (H,A) ]{\includegraphics[width=0.35\linewidth]{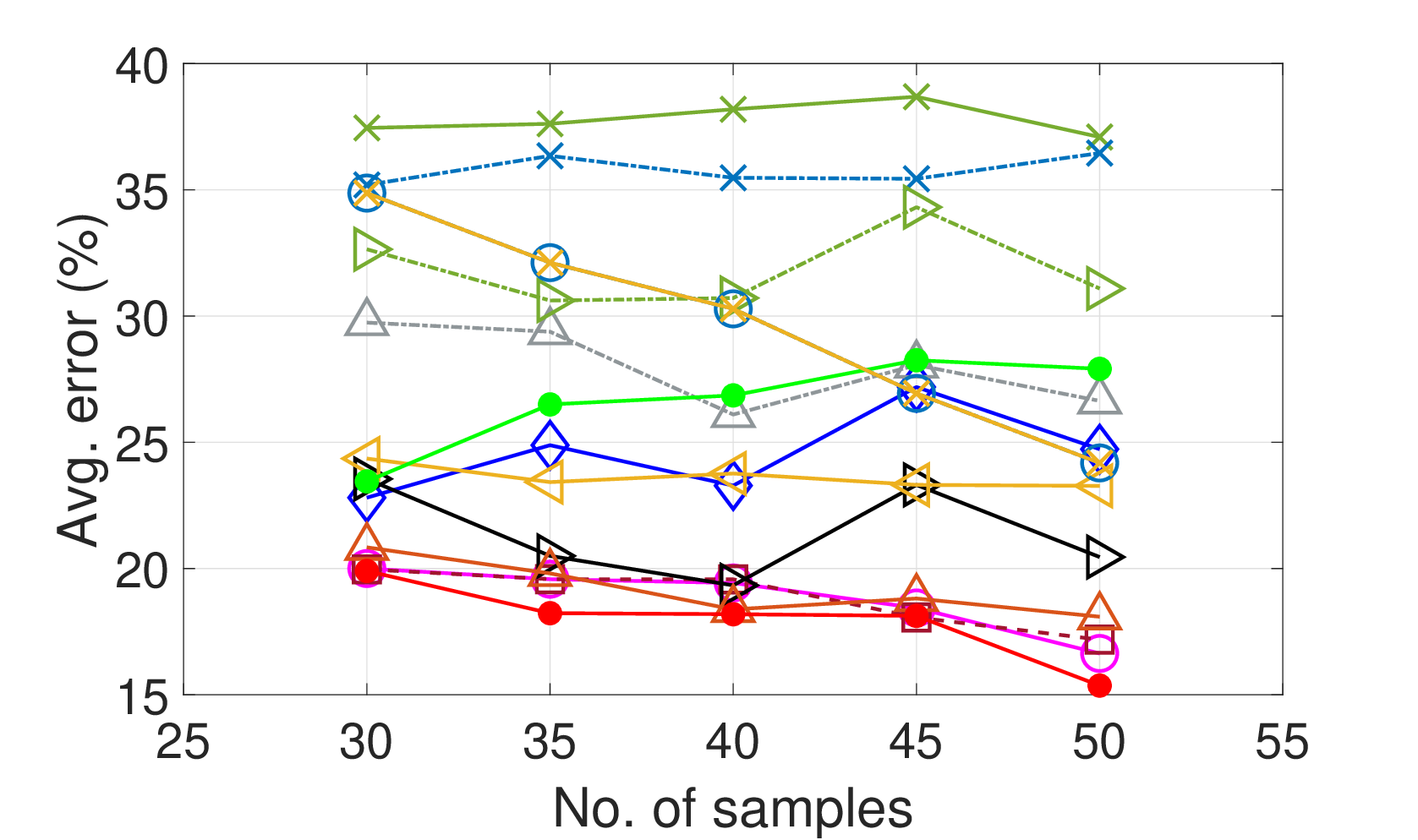}%
    	\label{subfig:WLHA}}
     \\
    \subfloat[NBI (B,M ) ]{\includegraphics[width=0.35\linewidth]{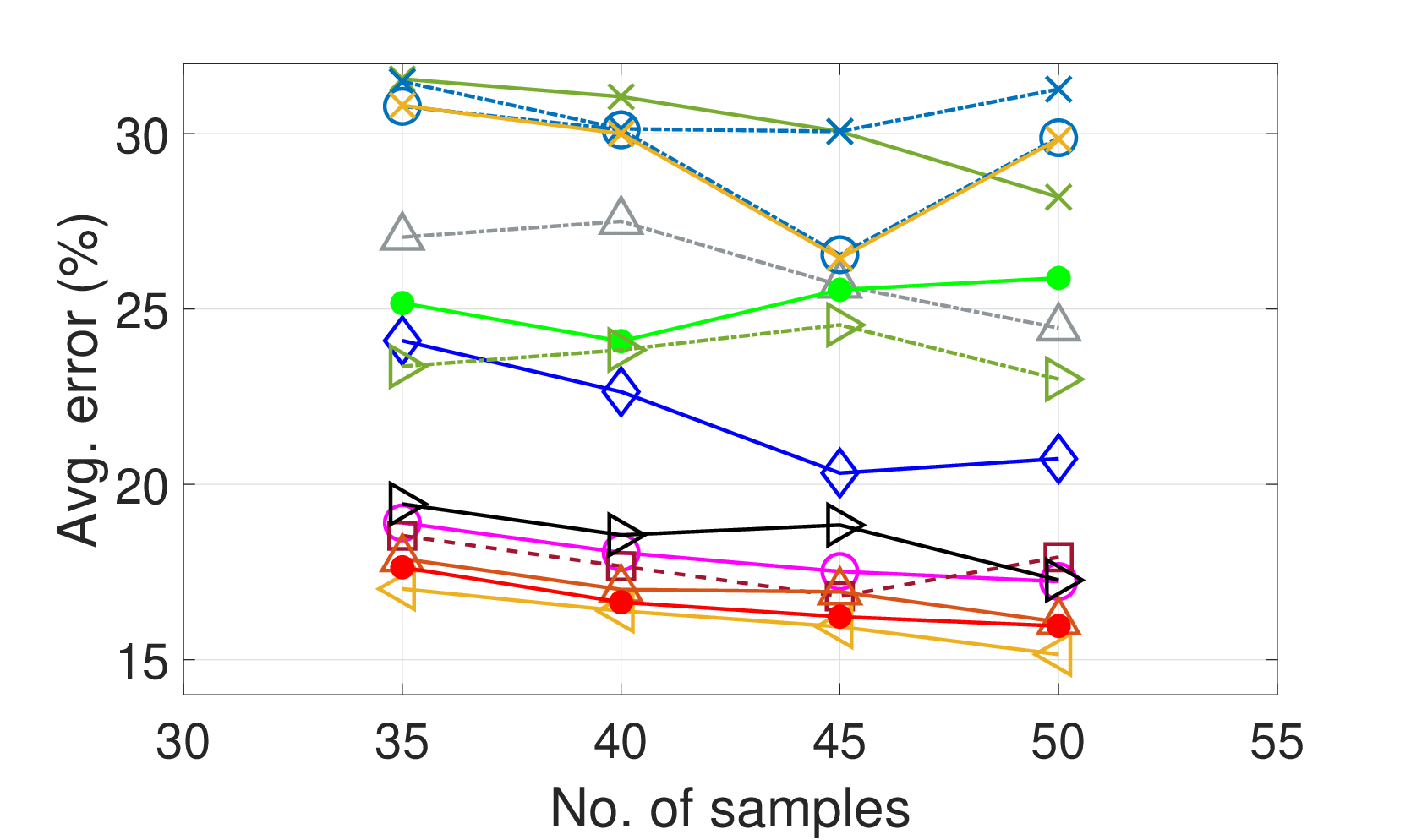}%
    	\label{subfig:NBIBM}}
     \\
  	\subfloat{\includegraphics[width=10cm, height=1cm]{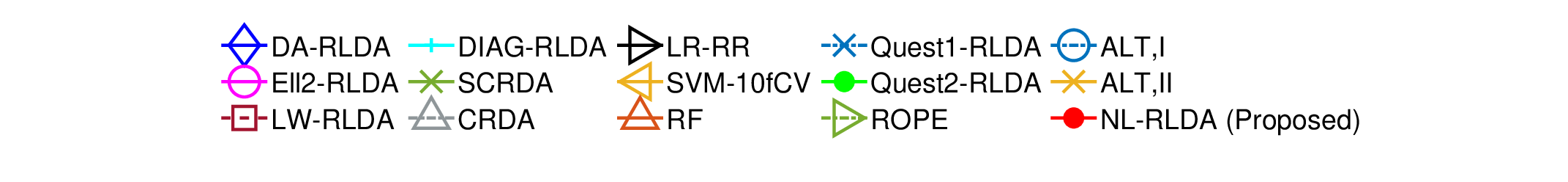}%
		}
	\caption{Performance comparison for different real datasets.}
	\label{fig:fig4}	
\end{figure*}

The average computational time of all methods is shown in Fig.~\ref{fig:ACT} using Matlab R2019a running on a 64-bit, Core(TM) i7-2600K 3.40 GHz Windows
PC, and  R version 4.0.1. For all methods that require cross-validation or searching, we consider a grid of 21 values. It is clear from Figs.~\subref*{subfig:ACT1} -- \subref{subfig:ACT3} that the proposed method has a reasonably low complex running time compared to some methods. 

\subsection{Real Data}
Although we derive our consistent estimator under the Gaussian assumption, it is interesting to examine its performance in real datasets. For this purpose, we test a few examples from the Phoneme dataset \cite{1995-hastie-penalized}, Statlog (Landsat satellite) dataset, USPS dataset \cite{1989-le-handwritten}, and  MINIST  dataset \cite{1995-grother-nist,1998-lecun-gradient}. 
We utilize the same grid-search strategy used with the synthetic data in the previous subsection. In the remainder of this subsection, we discuss the results obtained from the real-data examples. The properties of the real datasets used in this section are summarized in Table~\ref{tab:realdata}.
    
\begin{table}[!h]
\centering
    \begin{threeparttable} 
		\caption{Real data used for comparison}
        \label{tab:realdata}
  \begin{small}
	\begin{tabular}{@{}lccccc@{}}
		\toprule
		Dataset           &   $p$        & $n_0^*$  & $n_1^*$ & $\widehat{\pi}_0$ & $\widehat{\pi}_1$ \\ \midrule
		 Phonemes (`aa',`ao')   &   256 & 695  & 1022  & $40.48\%$   & $59.52\%$       \\ 		
		Statlog (gs,dgs)  &   36  & 1358 & 626   & $68.45\%$   & $31.55\%$       \\ 		
		MNIST (3,8)      &   400 & 2042 & 1918  & $51.57\%$   & $48.43\%$       \\
		USPS (4,9)        &   256 & 852  & 821   & $50.93\%$   & $49.07\%$       \\
        WL (B,M)          &   698 & 21  & 55   & $27.63\%$   & $72.37\%$       \\
        WL (H,A)          &   698 & 21  & 40   & $34.43\%$   & $65.57\%$       \\
        NBI (B,M)         &   698 & 21  & 55   & $27.63\%$   & $72.37\%$       \\
  \bottomrule
	\end{tabular}
      \begin{tablenotes}
        \item Note: $n_0^*$, $n_1^*$ are the population size of classes 0 and 1, respectively.
      \end{tablenotes}
    \end{small}
  \end{threeparttable}
\end{table}

$\bullet$ \textbf{The phonemes dataset} is extracted from the TIMIT database, a widely used speech recognition dataset. The phonemes dataset consists of five phoneme data of dimension $p=256$. As an example, we test the performance of the different classifiers using the two phonemes `aa' and `ao'. The task of distinguishing these two phonemes is the most confusing one in the dataset because the distance between the two classes is the shortest \cite{1995-hastie-penalized}. The total samples of the first class is $n^*_0 = 695$ and of the second class is $n^*_1 = 1022$.

Fig.~\subref*{subfig:phonemes}, plots the average misclassification rates versus the number of training samples $n \in [40,572]$. It can clearly be seen that the NL-RLDA method, again, achieves the best performance among all the methods.

$\bullet$ \textbf{Statlog (Landsat satellite) dataset } consists of a multi-spectral pixel values in a $3 \times 3$ neighborhoods in a satellite image of $p=36$. The task we consider here is to classify between grey soil `gs' (of population $n_0^* =1358$) and damp grey soil `dgs' ($n_1^*=626$).
Fig.~\subref*{subfig:statlog} plots the average error rate against training samples, $n \in [10,50]$. The figure reveals the superiority of the proposed method for $n < 40$.

$\bullet$ \textbf{The USPS dataset}  is a popular handwritten digits dataset \cite{1989-le-handwritten} of dimension $p=256$. We consider classifying the digits (4,9) of population size $n_0^* = 852$ and $n_1^* = 821$.
Fig.~\subref*{subfig:usps} shows the average error rate performance for $n \in [100, 150]$. It can be seen that the proposed NL-RLDA method and the DA-RLDA method show comparable error rates. These two methods offer the best performance among all the tested algorithms, with a small advantage for the proposed method. 

$\bullet$ \textbf{The MINIST dataset} is another widely-used handwritten digits dataset of dimension $p=400$. As our two classes, we consider the digit pair (3,8). 

Fig.~\subref*{subfig:minist} plots the average error rate versus the number of training samples $n \in [150, 800]$  obtained from population $n_0^* = 2042$ and $n_1^*=1918$. As can be seen from the figure, the performance of all methods improves as the number of training samples increases. The proposed NL-RLDA method shows an improved performance compared to the other methods when the samples are scarce.

$\bullet$ \textbf{Gastrointestinal Lesions in Regular Colonoscopy dataset}  includes features extracted from colonoscopic videos of gastrointestinal lesions. The dataset contains feature vectors of 698, ($p=698$) for $76$ lesions, which are classified into three types: Hyperplasic ($n^*_\text{H}=21$), Adenoma ($n^*_\text{A}=40$), and Serrated adenoma ($n^*_\text{S}=15$). Each lesion is recorded twice using two different types of lights: White Light (WL) and Narrow-Band Imaging (NBI), resulting in duplicate recordings for each lesion.  It is possible to simplify the classification problem by combining adenoma and serrated adenoma into one class, resulting in a binary classification problem. Under this classification, Hyperplasic lesions are classified as `Benign'($n^*_0=21$), while adenoma and serrated adenoma are classified as `Malignant' ($n^*_1 = 55$) \cite{colonoscopyDataset}. 

Fig.~\subref*{subfig:WLBM} -- \subref{subfig:NBIBM}  plots the average error rate versus the number of training samples for some classification examples from the dataset. Fig.~\subref*{subfig:WLBM} plots the performance when classifying (B,M) (`Benign',`Malignant') using the WL data for training samples $n \in [40, 60]$  obtained from population $n_B^* = 21$ and $n_M^*=55$. As can be seen from the figure, the performance of the proposed method shows an improved performance compared to the other methods when the samples are scarce. A similar behavior is revealed when classifying (H,A) (Hyperplasic,Adenoma) using the WL data in Fig.~\subref*{subfig:WLHA}. Finally, Fig.~\subref*{subfig:NBIBM} shows the performance of the classification task (B,M) using the NBI data. 
It is interesting to note that the proposed method can perform well when $p>>n$.

\section{Conclusion}
\label{sec:conc}
We have proposed a novel estimator of the linear discriminant analysis (LDA) pooled covariance matrix inverse. Starting from the original LDA score function, we have derived an expression for a regularized LDA (RLDA) score function based on the bounded data uncertainty model. This expression implicitly entails a nonlinear (NL) estimator of the pooled covariance matrix, leading to our proposed NL-RLDA classifier. We have derived the classifier’s asymptotic misclassification rate and a consistent estimator of the rate, which allows us to tune the regularization parameter value. Performance comparisons with common LDA classifiers that use linear and nonlinear covariance matrix estimators, and precision matrix estimators, demonstrate the superiority of our proposed classifier.

\appendices
\section{Mathematical tools}
\label{sec:append}
We form a data matrix $\Xm$ that is of a $p \times n$ dimension ($n=n_0 + n_1$) as follows:
\begin{equation}
\label{eqn:Xn}
\Xm= [\Xm_0, \Xm_1] = [\xv_{1}, \xv_{2},\ldots, \xv_{n_0}, \xv_{n_{0+1}}, \ldots, \xv_{n}],
\end{equation}
where $\xv_i \sim \mathcal{N} (\muv_0, \Sigmam)$, for $i=1,\ldots, n_0$ and $\xv_i \sim \mathcal{N} (\muv_1, \Sigmam)$, for $i=n_{0+1},\ldots, n$.   
Similarly,  
\begin{equation}
\Zm = [\Zm_0, \Zm_1] = [\zv_{1}, \zv_{2},\ldots, \zv_{n_0}, \zv_{n_{0+1}}, \ldots, \zv_{n}],
\end{equation}
where $\zv_i \sim \mathcal{N}(\boldsymbol{0}, {\bf I}_p)$.
Hence, the data model can be represented in the following matrix form: 
\begin{equation}
    \label{eqn:DataMatrixForm}
    \Xm = \Sigmam^{\frac{1}{2}} \Zm  + [\muv_0 \boldsymbol{1}^T_{n_0}, \muv_1 \boldsymbol{1}^T_{n_1} ], 
\end{equation}
where $\boldsymbol{1}_{n_0}$ and $\boldsymbol{1}_{n_1}$ are vectors of 1s of sizes $n_0$ and $n_1$, respectively.  We also can estimate the covariance matrix of each class ($i=0,1$) using matrix notation, 
\begin{equation}
    \Sm_i  = \dfrac{1}{n_i - 1} \Bm_i \Bm_i^T, 
\end{equation}
where $\Bm_i \triangleq \Xm_i - \mv_i \boldsymbol{1}_{n_i}^T $. 
The pooled sample covariance matrix is 
\begin{equation}
    \label{eqn:PooldedSCMmatrix}
    \Sm = \dfrac{1}{n-2} [(n_0-1)\Sm_0 + (n_1 -1) \Sm_1 ] 
\end{equation}
It can be easily shown that $\Sm$ can be written as 
\begin{equation}
    \Sm= {{1}\over {n-2}}\left({\Sigmam}^{1/2} \Zm \Gm \Zm^T\Sigmam^{1/2} \right),
\end{equation}
where $\Gm$ defines  the following matrix \cite{2015-zollanvari}: 
\begin{equation}
\label{eqn:Gn}
{\bf G}= {\bf I}_{n}-\left[\begin{IEEEeqnarraybox*}[][c]{,c/c,}
{{{\bf 1}_{n_{0}} {\bf 1}_{n_{0}}^{T}}\over {n_{0}}} &{\bf 0}_{{n_{0}}\times n_{1}}\\
{\bf 0}_{{n_{1}}\times n_{0}}&{{{\bf 1}_{n_{1}} {\bf 1}_{n_{1}}^{T}}\over{n_{1}}}
\end{IEEEeqnarraybox*}
\right].
\end{equation}

Next, we use the eigenvalue decomposition of ${\bf G}$ and let ${\Tm}$  be a diagonal matrix defined by the eigenvalues of ${\bf G}$, (i.e., $\Gm = \Um \Tm \Um^T$), we can write the following:
\begin{align}
    \Sm &= {{1}\over {n-2}}\left ({\Sigmam}^{1/2} \tilde{\Zm} {\Tm} \tilde{\Zm}^{T} {\Sigmam}^{1/2}\right), \label{eqn:SmModel}\\
    &={{{\bf Y} {\bf Y}^{T}}\over {\tilde{n}}}, \label{eqn:Sy}
\end{align}
where $\tilde{n} \triangleq n-2$, and $\Ym \triangleq {\Sigmam}^{1/2} \tilde{\Zm} {\Tm}$.  Note that $\Zm$ has the same distribution of $\tilde{\Zm}:=\Um\Zm$  (invariance under orthogonal transformation). Also, because  ${\bf G}$ has the special structure in (\ref{eqn:Gn}), ${\Tm}$  is a diagonal matrix with the diagonal elements  $[\boldsymbol{1}^T_{n_0-1},0, \boldsymbol{1}^T_{n_1-1},0]$, which makes $\Tm$ idempotent. 

The model \eqref{eqn:SmModel}  is a well-established model in the RMT literature (e.g., see \cite{2012-rubio}), and we can use some immediate results of this model.
\begin{lemma}
The following results hold true almost surly (see \cite{2012-rubio} and \cite{2015-zollanvari}):
\end{lemma}
\begin{align}
&\dfrac{1}{\tilde{n}} \text{tr}[\Thetam\Sm \Qm] \asymp \dfrac{1}{\tilde{n}} x \trace\Big[\Thetam \Sigmam (x\Sigmam-z{\bf I}_p)^{-1}\Big], \label{eqn:DE1}\\
&\dfrac{1}{\tilde{n}} \text{tr}[\Thetam\Sm \Qm^2] \asymp \dfrac{1}{\tilde{n}} (x-zx^\prime) \trace\Big[\Thetam \Sigmam (x\Sigmam-z{\bf I}_p)^{-2}\Big], \label{eqn:DE2}\\
&\dfrac{1}{\tilde{n}} \text{tr}[\Thetam\Qm\Sigmam\Qm] \asymp \dfrac{1}{\tilde{n}} \dfrac{1}{1-\phi \tilde{\phi}} \trace\Big[\Thetam \Sigmam (x\Sigmam-z{\bf I}_p)^{-2}\Big], \label{eqn:DE-1-1}
\end{align}	
where $\Thetam \in \mathbb{R}^{p \times p}$ is a bounded norm matrix, $\{e=e(z), x=x(z)\}$ is the unique solution for the following system of equations:
\begin{equation}
\begin{cases}
e = \frac{1}{\tilde{n}} \text{tr} \big[\Sigmam (x\Sigmam-z{\bf I}_p)^{-1}\big] \\
x = \frac{1}{\tilde{n}} \text{tr} \big[ {\Tm} (e{\Tm} +{\bf I}_{\tilde{n}})^{-1}\big].
\end{cases}
\label{eqn:systeqn}
\end{equation}
For the special structure of ${\Tm} $, we obtain $x$ as defined in  Table~\ref{tab:def}.

\section{}
\begin{IEEEproof} [Proof of Theorem \ref{thm:1}]
Let $\uv=[{\bf 1}_{n_{0}}^{T}/n_{0}, {\bf -1}_{n_{1}}^{T}/n_{1}]^{T}$, $\vv=[{\bf 1}_{n_{0}}^{T}/n_{0}, {\bf 1}_{n_{1}}^{T}/n_{1}]^{T}$ and $\muv =\muv_{0}- \muv_{1}$. Using (\ref{eqn:Xn}), we can write \cite{2015-zollanvari}
\begin{align}
G\big ( &\muv_{0}, \mv_{0}, \mv_{1}, \Hm\big) \nonumber\\
=&\, \left(\muv_{0}- {{ \mv_{0}+ \mv_{1}}\over {2}}\right) ^{T} \Hm\left ( \mv_{0}- \mv_{1}\right) \nonumber\\ 
=&\, {\boldsymbol{\mu} }_{0}^{T} \Hm \Xm \uv- {{1}\over {2}}\vv^{T} \Xm^{T} \Hm \Xm \uv \nonumber\\
= &\, {{1}\over {2}} {\boldsymbol{\mu} }^{T} \Hm {\boldsymbol{\mu} }- {{1}\over {2}} \vv^{T} \Zm^{T} {\Sigmam}^{1/2} \Hm {\Sigmam}^{1/2} \Zm \uv \nonumber \\& - {{1}\over {2}}({{\boldsymbol{\mu}}}_{0}+{{\boldsymbol{\mu}}}_{1})^{T} \Hm {\Sigmam}^{1/2} \Zm \uv - {{1}\over {2}} \vv^{T} \Zm^{T} {\Sigmam}^{1/2} \Hm {\boldsymbol{\mu} }.
\end{align}
In the Gaussian case, the sample mean $\mv_i$, $i=0,1$, is independent from the sample covariance matrices $\Sm_j$, $j=0,1$, under certain conditions, (see Theorem 11.15 in \cite{2016-schott-matrix}); therefore, $\Xm \uv$ and $\Xm \vv$ are independent of $\Hm$. Note that $\mathbb{E}[({\Sigmam}^{1/2} \Zm \vv) (\uv^{T} \Zm^{T} {\Sigmam}^{1/2})]= {{\Sigmam}\over {n_{0}}}- {{\Sigmam}\over {n_{1}}}$. Therefore,
\begin{equation}
G\big ( \muv_{0}, \mv_{0}, \mv_{1}, \Hm\big) 
=  \dfrac{1}{2}\text{tr} [\boldsymbol{\mu}\boldsymbol{\mu}^T \Hm] + (\dfrac{1}{n_1}-\dfrac{1}{n_0}) \text{tr} [ \Sigmam \Hm].
\label{eqn:Gexp}
\end{equation}
Finding the deterministic equivalents of (\ref{eqn:Gexp}) using (\ref{eqn:DE2}) results in (\ref{eqn:Gasymp}).
Similarly, we can find that
\begin{align}
D\big (& \mv_{0}, \mv_{1}, \Hm, {\Sigmam}\big) \nonumber \\
&= \text{tr} [\boldsymbol{\mu}\boldsymbol{\mu}^T \Hm\Sigmam \Hm] + \left(\dfrac{1}{n_1}+\dfrac{1}{n_0} \right) \text{tr} [ \Sigmam \Hm\Sigmam \Hm] 
\end{align}
In order to find  the deterministic equivalent of the form $\text{tr} [\Thetam\Hm\Sigmam \Hm]$, with $\Thetam = \Sigmam$ or $\Thetam = \muv\muv^T$,  for the proposed $\Hm$ in (\ref{eqn:HNL}), we can write
\begin{align}
\text{tr} [\Thetam \Hm\Sigmam \Hm]
&= \text{tr} [\Thetam \Sm \Qm^2\Sigmam \Sm \Qm^2]\nonumber \\ 
& = \text{tr} [\Thetam \Qm\Sigmam \Qm] 
+	z \text{tr} [\Thetam \Qm\Sigmam \Qm^2] \label{eqn:trThetaHSH}\\
&+	z \text{tr} [\Thetam \Qm^2\Sigmam \Qm] + z^2 \text{tr} [\Thetam \Qm^2\Sigmam \Qm^2].\nonumber  			   
\end{align}
The first deterministic equivalent of the first term in \eqref{eqn:trThetaHSH} can be obtained from (\ref{eqn:DE-1-1}). We note that if $\Thetam$ is symmetric, which is the case we have here, the following relation is valid for a product of symmetric matrices:
\begin{equation}
\text{tr} [\Thetam \Qm\Sigmam \Qm^2] = \text{tr} [\Thetam \Qm^2\Sigmam \Qm].
\end{equation}
Hence, the deterministic equivalent of the second and the third term in \eqref{eqn:trThetaHSH} can be obtained by differentiating both sides in (\ref{eqn:DE-1-1}) to obtain
\begin{align}
&\dfrac{1}{\tilde{n}} \text{tr}[\Thetam\Qm^2\Sigmam\Qm] \asymp\dfrac{1}{\tilde{n}}\Big( \dfrac{\phi \tilde{\phi}^\prime + \phi^\prime \tilde{\phi} + 2\frac{x^\prime}{x}}{(1-\phi \tilde{\phi})^2} \Big)\trace\Big[\Thetam \Sigmam (x\Sigmam-z{\bf I}_p)^{-2}\Big]  \nonumber\\
&+   2\Big(  \dfrac{1}{1-\phi\tilde{\phi}}-\dfrac{ \frac{x^\prime}{x}}{(1-\phi \tilde{\phi})^2}z \Big) \trace\Big[\Thetam \Sigmam (x\Sigmam-z{\bf I}_p)^{-3}\Big].
\end{align}
The deterministic equivalent of the last term in \eqref{eqn:trThetaHSH} is a bit more involved. We will use the following result from \cite{2009-rubio}, for any $z_1$, $z_2 \in \mathbb{C}^{+}$ :

\begin{align}
&\dfrac{1}{\tilde{n}} \text{tr}[\Thetam(\Sm-z_1{\bf I})^{-1}\Sigmam(\Sm-z_2{\bf I})^{-1}] \nonumber \\ 
&\asymp \dfrac{1}{\tilde{n}} \dfrac{\text{tr}\big[\Thetam {\bf P }(z_1) \Sigmam {\bf P} (z_2)\big]}{1-w(z_1)w(z_2)\frac{c}{\tilde{n}}\text{tr}\big[\Sigmam {\bf P }(z_1) \Sigmam {\bf P} (z_2)\big]} \label{eqn:ThetaHThetaH},
\end{align}
where $\Pm$ and $w$ are defined in   Table~\ref{tab:def}.
We differentiate \eqref{eqn:ThetaHThetaH}  twice, one with respect to $z_1$ and the other with respect to $z_2$, and finally substitute with $z_1=z_2=z$ to obtain the final result which is the second and the third term of (\ref{eqn:asym}) (without the factor $z^2$). 
\end{IEEEproof}

\section{}
\begin{IEEEproof} [Proof of Theorem \ref{thm:2}]
It can be shown that \cite{2020-khalili}
\begin{equation}
G\big ( \mv_{i}, \mv_{0}, \mv_{1}, \Hm\big) =	G\big ( { \boldsymbol{\mu}}_{i}, \mv_{0}, \mv_{1}, \Hm\big) - \dfrac{1}{n_i} \text{tr}[{\Sigmam \Hm}].
\end{equation}
Our goal is to achieve an almost sure convergence for the following: 
\begin{equation}
{\dfrac{1}{n_i} \text{tr}[{\Sigmam \Hm}]- \theta_G}  \rightarrow_{a.s} 0 , 
\end{equation}
where
\begin{equation}
\theta_G = \dfrac{1}{\tilde{n}}(x-zx^\prime) \trace\Big[\Sigmam^2 (x\Sigmam-z{\bf I}_p)^{-1}\Big].
\end{equation}
We note that from (\ref{eqn:DE1}), if $\Thetam = {\bf I}$, then
\begin{align}
\dfrac{1}{\tilde{n}} \text{tr}[ \Sm\Qm] \asymp \dfrac{1}{\tilde{n}} x\text{tr} [\Sigmam (x\Sigmam-z{\bf I})^{-1}].
\end{align}	
But  we know that $e=\dfrac{1}{\tilde{n}}\text{tr} [\Sigmam (x\Sigmam-z{\bf I})^{-1}] $ and $x=\dfrac{1}{1+e}$. Hence, 
\begin{align}
\dfrac{1}{\tilde{n}} \text{tr}[ \Sm\Qm] \asymp \dfrac{e}{1+e}. \label{eqn:easymp}
\end{align}	
From (\ref{eqn:easymp}), $\hat{e}$ can be estimated as follows:
\begin{equation}
\label{eqn:e_hat}
\hat{e} = \dfrac{\frac{1}{\tilde{n}} \text{tr}[ \Sm\Qm] }{1-\frac{1}{\tilde{n}} \text{tr}[ \Sm\Qm] },
\end{equation}
which satisfies  $e \asymp \hat{e}$. Note that $\hat{x}$ can be estimated from
\begin{equation}
\hat{x} = \dfrac{1}{1 + \hat{e}}.
\end{equation}
We can  differentiate  (\ref{eqn:systeqn}) to obtain
\begin{equation}
e^{\prime} = \dfrac{\frac{1}{\tilde{n}}\trace\Big[\Sigmam (x\Sigmam-z{\bf I}_p)^{-2}\Big]}{1 - \phi(1+e)^{-2}}.
\end{equation}
Hence, we obtain $\hat{\phi}$ 
\begin{equation}
\hat{\phi} = \dfrac{\hat{e}^\prime - \dfrac{\dfrac{1}{\tilde{n}} \text{tr}[\Sm \Qm^2]}{(\hat{x}-z\hat{x}^\prime)}}{\hat{e}^\prime(1+\hat{e})^{-2}}.
\label{eqn:estPhi}
\end{equation}
where the second term in the numerator is obtained from (\ref{eqn:DE2}). We can obtain $\hat{\theta}_G$ as in (\ref{eqn:thetaG}) after manipulation.
To prove (\ref{eqn:Dc}), we can write
\begin{align}
\mv^T \Sm{\bf Q}^2 \boldsymbol{{\Sigma}}\Sm{\bf Q}^2 \mv &=z^2 \mv^T {\bf Q}^2 \boldsymbol{{\Sigma}}{\bf Q}^2 \mv + \mv^T {\bf Q} \boldsymbol{{\Sigma}}{\bf Q} \mv \nonumber\\
&+2z \mv^T {\bf Q}^2 \boldsymbol{{\Sigma}}{\bf Q} \mv, \label{eqn:totalmQSQm}
\end{align}
where ${\bf Q}$ is defined in Table~\ref{tab:def}.
Then we find the consistent estimator of each term separately. Starting with the term $\mv^T {\bf Q} \boldsymbol{{\Sigma}}{\bf Q} \mv $, and noting that from (\ref{eqn:Sy}) we can express $\Sm$ in terms of the column vectors of ${\bf Y}$, ${\bf y}_i$, as follows:
\begin{align}
\mv^T {\bf Q}  \Sm{\bf Q} \mv &= \dfrac{1}{\tilde{n}}\sum_{i=1}^{\tilde{n}} \mv^T {\bf Q} {\bf y}_i{\bf y}_i^T{\bf Q} \mv.
\end{align}
Applying the matrix inversion lemma (see lemma 3 in \cite{2019-li-robust}), we obtain
\begin{align}
\mv^T {\bf Q}  \Sm{\bf Q} \mv&=  \dfrac{1}{\tilde{n}}\sum_{i=1}^{\tilde{n}}\dfrac{\mv^T {\bf Q}_i {\bf y}_i{\bf y}_i^T{\bf Q}_i \mv }{(1 + \frac{1}{\tilde{n}} { {\bf y}_i^T}{\bf Q}_i{\bf y}_i)^2},
\end{align}
where $\Qm_i   \triangleq \left(  \frac{1}{\tilde{n} }\sum_{j\neq i}^{\tilde{n}} \yv_j\yv_j^T - z \Id_p  \right)^{-1} $.
Noticing that 
$\frac{1}{\tilde{n}} {\bf y}_i^T{\bf Q}_i{\bf y}_i \asymp  \frac{1}{\tilde{n}}\text{tr}[ \boldsymbol{{\Sigma}} {\bf Q}_i]$, we write
\begin{align}
\mv^T {\bf Q}  \Sm{\bf Q} \mv& \asymp \dfrac{1}{\tilde{n}}\sum_{i=1}^{\tilde{n}}\dfrac{\mv^T {\bf Q}_i {\bf y}_i{\bf y}_i^T{\bf Q}_i \mv }{(1 + \frac{1}{\tilde{n}}  \text{tr}[ \boldsymbol{{\Sigma}} {\bf Q}_i])^2}.
\end{align}
Taking the expectation over  ${\bf y}_i$, followed by rank-one perturbation lemma \cite{2016-muller-random}, \cite{2020-niyazi-asymptotic}, we obtain 
\begin{align}
\mv^T {\bf Q}  \Sm{\bf Q} \mv& \asymp \dfrac{1}{\tilde{n}}\sum_{i=1}^{\tilde{n}}\dfrac{\mv^T {\bf Q}_i \boldsymbol{{\Sigma}}{\bf Q}_i \mv }{(1 + \frac{1}{\tilde{n}}  \text{tr}[ \boldsymbol{{\Sigma}} {\bf Q}])^2}, 
\end{align}
which can be written as follows:
\begin{align}
\dfrac{1}{\tilde{n}}\sum_{i=1}^{\tilde{n}}\dfrac{\mv^T {\bf Q}_i \boldsymbol{{\Sigma}}{\bf Q}_i \mv }{(1 + \frac{1}{\tilde{n}}  \text{tr}[ \boldsymbol{{\Sigma}} {\bf Q}])^2}& = \dfrac{1}{\tilde{n}}\sum_{i=1}^{\tilde{n}}\dfrac{\mv^T ({\bf Q}_i-{\bf Q}) \boldsymbol{{\Sigma}}{\bf Q}_i \mv }{(1 + \frac{1}{\tilde{n}}  \text{tr}[ \boldsymbol{{\Sigma}} {\bf Q}])^2} \nonumber\\
&+  \dfrac{1}{\tilde{n}}\sum_{i=1}^{\tilde{n}}\dfrac{\mv^T{\bf Q} \boldsymbol{{\Sigma}}{\bf Q} \mv }{(1 + \frac{1}{\tilde{n}}  \text{tr}[ \boldsymbol{{\Sigma}} {\bf Q}])^2} \nonumber \\
&+  \dfrac{1}{\tilde{n}}\sum_{i=1}^{\tilde{n}}\dfrac{\mv^T {\bf Q} \boldsymbol{{\Sigma}}({\bf Q}_i-{\bf Q}) \mv }{(1 + \frac{1}{\tilde{n}}  \text{tr}[ \boldsymbol{{\Sigma}} {\bf Q}])^2}
\end{align}
The difference ${\bf Q}_i-{\bf Q}$ converges to zero, so we can write
\begin{align}
\mv^T {\bf Q}  \Sm{\bf Q} \mv&\asymp  \dfrac{\mv^T{\bf Q} \boldsymbol{{\Sigma}}{\bf Q} \mv }{(1 + \frac{1}{\tilde{n}}  \text{tr}[ \boldsymbol{{\Sigma}} {\bf Q}])^2}
\end{align}
Noting that
\begin{equation}
\frac{1}{\tilde{n}}  \text{tr}[ \boldsymbol{{\Sigma}} {\bf Q}] \asymp \dfrac{\frac{1}{\tilde{n}} \text{tr}[\Sm {\bf Q}]}{1-\frac{1}{\tilde{n}}\text{tr}[\Sm {\bf Q}]},
\end{equation}
we can write
\begin{equation}
\label{eqn:mQSQm}
\mv^T{\bf Q} \boldsymbol{{\Sigma}}{\bf Q} \mv \asymp \Big( 1 +  \dfrac{\frac{1}{\tilde{n}} \text{tr}[\Sm {\bf Q}]}{1-\frac{1}{\tilde{n}}\text{tr}[\Sm {\bf Q}]}\Big)^2 \mv^T {\bf Q}  \Sm{\bf Q} \mv
\end{equation}
We can follow the same steps used to obtain \eqref{eqn:mQSQm} to reach
\begin{equation}
\label{eqn:mQ2SQ2m}
\mv^T{\bf Q}^2 \boldsymbol{{\Sigma}}{\bf Q}^2 \mv \asymp \Big( 1 +  \dfrac{\frac{1}{\tilde{n}} \text{tr}[\Sm {\bf Q}]}{1-\frac{1}{\tilde{n}}\text{tr}[\Sm {\bf Q}]}\Big)^4 \mv^T {\bf Q}^2  \Sm{\bf Q}^2 \mv.
\end{equation}

For the term, $ \mv^T {\bf Q}^2 \boldsymbol{{\Sigma}}{\bf Q} \mv$ we can approach it differently by differentiating   (\ref{eqn:mQSQm}) and noting that 
all the $\mv\mv^T$, $\boldsymbol{ \Sigma}$ and ${\bf Q}$ are symmetric matrices. The consistent estimator of  $ \mv^T {\bf Q}^2 \boldsymbol{{\Sigma}}{\bf Q} \mv$  is obtained from 
\begin{align}
&	\mv^T {\bf Q}^2 \boldsymbol{{\Sigma}}{\bf Q} \mv \nonumber\\
&\asymp \hat{e}^\prime (1+\hat{e})\mv^T {\bf Q} \Sm{\bf Q} \mv+(1+\hat{e})^2\mv^T {\bf Q}^2 \Sm{\bf Q} \mv \label{eqn:mQ2SQm}.
\end{align}
By substituting (\ref{eqn:mQSQm}), (\ref{eqn:mQ2SQ2m}) and  (\ref{eqn:mQ2SQm}) in (\ref{eqn:totalmQSQm}) and noticing that $z=-\gamma$, we obtain the consistent estimator of $D\big ( \mv_{0},\mv_{1}, \Hm, \Sigmam\big)$ in  (\ref{eqn:Dc}).
\end{IEEEproof}

\bibliographystyle{IEEEtran}
\bibliography{IEEEabrv,RLDAbibfile}

\begin{thebibliography}{10}
\providecommand{\url}[1]{#1}
\csname url@samestyle\endcsname
\providecommand{\newblock}{\relax}
\providecommand{\bibinfo}[2]{#2}
\providecommand{\BIBentrySTDinterwordspacing}{\spaceskip=0pt\relax}
\providecommand{\BIBentryALTinterwordstretchfactor}{4}
\providecommand{\BIBentryALTinterwordspacing}{\spaceskip=\fontdimen2\font plus
\BIBentryALTinterwordstretchfactor\fontdimen3\font minus
  \fontdimen4\font\relax}
\providecommand{\BIBforeignlanguage}[2]{{%
\expandafter\ifx\csname l@#1\endcsname\relax
\typeout{** WARNING: IEEEtran.bst: No hyphenation pattern has been}%
\typeout{** loaded for the language `#1'. Using the pattern for}%
\typeout{** the default language instead.}%
\else
\language=\csname l@#1\endcsname
\fi
#2}}
\providecommand{\BIBdecl}{\relax}
\BIBdecl

\bibitem{1936-fisher-use}
R.~A. Fisher, ``The use of multiple measurements in taxonomic problems,''
  \emph{Annals of eugenics}, vol.~7, no.~2, pp. 179--188, 1936.

\bibitem{2012-devijver-pattern}
P.~A. Devijver and J.~Kittler, \emph{Pattern recognition theory and
  applications}.\hskip 1em plus 0.5em minus 0.4em\relax Springer Science \&
  Business Media, 2012, vol.~30.

\bibitem{2015-chen-shrunken}
C.~Chen, Z.-M. Zhang, M.-L. Ouyang, X.~Liu, L.~Yi, Y.-Z. Liang, and C.-P.
  Zhang, ``Shrunken centroids regularized discriminant analysis as a promising
  strategy for metabolomics data exploration,'' \emph{Journal of Chemometrics},
  vol.~29, no.~3, pp. 154--164, 2015.

\bibitem{2013-jin-motor}
X.~Jin, M.~Zhao, T.~W. Chow, and M.~Pecht, ``Motor bearing fault diagnosis
  using trace ratio linear discriminant analysis,'' \emph{IEEE Transactions on
  Industrial Electronics}, vol.~61, no.~5, pp. 2441--2451, 2013.

\bibitem{1997-belhumeur-eigenfaces}
P.~N. Belhumeur, J.~P. Hespanha, and D.~J. Kriegman, ``Eigenfaces vs.
  fisherfaces: Recognition using class specific linear projection,'' \emph{IEEE
  Transactions on pattern analysis and machine intelligence}, vol.~19, no.~7,
  pp. 711--720, 1997.

\bibitem{2017-wang-locality}
Q.~Wang, Z.~Meng, and X.~Li, ``Locality adaptive discriminant analysis for
  spectral--spatial classification of hyperspectral images,'' \emph{IEEE
  Geoscience and Remote Sensing Letters}, vol.~14, no.~11, pp. 2077--2081,
  2017.

\bibitem{1962-anderson-introduction}
T.~W. Anderson, ``An introduction to multivariate statistical analysis,'' Wiley
  New York, Tech. Rep., 1962.

\bibitem{2020-khalili}
K.~{Elkhalil}, A.~{Kammoun}, R.~{Couillet}, T.~Y. {Al-Naffouri}, and
  M.~{Alouini}, ``A large dimensional study of regularized discriminant
  analysis,'' \emph{IEEE Transactions on Signal Processing}, vol.~68, pp.
  2464--2479, 2020.

\bibitem{2009-bandos-classification}
T.~V. Bandos, L.~Bruzzone, and G.~Camps-Valls, ``Classification of
  hyperspectral images with regularized linear discriminant analysis,''
  \emph{IEEE Transactions on Geoscience and Remote Sensing}, vol.~47, no.~3,
  pp. 862--873, 2009.

\bibitem{2004-ledoit-well}
O.~Ledoit and M.~Wolf, ``A well-conditioned estimator for large-dimensional
  covariance matrices,'' \emph{Journal of multivariate analysis}, vol.~88,
  no.~2, pp. 365--411, 2004.

\bibitem{2005-schafer-shrinkage}
J.~Sch{\"a}fer and K.~Strimmer, ``A shrinkage approach to large-scale
  covariance matrix estimation and implications for functional genomics,''
  \emph{Statistical applications in genetics and molecular biology}, vol.~4,
  no.~1, 2005.

\bibitem{2019-ollila-optimal}
E.~Ollila and E.~Raninen, ``Optimal shrinkage covariance matrix estimation
  under random sampling from elliptical distributions,'' \emph{IEEE
  Transactions on Signal Processing}, vol.~67, no.~10, pp. 2707--2719, 2019.

\bibitem{2010-chen-shrinkage}
Y.~Chen, A.~Wiesel, Y.~C. Eldar, and A.~O. Hero, ``Shrinkage algorithms for
  mmse covariance estimation,'' \emph{IEEE Transactions on Signal Processing},
  vol.~58, no.~10, pp. 5016--5029, 2010.

\bibitem{2008-stoica-using}
P.~Stoica, J.~Li, X.~Zhu, and J.~R. Guerci, ``On using a priori knowledge in
  space-time adaptive processing,'' \emph{IEEE transactions on signal
  processing}, vol.~56, no.~6, pp. 2598--2602, 2008.

\bibitem{2011-fisher-improved}
T.~J. Fisher and X.~Sun, ``Improved stein-type shrinkage estimators for the
  high-dimensional multivariate normal covariance matrix,'' \emph{Computational
  Statistics \& Data Analysis}, vol.~55, no.~5, pp. 1909--1918, 2011.

\bibitem{2013-halbe-regularized}
Z.~Halbe, M.~Bortman, and M.~Aladjem, ``Regularized mixture density estimation
  with an analytical setting of shrinkage intensities,'' \emph{IEEE
  transactions on neural networks and learning systems}, vol.~24, no.~3, pp.
  460--470, 2013.

\bibitem{2012-chen-shrinkage}
X.~Chen, Z.~J. Wang, and M.~J. McKeown, ``Shrinkage-to-tapering estimation of
  large covariance matrices,'' \emph{IEEE Transactions on Signal Processing},
  vol.~60, no.~11, pp. 5640--5656, 2012.

\bibitem{1989-friedman-regularized}
J.~H. Friedman, ``Regularized discriminant analysis,'' \emph{Journal of the
  American statistical association}, vol.~84, no. 405, pp. 165--175, 1989.

\bibitem{2014-lancewicki-multi}
T.~Lancewicki and M.~Aladjem, ``Multi-target shrinkage estimation for
  covariance matrices,'' \emph{IEEE Transactions on Signal Processing},
  vol.~62, no.~24, pp. 6380--6390, 2014.

\bibitem{2019-zhang-improved}
B.~Zhang, J.~Zhou, and J.~Li, ``Improved covariance matrix estimators by
  multi-penalty regularization,'' in \emph{2019 22th International Conference
  on Information Fusion (FUSION)}.\hskip 1em plus 0.5em minus 0.4em\relax IEEE,
  2019, pp. 1--7.

\bibitem{2021-raninen-linear}
E.~Raninen, D.~E. Tyler, and E.~Ollila, ``Linear pooling of sample covariance
  matrices,'' \emph{IEEE Transactions on Signal Processing}, 2021.

\bibitem{1962-horel-applications}
A.~Horel, ``Applications of ridge analysis to regression problems,''
  \emph{Chem. Eng. Progress.}, vol.~58, pp. 54--59, 1962.

\bibitem{1976-di-application}
P.~J. Di~Pillo, ``The application of bias to discriminant analysis,''
  \emph{Communications in Statistics-Theory and Methods}, vol.~5, no.~9, pp.
  843--854, 1976.

\bibitem{2012-ledoit-nonlinear}
O.~Ledoit and M.~Wolf, ``Nonlinear shrinkage estimation of large-dimensional
  covariance matrices,'' \emph{The Annals of Statistics}, vol.~40, no.~2, pp.
  1024--1060, 2012.

\bibitem{2015-ledoit-spectrum}
------, ``Spectrum estimation: A unified framework for covariance matrix
  estimation and pca in large dimensions,'' \emph{Journal of Multivariate
  Analysis}, vol. 139, pp. 360--384, 2015.

\bibitem{2018-ledoit-optimal}
------, ``Optimal estimation of a large-dimensional covariance matrix under
  stein’s loss,'' \emph{Bernoulli}, vol.~24, no.~4B, pp. 3791--3832, 2018.

\bibitem{2020-ledoit-analytical}
------, ``Analytical nonlinear shrinkage of large-dimensional covariance
  matrices,'' \emph{The Annals of Statistics}, vol.~48, no.~5, pp. 3043--3065,
  2020.

\bibitem{2020-ledoit-power}
------, ``The power of (non-) linear shrinking: a review and guide to
  covariance matrix estimation,'' \emph{University of Zurich, Department of
  Economics, Working Paper}, no. 323, 2020.

\bibitem{2014-abadir-design}
K.~M. Abadir, W.~Distaso, and F.~{\v{Z}}ike{\v{s}}, ``Design-free estimation of
  variance matrices,'' \emph{Journal of Econometrics}, vol. 181, no.~2, pp.
  165--180, 2014.

\bibitem{2017-kuismin-estimation}
M.~O. Kuismin and M.~J. Sillanp{\"a}{\"a}, ``Estimation of covariance and
  precision matrix, network structure, and a view toward systems biology,''
  \emph{Wiley Interdisciplinary Reviews: Computational Statistics}, vol.~9,
  no.~6, p. e1415, 2017.

\bibitem{2020-kang-improvedprecision}
X.~Kang and X.~Deng, ``An improved modified cholesky decomposition approach for
  precision matrix estimation,'' \emph{Journal of Statistical Computation and
  Simulation}, vol.~90, no.~3, pp. 443--464, 2020.

\bibitem{2020-bilgrau-targeted}
A.~E. Bilgrau, C.~F. Peeters, P.~S. Eriksen, M.~B{\o}gsted, and W.~N.
  Van~Wieringen, ``Targeted fused ridge estimation of inverse covariance
  matrices from multiple high-dimensional data classes,'' \emph{The Journal of
  Machine Learning Research}, vol.~21, no.~1, pp. 946--997, 2020.

\bibitem{2019-van-generalized}
W.~N. van Wieringen, ``The generalized ridge estimator of the inverse
  covariance matrix,'' \emph{Journal of Computational and Graphical
  Statistics}, vol.~28, no.~4, pp. 932--942, 2019.

\bibitem{2016-van-ridge}
W.~N. Van~Wieringen and C.~F. Peeters, ``Ridge estimation of inverse covariance
  matrices from high-dimensional data,'' \emph{Computational Statistics \& Data
  Analysis}, vol. 103, pp. 284--303, 2016.

\bibitem{2017-kuismin-precision}
M.~Kuismin, J.~Kemppainen, and M.~Sillanp{\"a}{\"a}, ``Precision matrix
  estimation with rope,'' \emph{Journal of Computational and Graphical
  Statistics}, vol.~26, no.~3, pp. 682--694, 2017.

\bibitem{2021-auguin-large}
N.~Auguin, D.~Morales-Jimenez, and M.~R. McKay, ``Large-dimensional
  characterization of robust linear discriminant analysis,'' \emph{IEEE
  Transactions on Signal Processing}, vol.~69, pp. 2625--2638, 2021.

\bibitem{2015-zollanvari}
A.~{Zollanvari} and E.~R. {Dougherty}, ``Generalized consistent error estimator
  of linear discriminant analysis,'' \emph{IEEE Transactions on Signal
  Processing}, vol.~63, no.~11, pp. 2804--2814, 2015.

\bibitem{2013-gareth-introduction}
J.~Gareth, W.~Daniela, H.~Trevor, and T.~Robert, \emph{An introduction to
  statistical learning: with applications in R}.\hskip 1em plus 0.5em minus
  0.4em\relax Spinger, 2013.

\bibitem{2007-guo-regularized}
Y.~Guo, T.~Hastie, and R.~Tibshirani, ``Regularized linear discriminant
  analysis and its application in microarrays,'' \emph{Biostatistics}, vol.~8,
  no.~1, pp. 86--100, 2007.

\bibitem{1997-chandrasekaran-parameter}
S.~Chandrasekaran, G.~Golub, M.~Gu, and A.~H. Sayed, ``Parameter estimation in
  the presence of bounded modeling errors,'' \emph{IEEE Signal Processing
  Letters}, vol.~4, no.~7, pp. 195--197, 1997.

\bibitem{2016-muller-random}
A.~Muller and M.~Debbah, ``Random matrix theory tutorial. introduction to
  deterministic equivalents,'' \emph{Traitement du signal}, vol.~33, no. 2-3,
  pp. 223--248, 2016.

\bibitem{2016-suliman-RMT}
M.~Suliman, T.~Ballal, A.~Kammoun, and T.~Y. Al-Naffouri, ``Constrained
  perturbation regularization approach for signal estimation using random
  matrix theory,'' \emph{IEEE Signal Processing Letters}, vol.~23, no.~12, pp.
  1727--1731, 2016.

\bibitem{2016-benaych-spectral}
F.~Benaych-Georges and R.~Couillet, ``Spectral analysis of the gram matrix of
  mixture models,'' \emph{ESAIM: Probability and Statistics}, vol.~20, pp.
  217--237, 2016.

\bibitem{2016-bakir-efficient}
D.~Bakir, A.~P. James, and A.~Zollanvari, ``An efficient method to estimate the
  optimum regularization parameter in rlda,'' \emph{Bioinformatics}, vol.~32,
  no.~22, pp. 3461--3468, 2016.

\bibitem{2009-hastie-elements}
T.~Hastie, R.~Tibshirani, J.~H. Friedman, and J.~H. Friedman, \emph{The
  elements of statistical learning: data mining, inference, and
  prediction}.\hskip 1em plus 0.5em minus 0.4em\relax Springer, 2009, vol.~2.

\bibitem{2018-ollila-matlab}
\BIBentryALTinterwordspacing
E.~Ollila and E.~Raninen. \BIBforeignlanguage{en}{Matlab regularizedscm toolbox
  version 1.0}. [Online]. Available:
  \url{http://users.spa.aalto.fi/esollila/regscm/,}
\BIBentrySTDinterwordspacing

\bibitem{2018-tabassum-compressive}
M.~N. Tabassum and E.~Ollila, ``Compressive regularized discriminant analysis
  of high-dimensional data with applications to microarray studies,'' in
  \emph{2018 IEEE International Conference on Acoustics, Speech and Signal
  Processing (ICASSP)}.\hskip 1em plus 0.5em minus 0.4em\relax IEEE, 2018, pp.
  4204--4208.

\bibitem{1999-mclachlan-mahalanobis}
G.~J. McLachlan, ``Mahalanobis distance,'' \emph{Resonance}, vol.~4, no.~6, pp.
  20--26, 1999.

\bibitem{1995-hastie-penalized}
T.~Hastie, A.~Buja, and R.~Tibshirani, ``Penalized discriminant analysis,''
  \emph{The Annals of Statistics}, pp. 73--102, 1995.

\bibitem{1989-le-handwritten}
Y.~Le~Cun, B.~Boser, J.~S. Denker, D.~Henderson, R.~E. Howard, W.~Hubbard, and
  L.~D. Jackel, ``Handwritten digit recognition with a back-propagation
  network,'' in \emph{Proceedings of the 2nd International Conference on Neural
  Information Processing Systems}, 1989, pp. 396--404.

\bibitem{1995-grother-nist}
P.~J. Grother, ``Nist special database 19 handprinted forms and characters
  database,'' \emph{National Institute of Standards and Technology}, 1995.

\bibitem{1998-lecun-gradient}
Y.~LeCun, L.~Bottou, Y.~Bengio, and P.~Haffner, ``Gradient-based learning
  applied to document recognition,'' \emph{Proceedings of the IEEE}, vol.~86,
  no.~11, pp. 2278--2324, 1998.

\bibitem{colonoscopyDataset}
P.~Mesejo and D.~Pizarro, ``{Gastrointestinal Lesions in Regular
  Colonoscopy},'' UCI Machine Learning Repository, 2016, {DOI}:
  https://doi.org/10.24432/C5V02D.

\bibitem{2012-rubio}
F.~{Rubio}, X.~{Mestre}, and D.~P. {Palomar}, ``Performance analysis and
  optimal selection of large minimum variance portfolios under estimation
  risk,'' \emph{IEEE Journal of Selected Topics in Signal Processing}, vol.~6,
  no.~4, pp. 337--350, 2012.

\bibitem{2016-schott-matrix}
J.~R. Schott, \emph{Matrix analysis for statistics}.\hskip 1em plus 0.5em minus
  0.4em\relax John Wiley \& Sons, 2016.

\bibitem{2009-rubio}
F.~{Rubio} and X.~{Mestre}, ``Consistent reduced-rank lmmse estimation with a
  limited number of samples per observation dimension,'' \emph{IEEE
  Transactions on Signal Processing}, vol.~57, no.~8, pp. 2889--2902, 2009.

\bibitem{2019-li-robust}
Q.~Li, P.~de~Kerret, D.~Gesbert, and N.~Gresset, ``Robust regularized zf in
  cooperative broadcast channel under distributed csit,'' \emph{IEEE
  Transactions on Information Theory}, vol.~66, no.~3, pp. 1845--1860, 2019.

\bibitem{2020-niyazi-asymptotic}
L.~B. Niyazi, A.~Kammoun, H.~Dahrouj, M.-S. Alouini, and T.~Y. Al-Naffouri,
  ``Asymptotic analysis of an ensemble of randomly projected linear
  discriminants,'' \emph{IEEE Journal on Selected Areas in Information Theory},
  vol.~1, no.~3, pp. 914--930, 2020.

\end{thebibliography}

\end{document}